\documentclass[nohyperref]{article}

\usepackage{microtype}
\usepackage{graphicx}
\usepackage{subfigure}
\usepackage{booktabs} 

\usepackage{hyperref}

\usepackage[utf8]{inputenc} 
\usepackage[T1]{fontenc}    
\usepackage{tcolorbox}
\usepackage{wrapfig}
\usepackage{hyperref}
\usepackage{algorithm2e}
\RestyleAlgo{ruled}

\hypersetup{colorlinks = true,
            linkcolor = blue,
            urlcolor  = blue,
            citecolor = blue,
            anchorcolor = blue}
\usepackage{url}            
\usepackage{booktabs}       
\usepackage{amsfonts}       
\usepackage{nicefrac}       
\usepackage{microtype}      
\usepackage{xcolor}         
\usepackage{graphicx}
\usepackage{subfigure}
\usepackage{amsmath}
\usepackage{tabularx}
\usepackage{wrapfig}

\usepackage[accepted]{icml2023}

\usepackage{amsmath}
\usepackage{amssymb}
\usepackage{mathtools}
\usepackage{amsthm}

\usepackage[capitalize,noabbrev]{cleveref}

\usepackage[textsize=tiny]{todonotes}

\newcommand{\namel}{\textsc{IER}}
\newcommand{\namek}{\textsc{H-IER}}
\newcommand{\namelo}{\textsc{RER}}
\newcommand{\namep}{\textsc{PER}}
\newcommand{\nameu}{\textsc{UER}}
\newcommand{\nameh}{\textsc{HER}}
\newcommand{\namet}{\textsc{TER}}
\newcommand{\nameo}{\textsc{OER}}
\newcommand{\nameul}{\textsc{U-IER}}

\icmltitlerunning{Introspective Experience Replay}

\begin{document}

\twocolumn[
\icmltitle{Introspective Experience Replay: Look Back When Surprised}



\icmlsetsymbol{equal}{*}

\begin{icmlauthorlist}
\icmlauthor{Ramnath Kumar}{goog}
\icmlauthor{Dheeraj Nagaraj}{goog}
\end{icmlauthorlist}

\icmlaffiliation{goog}{Google AI Research Lab,Bengaluru, India 560016}

\icmlcorrespondingauthor{Ramnath Kumar}{ramnathk@google.com}
\icmlcorrespondingauthor{Dheeraj Nagaraj}{dheerajnagaraj@google.com}

\icmlkeywords{Machine Learning, ICML}

\vskip 0.3in
]



\printAffiliationsAndNotice{}  

\begin{abstract}


In reinforcement learning (RL), experience replay-based sampling techniques play a crucial role in promoting convergence by eliminating spurious correlations. However, widely used methods such as uniform experience replay (\nameu) and prioritized experience replay (\namep) have been shown to have sub-optimal convergence and high seed sensitivity respectively. To address these issues, we propose a novel approach called Introspective Experience Replay (\namel) that selectively samples batches of data points prior to surprising events. Our method builds upon the theoretically sound reverse experience replay (\namelo) technique, which has been shown to reduce bias in the output of Q-learning-type algorithms with linear function approximation. However, this approach is not always practical or reliable when using neural function approximation. Through empirical evaluations, we demonstrate that \namel~with neural function approximation yields reliable and superior performance compared to \nameu, \namep, and hindsight experience replay (\nameh) across most tasks.

\end{abstract}

\section{Introduction}



Reinforcement learning (RL) involves learning with dependent data derived from trajectories of Markov processes. In this setting, the iterations of descent algorithms designed for i.i.d. data co-evolve with the trajectory at hand, leading to poor convergence. Experience replay \citep{lin1992self} involves storing the received data points in a large buffer and producing a random sample from this buffer whenever the learning algorithm requires it. Therefore experience replay is usually deployed with popular algorithms like DQN, DDPG, and TD3 to achieve state-of-the-art performance \citep{mnih2015human,lillicrap2015continuous}. It has been shown experimentally \citep{mnih2015human} and theoretically \citep{bresler2020least} that these learning algorithms for Markovian data show sub-par performance without experience replay. 

The simplest and most widely used experience replay method is the uniform experience replay (\nameu), where the data points stored in the buffer are sampled uniformly at random every time a data point is queried \citep{mnih2015human}. However, \nameu~might pick uninformative data points most of the time, which may slow down the convergence. For this reason, optimistic experience replay (\nameo) and prioritized experience replay (\namep)~\citep{schaul2015prioritized} were introduced, where samples with higher TD error (i.e., `surprise') are sampled more often from the buffer. Optimistic experience replay (originally called ``greedy TD-error prioritization'') was shown to have a high bias\footnote{We use the term `bias' here as used in \cite{schaul2015prioritized}, which means biased with respect to the empirical distribution over the replay buffer}. This leads to the algorithm predominantly selecting rare data points and ignoring the rest. Prioritized experience replay was proposed to solve this issue \citep{schaul2015prioritized} by using a sophisticated sampling approach. However, as shown in our experiments outside of the Atari environments, \namep~still suffers from a similar problem, and its performance can be highly sensitive to seed and hyper-parameter. Although this speeds up the learning process in many cases, its performance can be quite bad and erratic due to picking and choosing only specific data points. The design of experience replay continues to be an active field of research. Several other experience replay techniques like Hindsight experience replay (\nameh)~\citep{andrychowicz2017hindsight}, Reverse Experience Replay (\namelo)~\citep{rotinov2019reverse}, and Topological Experience Replay (\namet)~\citep{hong2022topological} have been proposed. An overview of these methods is discussed in Section~\ref{rel_works}. 

\begin{figure}[hbt!]
\centering
\includegraphics[width=0.99\linewidth]{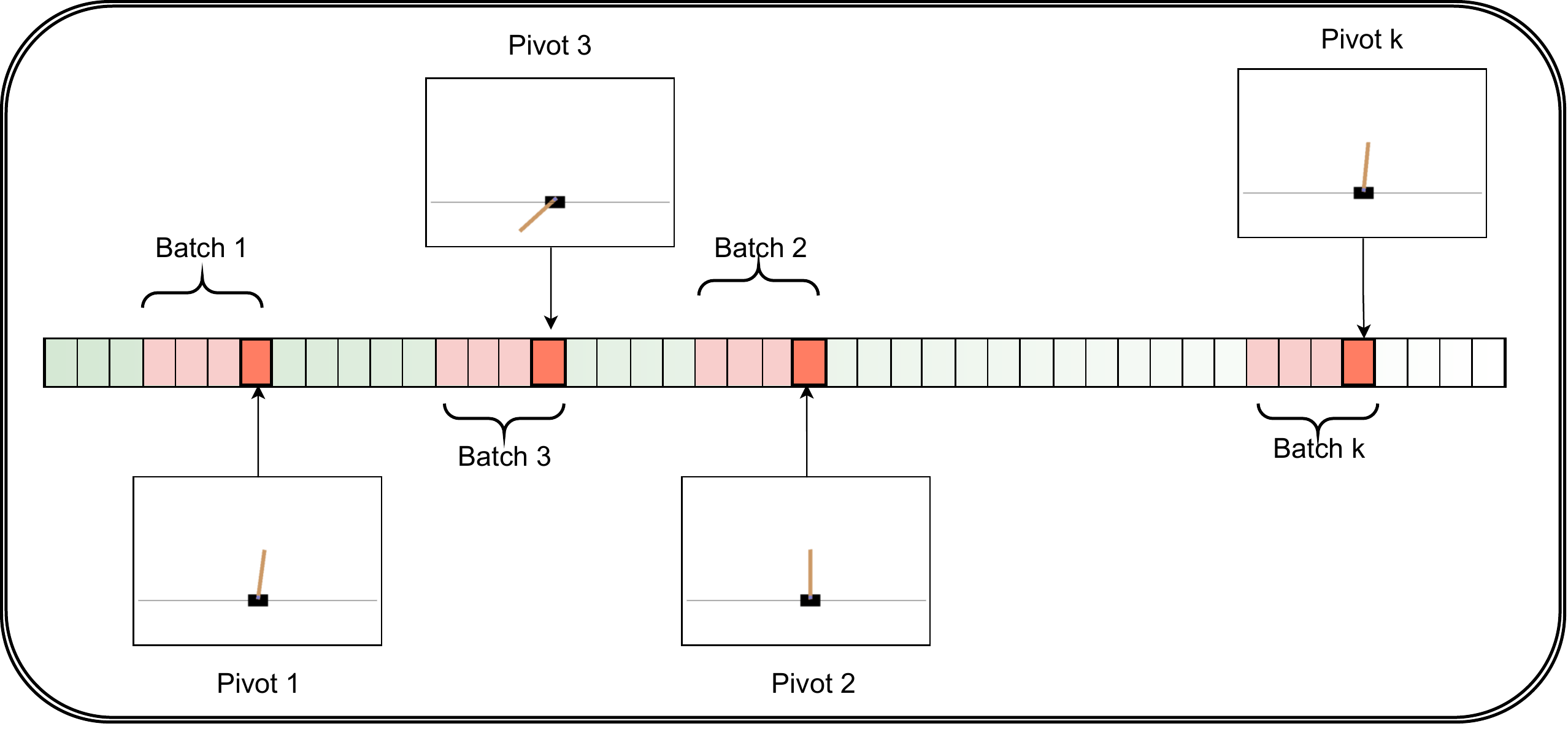}
\label{fig:rer++}
\caption{An illustration of our proposed methodology when selecting $k$ batches in the CartPole environment. The red color is used to indicate the batches being sampled from the replay buffer. The green samples are the un-sampled states from the buffer. The arrow explicitly points the pivots and the snapshot of the surprising state encountered.}
\label{fig:ro}
\end{figure}

Even though these methods are widely deployed in practice, theoretical analyses have been very limited. Recent results on learning dynamical systems \citep{kowshik2021streaming,kowshik2021near} showed rigorously in a theoretical setting that \namelo~is the conceptually-grounded algorithm when learning from Markovian data. Furthermore, this work was extended to the RL setting in \cite{agarwal2021online} to achieve efficient Q learning with linear function approximation. The \namelo~technique achieves good performance since reverse order sampling of the data points prevents the build-up of spurious correlations in the learning algorithm. In this paper, we build on this line of work and introduce \textbf{Introspective Experience Replay} (\namel). Roughly speaking, \namel~first picks top $k$ `pivot' points from a large buffer according to their TD error. It then returns batches of data formed by selecting the consecutive points \emph{temporally} before these pivot points. In essence, the algorithm \emph{looks back when surprised}. The intuition behind our approach is linked to the fact that the agent should always associate outcomes to its past actions, just like in \namelo. The summary of our approach is shown in Figure~\ref{fig:rer++}. This technique is an amalgamation of Reverse Experience Replay (\namelo) and Optimistic Experience Replay (\nameo), which only picks the points with the highest TD error. 



Our main findings are summarized below:

\textbf{Better Performance Against SOTA:}
Our proposed methodology (\namel) outperforms previous state-of-the-art baselines such as \namep, \nameu~and \nameh~on most environments (see Table~\ref{tab:initial_results}, Section~\ref{results}).

\textbf{Conceptual Understanding:} We consider a simple toy example where we understand the differences between \nameu, \namelo, \nameo~and \namel~(Section~\ref{sec:toy_example}). This study illustrates the better performance of our proposed method by showing a) why naive importance sampling, like in \nameo~suffers from poor performance, often failing to learn any non-trivial policy and b) why techniques like \nameu~and \namelo~are slow to learn. 

\textbf{Forward vs. Reverse:} We show that the temporal direction (forward/reverse) to consider after picking the pivot is non-trivial. We show empirically that \namel~performs much better than its forward counterpart, IER (F) (see Table~\ref{table-temporl-results}). This gives evidence that causality plays a role in the success of \namel. 

\textbf{Whole vs. Component Parts:} Our method (\namel) is obtained by combining two methods \namelo~(which picks the samples in the reverse order as received) and \nameo~(which greedily picks the samples with the largest TD error). We show that neither of these components performs well compared to their amalgamation, \namel~(Table~\ref{table:ablation}). 

\textbf{Minimal hyperparameter tuning:} Our proposed methodology uses minimal hyperparameter tuning. We use the same policy network architecture, learning rate, batch size, and all other parameters across all our runs for a given environment. These hyperparameters are selected based on the setting required to achieve SOTA performance on \nameu. However, we have few options available, a \textit{Hindsight} flag, leading to the H-IER sampler, and the \textit{Uniform Mixing fraction} leading to the \nameul~sampler. Furthermore, for most of our experiments, we use a mixing fraction of $0$.

\begin{table}[!hbt]\centering
    \vspace{-10pt}
\caption{\textbf{\namel~outperforms previous state-of-the-art baselines}. These baselines include samplers such as \nameu~\citep{mnih2013playing}, \namep~\citep{schaul2015prioritized}, and \nameh~\citep{andrychowicz2017hindsight} across many environments. Results are from 13 different environments that cover a broad category of MDPs. These include a few Atari environments (previously used to support the efficacy of \nameu, \namep), some Robotics environments (previously used to support the efficacy of \nameh), and many other classes of environments, including Classic Control, Box 2D, and Mujoco. More details about these experiments and their setup have been discussed in Section~\ref{results}.
}
\vspace*{1mm}
\label{tab:initial_results}
\resizebox{0.8\linewidth}{!}{
\begin{tabular}{lcccc}
\toprule
Experience Replay Method & {UER} & {PER} & {HER} & {IER} \\ \midrule
Best Performance Frequency & 1 & 0 & 1  & {\color[HTML]{3166FF} \textbf{11}}\\ \bottomrule
\end{tabular}
}
    \vspace{-10pt}
\end{table}

\section{Related Works and Comparison}
\label{rel_works}
\subsection{Experience Replay Techniques}
Experience replay involves storing consecutive temporally dependent data in a (large) buffer in a FIFO order. Whenever a learning algorithm queries for batched data, the experience replay algorithm returns a sub-sample from this buffer such that this data does not hinder the learning algorithms due to spurious correlations. The most basic form of experience replay is \nameu~\citep{lin1992self} which samples the data in the replay buffer uniformly at random. This approach has significantly improved the performance of off-policy RL algorithms like DQN \citep{mnih2015human}. Several other methods of sampling from the buffer have been proposed since; \namep~\citep{schaul2015prioritized} samples experiences from a probability distribution which assigns higher probability to experiences with significant TD error and is shown to boost the convergence speed of the algorithm. This outperforms \nameu~in most Atari environments. \nameh~\citep{andrychowicz2017hindsight} works in the "what if" scenario, where even a bad policy can lead the agent to learn what not to do and nudge the agent towards the correct action. There have also been other approaches such as \cite{liu2019competitive, fang2018dher, fang2019curriculum} have adapted \nameh~in order to improve the overall performance with varying intuition.
\namelo~processes the data obtained in a buffer in the reverse temporal order. We refer to the following sub-section for a detailed review of this and related techniques. We will also consider `optimistic experience replay' (OER), the naive version of \namep, where at each step, only top $B$ elements in the buffer are returned when batched data is queried. This approach can become highly selective to the point of completely ignoring certain data points, leading to poor performance. This is mitigated by a sophisticated sampling procedure employed in \namep. Other works such as \cite{fujimoto2020equivalence, pan2022understanding} attempt to study \namep~and address some of its shortcomings, and \cite{lahire2021large} introduces 'large batch experience replay' (LaBER), which reduces the stochastic noise in gradients while keeping the updates unbiased with respect to the empirical data distribution in the replay buffer 

\subsection{Reverse Sweep Techniques}
\label{sec:rev_sweep}
Reverse sweep or backward value iteration refers to methods that process the data as received in reverse order. This has been studied in the context of planning tabular MDPs \citep{dai2007prioritizing,grzes2013convergence}. We refer to Section~\ref{understanding_rer} for a brief overview of why these methods are considered. However, this line of work assumes that the MDP and the transition functions are known. Inspired by the behavior of biological networks, \cite{rotinov2019reverse} proposed reverse experience replay where the experience replay buffer is replayed in a LIFO order. Since \namelo~forms mini-batches with consecutive data points can be unreliable with Neural approximation (i.e., it does not learn consistently well across different environments). Therefore, the iterations of \ namelo~are `mixed' with \nameu. However, the experiments are limited and do not demonstrate that this method outperforms even \nameu. A similar procedure named Episodic Backward Update (EBU) is introduced in \cite{lee2019sample}. However, to ensure that the pure \namelo~works well, the EBU method also seeks to change the target for Q learning instead of just changing the sampling scheme in the replay buffer.
The reverse sweep was independently rediscovered as \namelo~in the context of streaming linear system identification in \cite{kowshik2021streaming}, where SGD with reverse experience replay was shown to achieve near-optimal performance. In contrast, naive SGD was significantly sub-optimal due to the coupling between the Markovian data and the SGD iterates. The follow-up work \cite{agarwal2021online} analyzed off-policy Q learning with linear function approximation and reverse experience replay to provide near-optimal convergence guarantees using the unique super martingale structure endowed by reverse experience replay. 
\cite{hong2022topological} considers topological experience replay, which executes reverse replay over a directed graph of observed transitions. Mixed with \namep~enables non-trivial learning in some challenging environments. Another line of work \citep{florensa2017reverse,moore1993prioritized,goyal2018recall,schroecker2019generative} considers reverse sweep with access to a simulator or using a fitted generative model. On the other hand, our work only seeks on-policy access to the MDP.

\section{Background and Proposed Methodology}
\label{prelim}
We consider episodic reinforcement learning \citep{sutton2018reinforcement}, where at each time step an agent takes actions $a_t$ in an uncertain environment with state $s_t$, and receives a reward $r_t$. The environment then evolves into a new state $s_{t+1}$ whose law depends only on $s_t,a_t$. Our goal is to (approximately) find the policy $\pi^{*}$ which maps the environmental state $s$ to an action $a$ such that when the agent takes the action $a_t = \pi^{*}(s_t)$, the discounted reward $\mathbb{E}\left[\sum_{t=0}^{\infty}\gamma^{t}r_t\right]$ is maximized. To achieve this, we consider algorithms like DQN (\cite{mnih2015human}), DDPG (\cite{lillicrap2015continuous}) and TD3 (\cite{fujimoto2018addressing}), which routinely use experience replay buffers. In this paper, we introduce a new experience replay method, \namel, and investigate the performance of the aforementioned RL algorithms with this modification. In this work, when we say ``return'', we mean discounted episodic reward. 
\begin{algorithm*}
\caption{Our proposed Introspective Experience Replay (\namel) for Reinforcement Learning}\label{alg:rev_plus}
 \KwIn{Data collection mechanism $\mathbb{T}$, Data buffer $\mathcal{H}$, Batch size $B$, grad steps per Epoch $G$, number of episodes $N$, Importance function $I$, learning procedure $\mathbb{A}$, Uniform Mixing fraction $p$}
 $n \leftarrow 0$\;
 
 \While{$n < N$}
 { $n \leftarrow n+1$\;

  $\mathcal{H} \leftarrow \mathbb{T}(\mathcal{H})$ \hspace*{\fill} \tcp*{\footnotesize Add a new episode to the buffer}

$\mathcal{I} \leftarrow I(\mathcal{H})$ \hspace*{\fill} \tcp*{\footnotesize Compute importance of each data point in the buffer} 

 $P \leftarrow \mathsf{Top}(\mathcal{I};G)$ \hspace*{\fill} \tcp*{\footnotesize Obtain index of top $G$ elements of $\mathcal{I}$} 
 $g \leftarrow 0$\;

 \While{$g< G$}{
 \eIf{$g<(1-p)G$}{
 $D \leftarrow \mathcal{H}[P[g]-B,P[g]]$ \hspace*{\fill} \tcp*{\footnotesize Load batch of previous $B$ examples from pivot $P[g]$} 
 }{
 $D \leftarrow \mathcal{H}[\mathsf{Uniform}(\mathcal{H}, B)]$ \hspace*{\fill} \tcp*{\footnotesize Randomly chose $B$ indices from buffer} 
 }
 $g\leftarrow g+1$\;

 $\mathbb{A}(D)$\hspace*{\fill} \tcp*{\footnotesize Run the learning algorithm with batch data $D$} 
 }
 }
\end{algorithm*}

\subsection{Methodology}
\label{method}
We now describe our main method in a general way where we assume that we have access to a data collection mechanism $\mathbb{T}$ which samples new data points. This then appends the sampled data points to a buffer $\mathcal{H}$ and discards some older data points. The goal is to run an iterative learning algorithm $\mathbb{A}$, which learns from batched data of batch size $B$ in every iteration. We also consider an important metric $I$ associated with the problem. At each step, the data collection mechanism $\mathbb{T}$ collects a new episode and appends it to the buffer $\mathcal{H}$ and discards some old data points, giving us the new buffer as $\mathcal{H} \leftarrow \mathbb{T}(\mathcal{H})$. We then sort the entries of $\mathcal{H}$ based on the importance metric $I$ and store the indices of the top $G$ data points in an array $P = [P[0],\dots,P[G-1]]$. Then for every index in $P$, we run the learning algorithm $\mathbb{A}$ with the batch $D = (\mathcal{H}(P[i]),\dots,\mathcal{H}(P[i]-B+1))$. In some cases, we can `mix'\footnote{Mixing here denotes sampling with a given probability from one sampler A, and filling the remaining samples of a batch with sampler B.} this with the standard \nameu~sampling mechanism to reduce bias of the stochastic gradients with respect to the empirical distribution in the buffer, as shown below. Our experiments show that this amalgamation helps convergence in certain cases. We describe this procedure in Algorithm~\ref{alg:rev_plus}. 



In the reinforcement learning setting, $\mathbb{T}$ runs an environment episode with the current policy and appends the transitions and corresponding rewards to the buffer $\mathcal{H}$ in the FIFO order, maintaining a total of $1E6$ data points, usually. We choose $\mathbb{A}$ to be an RL algorithm like TD3 or DQN or DDPG. The importance function $I$ is the magnitude of the TD error with respect to the current Q-value estimate provided by the algorithm $\mathbb{A}$ (i.e., $I = |Q(s,a) - R(s,a) - \gamma \sup_{a^{\prime}}Q^{\mathsf{target}}(s^{\prime},a^{\prime})|$). When the data collection mechanism ($\mathbb{T}$) is the same as in \nameu, we will call this method \textbf{\namel}. In optimistic experience replay (\nameo), we take $\mathbb{T}$ to be the same as in \nameu. However, we query top $BG$ data points from the buffer $\mathcal{H}$ and return $G$ disjoint batches each of size $B$ from these `important' points. It is clear that \namel~is a combination of \nameo~and \namelo. Notice that we can also consider the data collection mechanism like that of \nameh, where examples are labeled with different goals, i.e. $\mathbb{T}$ has now been made different, keeping the sampling process exactly same as before. In this case, we will call our algorithm \textbf{\namek}. Our experiment in Enduro and Acrobat depicts an example of this successful coalition. We also consider the \namelo~method, which served as a motivation for our proposed approach. Under this sampling methodology, the batches are drawn from $\mathcal{H}$ in the temporally reverse direction. This approach is explored in the works mentioned in Section~\ref{sec:rev_sweep}. We discuss this methodology in more detail in Appendix~\ref{rer}. 

\subsection{Didactic Toy Example}
\label{sec:toy_example}
In this section, we discuss the working of \namel~on a simple environment such as GridWorld-1D, and compare this with some of our baselines such as \nameu, \nameo, \namelo, and IER (F). In this environment, the agent lives on a discrete 1-dimensional grid of size 40 with a max-timestep of 1000 steps, and at each time step, the agent can either move left or right by one step. The agent starts from the \textit{starting state} (S; $[6]$), the goal of the agent is to reach \textit{goal state} (G; $[40]$) getting a reward of $+1$, and there is also a \textit{trap state} (T; $[3]$), where the agents gets a reward of $-2$. The reward in every other state is $0$. For simplicity, we execute an offline exploratory policy where the agent moves left or right with a probability of half and obtain a buffer of size 30000. The rewarding states occur very rarely in the buffer since it is hard to reach for this exploration policy. The episode ends upon meeting either of two conditions: (i) the agent reaches the terminal state, which is the \textit{goal state}, or (ii) the agent has exhausted the max-timestep condition and has not succeeded in reaching any terminal state. An overview of our toy environment is depicted in Figure~\ref{fig:gridworld_env}. Other hyperparameters crucial to replicating this experiment are described in Appendix~\ref{appendix-hyperparam}.

In this example, reaching the goal state as quickly as possible is vital to receive a positive reward and avoid the fail state. Therefore, it is essential to understand the paths which reach the goal state.
Figure~\ref{fig:gridworld_count} depicts the number of times each state occurs in the buffer. Furthermore, the remaining subplots of Figure~\ref{fig:gridworld_v_function} depict the \textit{Absolute Frequency} of our off-policy algorithm trained in this environment. A state's ``absolute frequency'' is the number of times the replay technique samples a given state during the algorithm's run. The experiments on this simple didactic toy environment do highlight a few interesting properties: \newline
\textbf{Comparison of \nameu~and \namel:} Since the goal state appears very rarely in buffer, \nameu~and \namelo~rarely sample the goal state and hence do not manage to learn effectively. While \namelo~naturally propagates the information about the reward back in time to the states that led to the reward, it does not often sample the rewarding state. \newline
\textbf{Limitation of \nameo:}  While \nameo~samples a lot from the states close to the goal state, the information about the reward does not propagate to the start state. We refer to the bottleneck in Figure~\ref{fig3} where some intermediate states are not sampled. \newline
\textbf{Advantage of \namel:} \namel~prioritizes sampling from the goal state and propagates the reward backward so that the entire path leading to the reward is now aware of how to reach the reward. Therefore, a combination of \namelo~and \nameo~reduces the sampling bias in \nameo~by preventing the bottlenecks seen in Figure~\ref{fig3}. \newline 
\textbf{Bottleneck of IER (F):} 
IER (F) has a more significant bottleneck when compared to \namelo~and chooses to sample the non-rewarding middle states most often. Also, note that whenever IER (F) chooses the goal state as the pivot, it selects the rest of the batch to overflow into the next episode, which begins at the starting state. This does not allow the algorithm to effectively learn the path which \emph{led} to the goal state. 


\begin{figure*}[hbt!]
\centering
\subfigure[GridWorld-1D environment]{%
\label{fig:gridworld_env}
\includegraphics[width=0.33\linewidth]{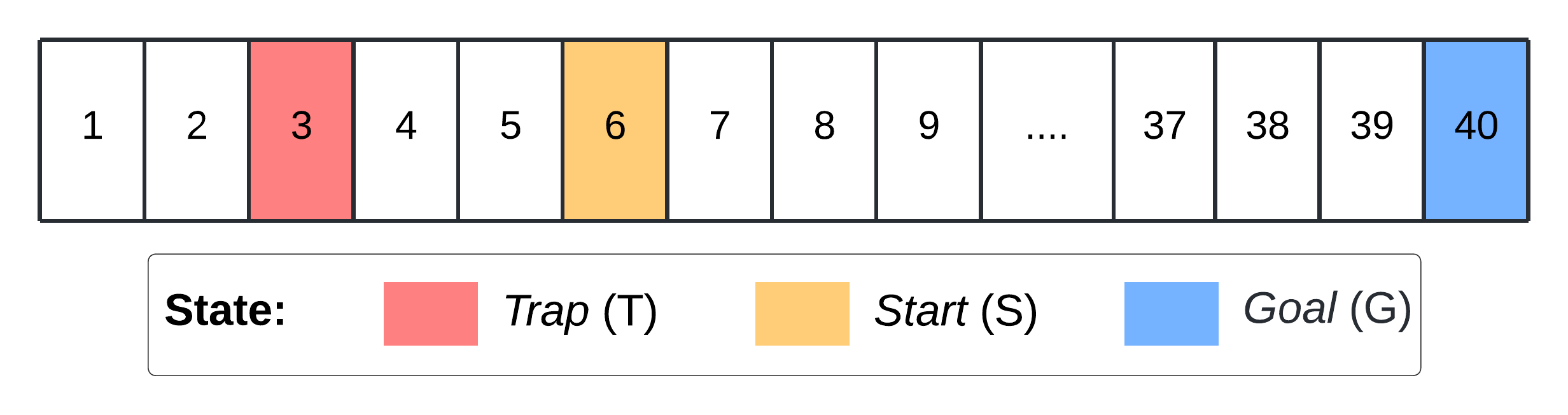}}
\quad
\subfigure[Buffer]{%
\label{fig:gridworld_count}
\includegraphics[width=0.22\linewidth]{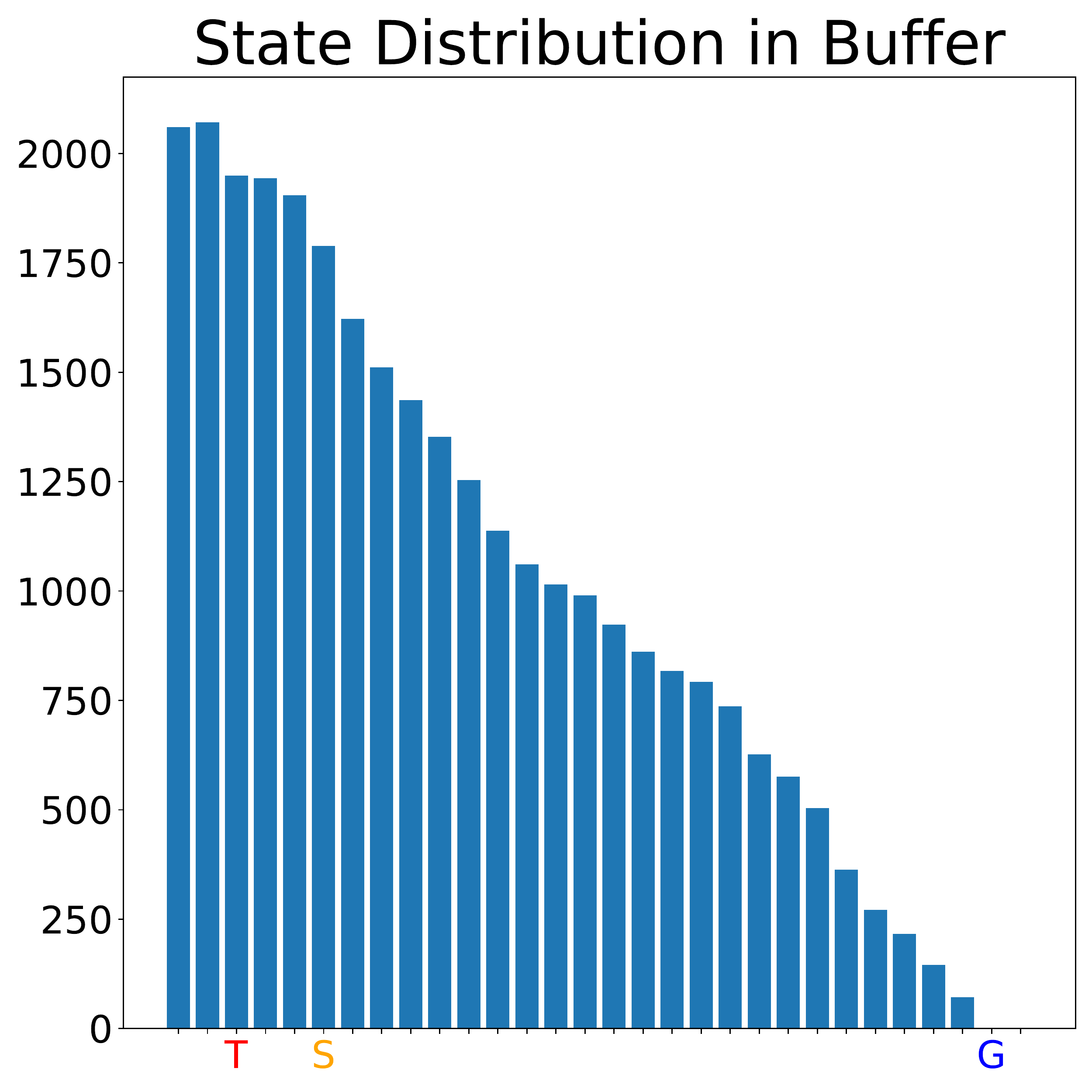}}
\quad
\subfigure[UER]{%
\label{fig1}
\includegraphics[width=0.22\linewidth]{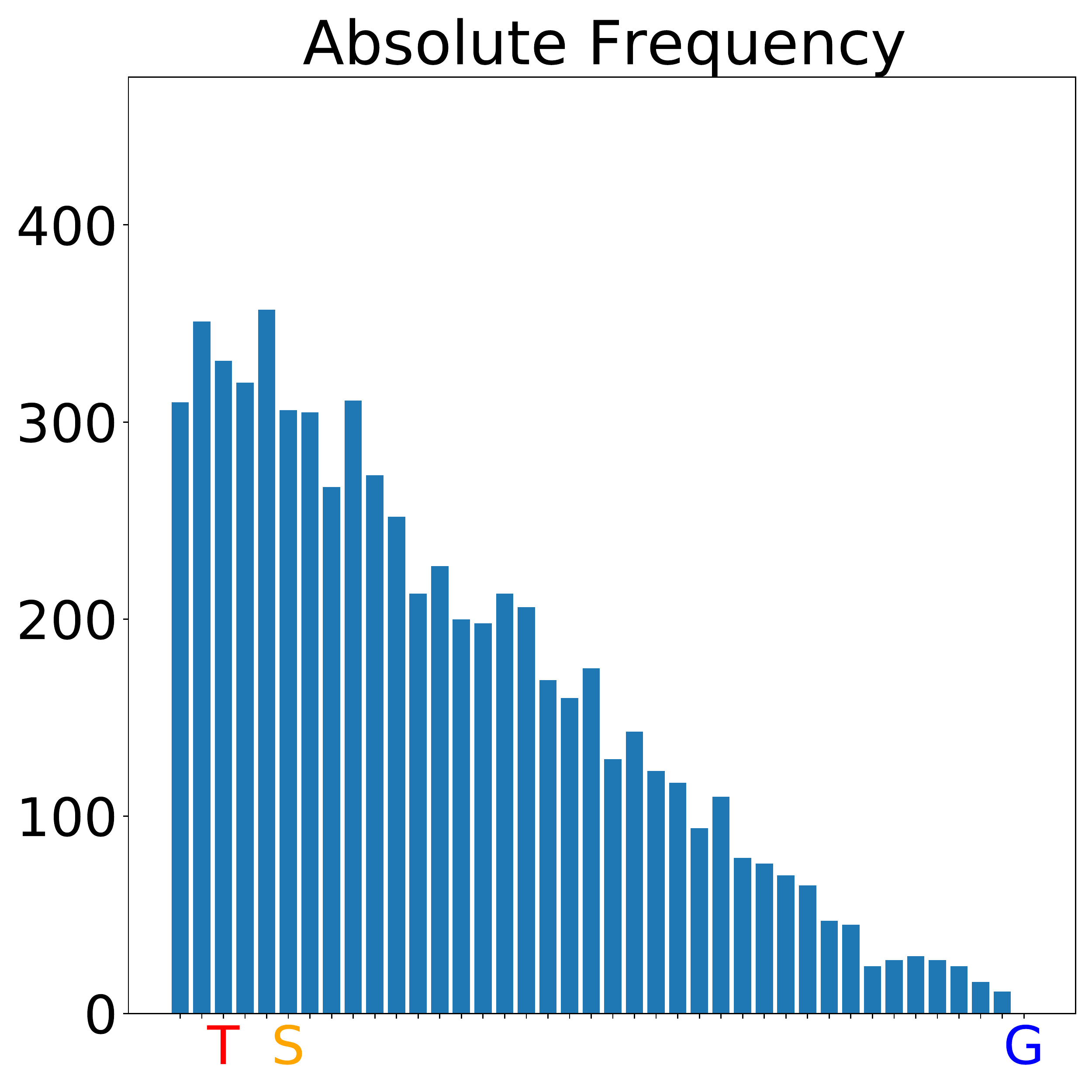}}

\subfigure[RER]{%
\label{fig2}
\includegraphics[width=0.22\linewidth]{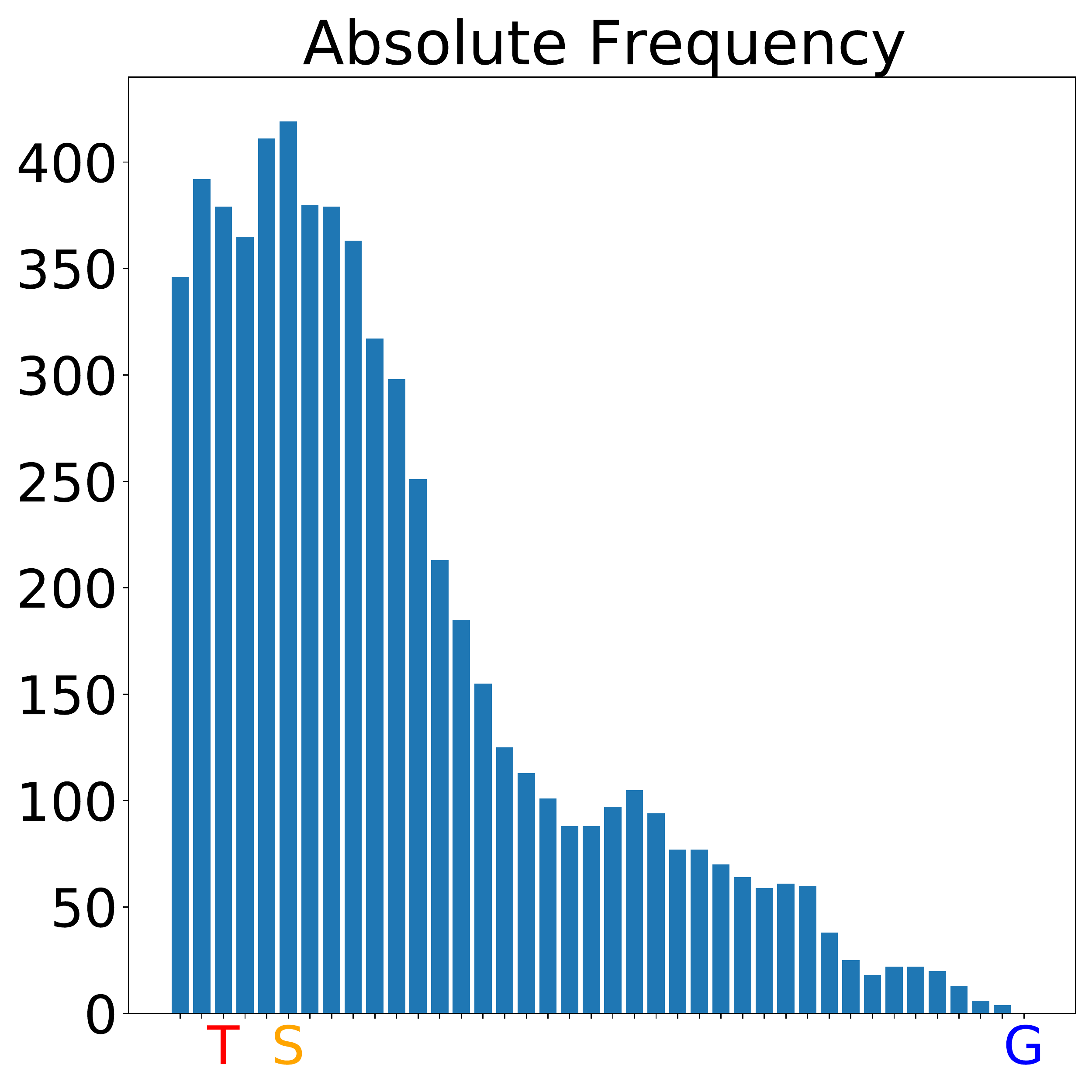}}
\quad
\subfigure[OER]{%
\label{fig3}
\includegraphics[width=0.22\linewidth]{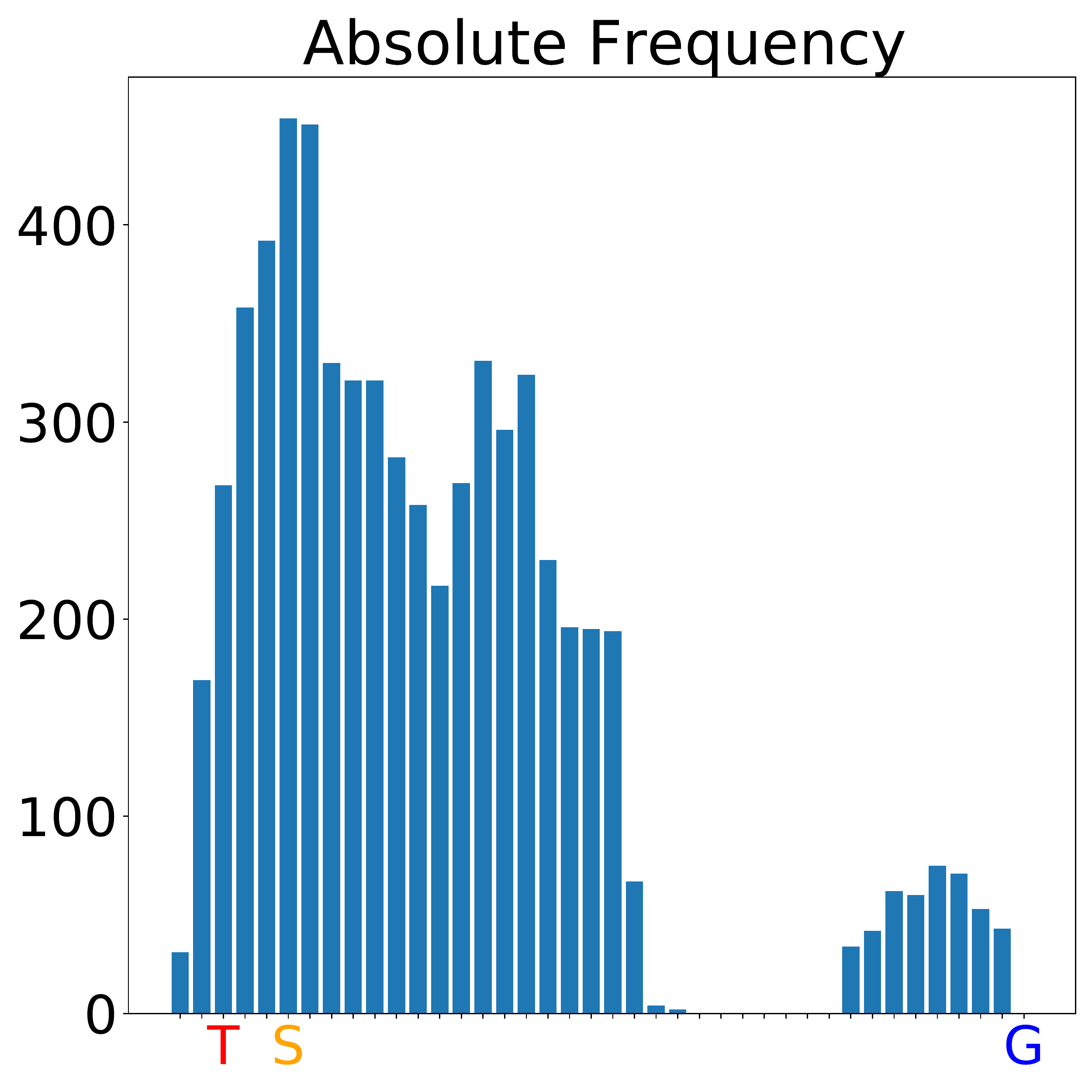}}
\quad
\subfigure[IER]{%
\label{fig4}
\includegraphics[width=0.22\linewidth]{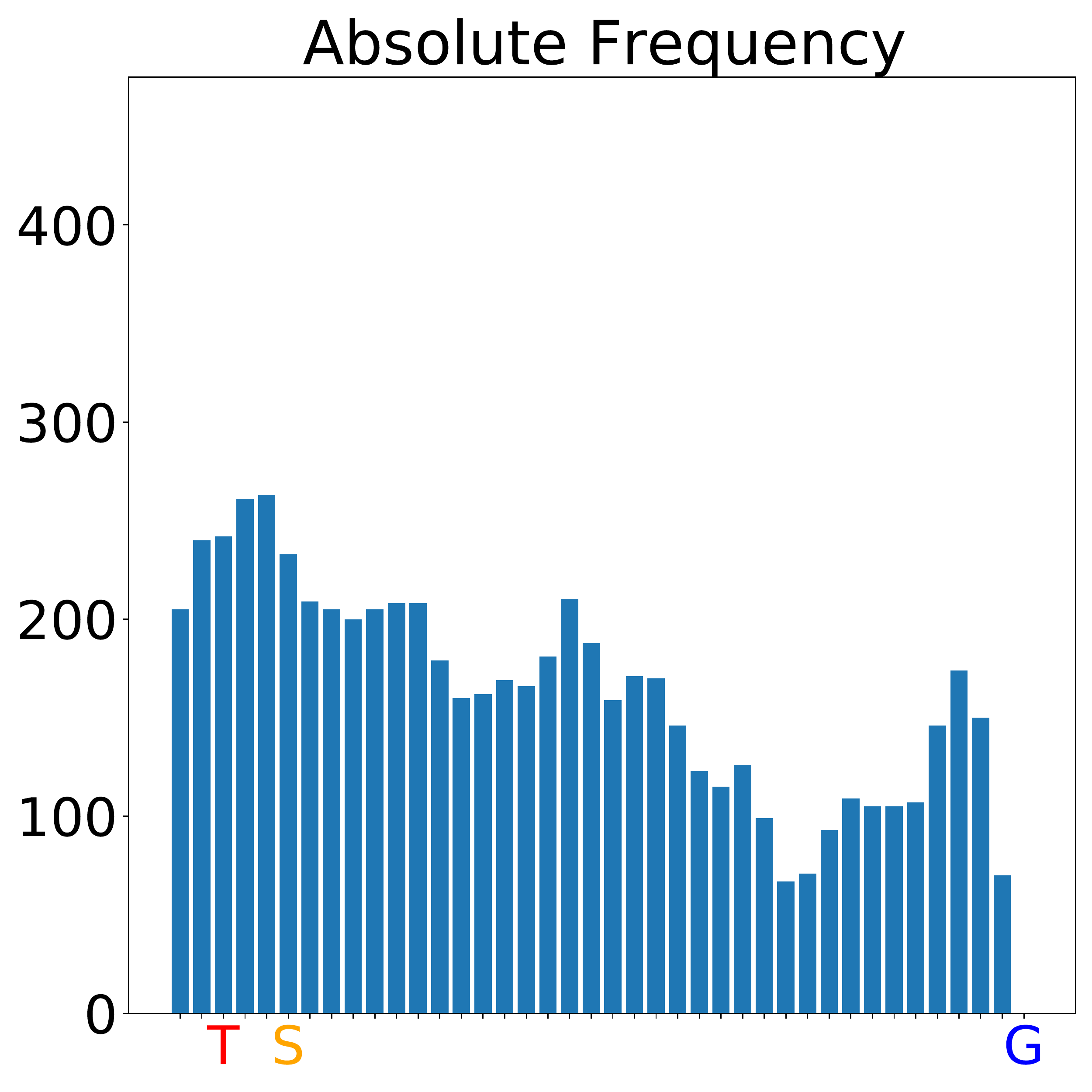}}
\quad
\subfigure[IER (F)]{%
\label{fig5}
\includegraphics[width=0.22\linewidth]{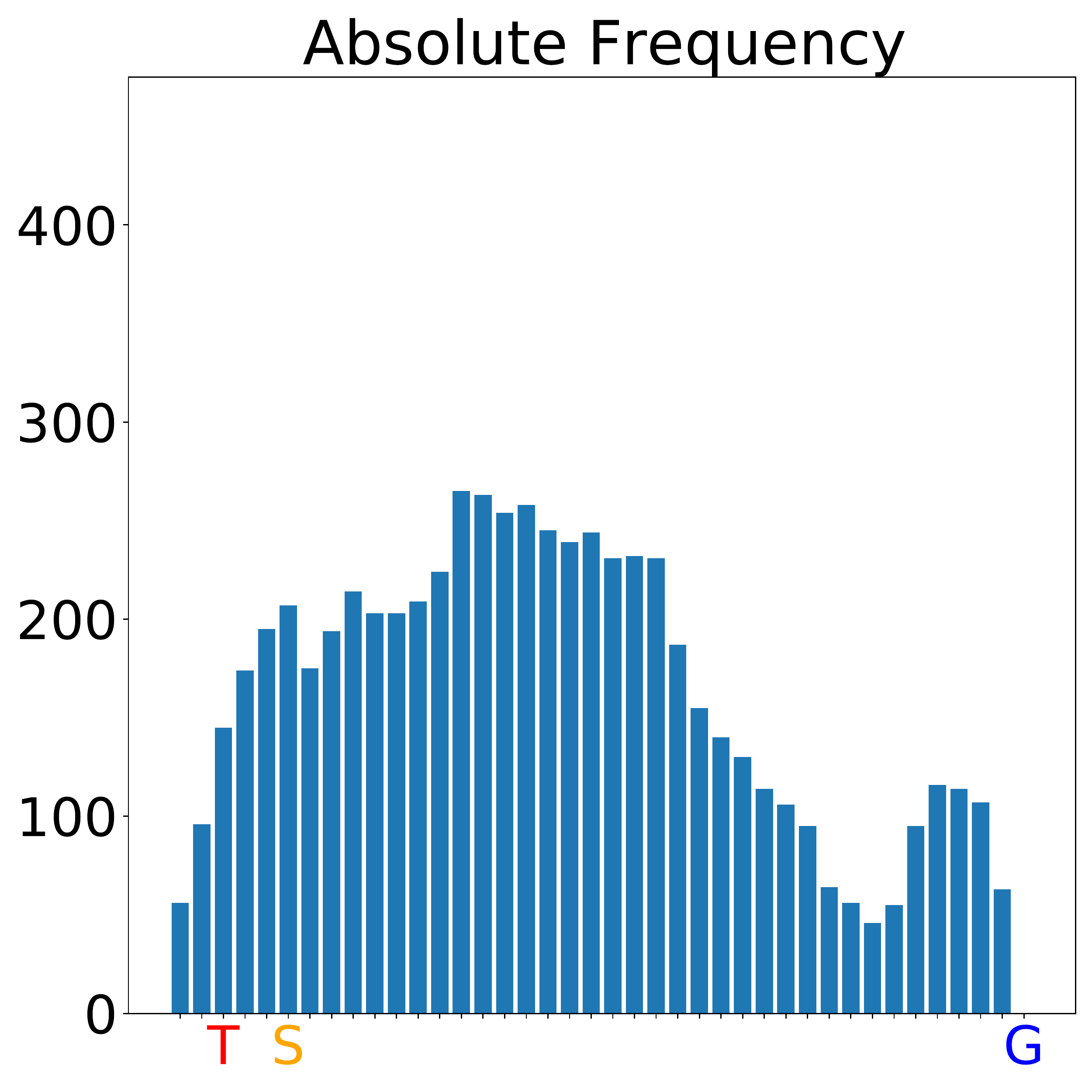}}
\caption{Gridworld-1D environment is depicted in Figure~\ref{fig:gridworld_env}. Distribution of states in the buffer (Figure~\ref{fig:gridworld_count}) and relative frequency of different experience replay samplers on the didactic example of GridWorld-1D environment (Figure~\ref{fig1};\ref{fig2};\ref{fig3};\ref{fig4};\ref{fig5}).}
\label{fig:gridworld_v_function}
\end{figure*}

The toy example above models several salient features in more complicated RL environments. 

(i) In the initial stages of learning, the exploratory policy is essentially random, and such a naive exploratory policy does not often lead to non-trivial rewards.

(ii)  Large positive and negative reward states (the goal and trap states), and their neighbors provide the pivot state for \namel~and \nameo.  

\begin{figure*}[hbt]
\centering
\subfigure[UER]{%
\includegraphics[width=0.3\linewidth]{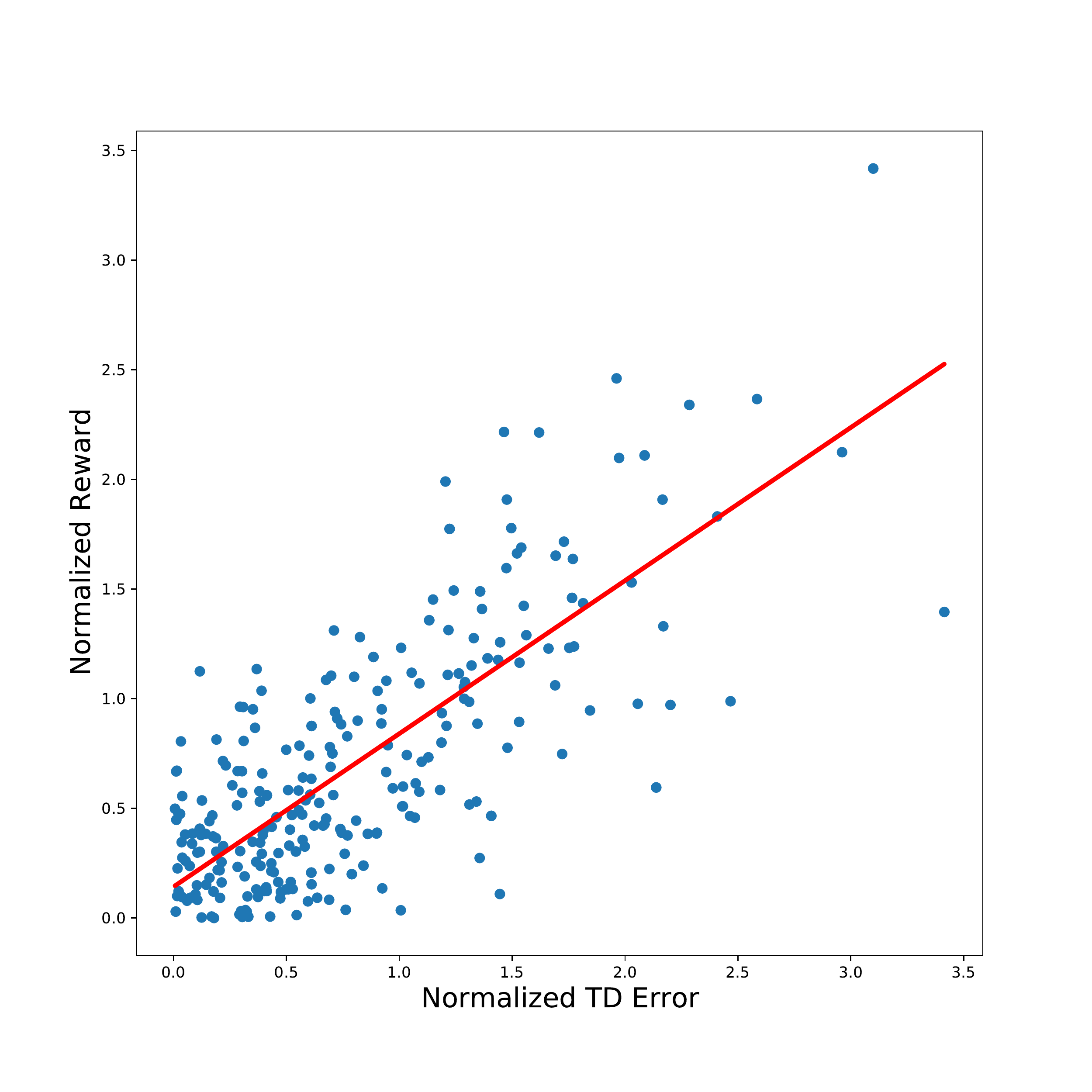}
\label{fig:uer_ant_scatterplot}}
\subfigure[RER]{%
\includegraphics[width=0.3\linewidth]{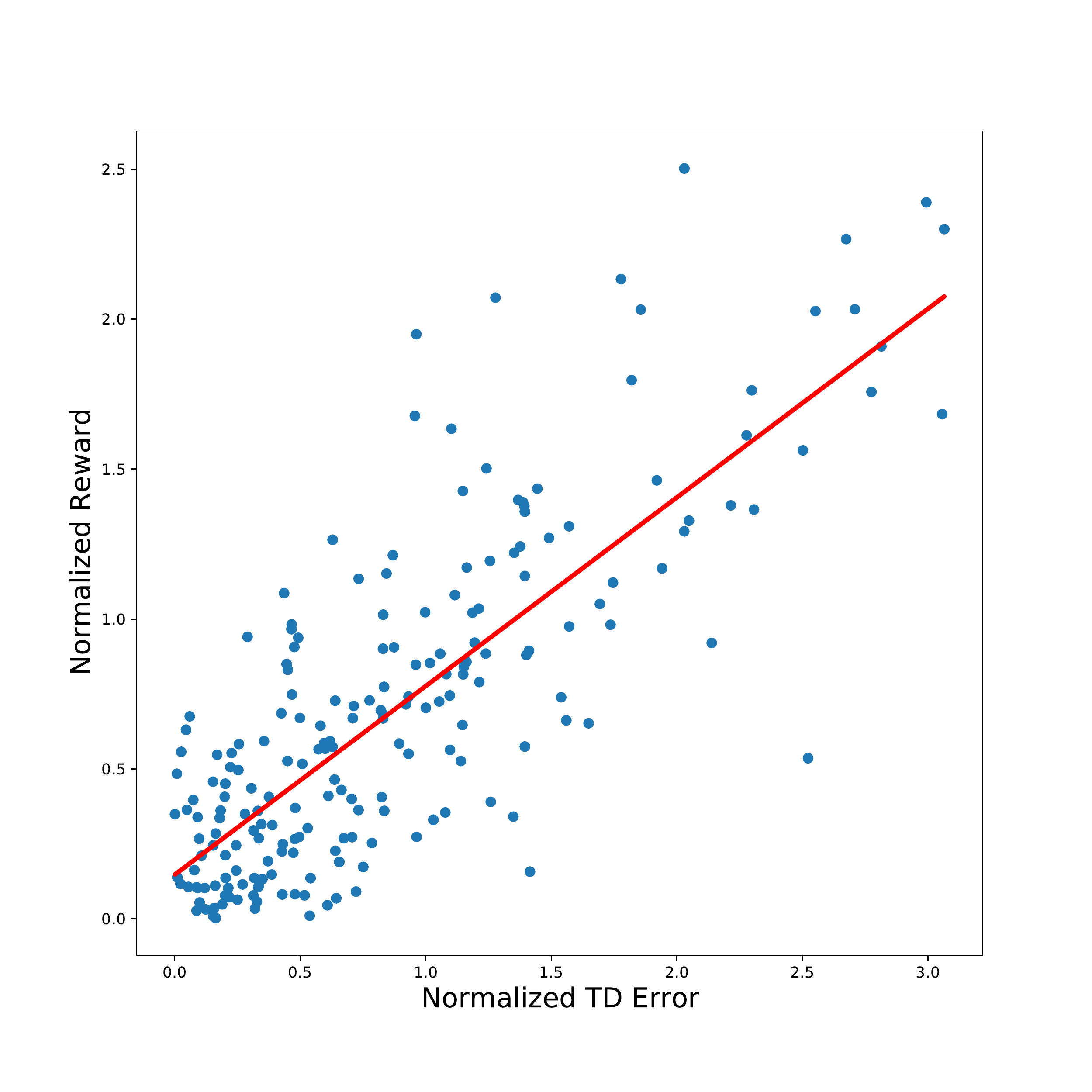}
\label{fig:rer_ant_scatterplot}}
\subfigure[IER]{%
\includegraphics[width=0.3\linewidth]{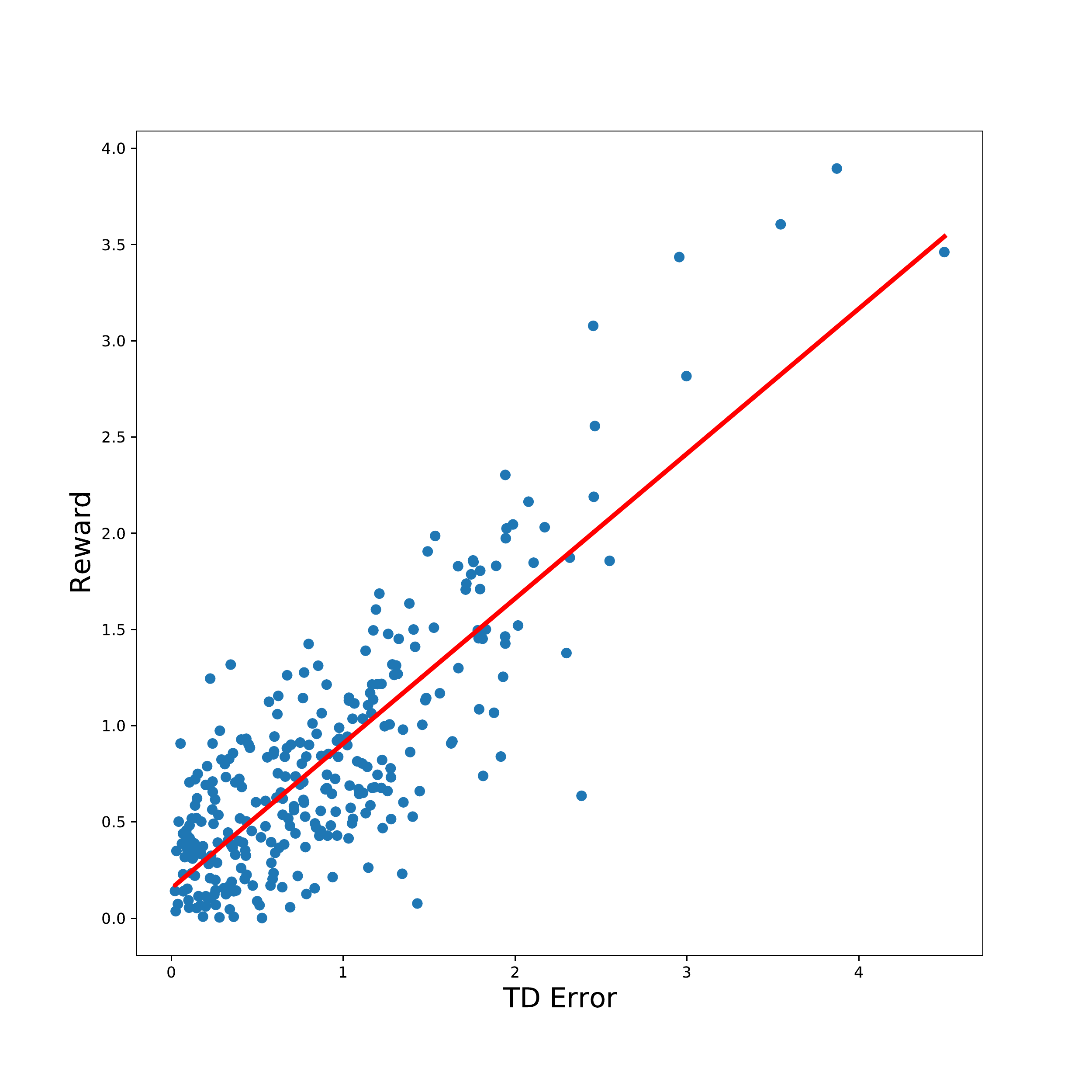}
\label{fig:rer++_ant_scatterplot}}

\caption{Relationship between absolute values of TD Error (Surprise factor) and Reward for the Ant environment.}
\label{fig:scatterplot_ant}
\end{figure*}

We show empirically that this holds in more complicated environments as well. Figure~\ref{fig:scatterplot_ant} depicts the surprise vs. reward for the Ant environment. Here we see a strong correlation between absolute reward and TD error (``Surprise factor'').

\section{Understanding Reverse Replay}
\label{understanding_rer}
There are various conceptual ways we can look at \namelo~and \namel. This section outlines some of the motivations behind using this technique. Theoretical works such as \cite{agarwal2021online,kowshik2021streaming} have established rigorous theoretical guarantees for \namelo~by utilizing super martingale structures. This structure is not present in forward replay techniques (i.e., the opposite of reverse replay) as shown in \cite{kowshik2021near}. We refer to Appendix~\ref{ablation_temporal}, where we show via an ablation study that going forward in time instead of reverse does not work very well. We give the following explanations for the success of \namel. \newline
\textbf{Propogation of Sparse Rewards:} In many RL problems, non-trivial rewards are sparse and only received at the goal states. Therefore, processing the data backward in time from such goal states helps the algorithm learn about the states that led to this non-trivial reward. Our study (see Figure \ref{fig:scatterplot_ant} and Appendix~\ref{sparse_surprise} for further details) shows that in many environments \namel~picks pivots which are the states with large (positive or negative) rewards, enabling effective learning. \newline
\textbf{More spread than \nameo:} \nameo, which greedily chooses the examples with the largest TD error to learn from, performs very poorly since it is overtly selective to the point of ignoring most of the states. To illustrate this phenomenon, we refer to the didactic example in Section~\ref{sec:toy_example}. One possible way of viewing \namel~is that \namelo~is used to reduce this affinity to picking a minimal subset of states in \nameo. \namep~is designed to achieve a similar outcome with a sophisticated and computationally expensive sampling scheme over the buffer. \newline
\textbf{Causality:} MDPs have a natural causal structure: actions and events in the past influence the events in the future. Therefore, whenever we see a surprising or unexpected event, we can understand why or how it happened by looking into the past. Further theoretical work is needed to realize \namel~better.

We further illustrate the same by referring to the straightforward didactic example (Section~\ref{sec:toy_example}), where we can see the effects of each of the experience replay methods. We also demonstrate superior performance on more complicated environments (Section~\ref{results}), showcasing the robustness of our approach with minimal hyperparameter tuning.

\section{Experimental Results}
\label{results}

In this section, we briefly discuss our experimental setup as well as the results of our experiments.
\paragraph{Environments:} We evaluate our approach on a diverse class of environments, such as (i) Environments with low-dimensional state space (including classic control and Box-2D environments), (ii) Multiple joint dynamic simulations and robotics environments (including Mujoco and Robotics environments), and (iii) Human-challenging environments (such as Atari environments). Note that previous seminal papers in the field of experience replay, such as \cite{mnih2013playing}, \cite{schaul2015prioritized}, and \cite{andrychowicz2017hindsight}, showed the efficacy of their approach on a subset of these classes. For instance, \nameu and \namep~was shown to work well on Atari games. Furthermore, \nameh~was effective in the Robotics environments such as FetchReach. In this work, we perform a more extensive study to show the robustness and effectiveness of our model not just in Atari and Robotics environments but also in Mujoco, Box2D, and Classic Control environments. Due to computational limitations and the non-reproducibility of some baselines, we could not extend our experiments to some Atari environments. We refer to Appendix~\ref{appendix-env} for a brief description of the environments used.

\paragraph{Hyperparameters:}
Refer to Appendix~\ref{appendix-hyperparam} for the exact hyperparameters used. Across all our experiments on various environments, we use a standard setting for all the different experience replay buffers. This classic setting is set so we can reproduce state-of-the-art performance using \nameu~on the respective environment. For most of our experiments, we set the uniform mixing fraction ($p$) from Algorithm~\ref{alg:rev_plus} to be 0. We use a non-zero $p$ value only for a few environments to avoid becoming overtly selective while training, as described in Appendix~\ref{appendix-hyperparam}. For \namep, we tune the $\alpha$ and $\beta$ hyperparameters used in the \cite{schaul2015prioritized} paper across all environments other than Atari. The default values of $\alpha=0.4$ and $\beta=0.6$ are robust on Atari environments as shown by extensive hyperparameter search by \cite{schaul2015prioritized}. We detail the results from our grid search in Appendix~\ref{appendix:per_grid_search}.


\paragraph{Metric:} To compare the different models, we use the \textit{Top-K seeds moving average return} as the evaluation metric across all our runs. Top-K seeds here mean we take the average of $k=3$ seeds that gave the best performance out of a total of $n=5$ trials. This measures the average performance of the top $k/n$ quantile of the trajectories. This factors in the seed sensitivity and is robust despite the highly stochastic learning curves encountered in RL, where there is a non-zero probability that non-trivial policies are not learned after many steps. It is common to use top-1 and Top-K trials to be selected from among many trials in the reinforcement learning literature (see \cite{schaul2015prioritized};\cite{sarmad2019rl};\cite{wu2017scalable};\cite{mnih2016asynchronous}). Furthermore, we motivate our design choice of using the Top-K metric as our metric of choice from our study in Appendix~\ref{appendix:metrics}. We show that the Top-K metric is more stable and suitable for comparing the performances of different RL agents, which generally suffer from highly volatile learning curves, as described above. 
Moving average with a given window size is taken for learning curves (with a window size of 20 for FetchReach and 50 for all others) to reduce the variation in return which is inherently present in each epoch. We argue that taking a moving average is essential since, usually, pure noise can be leveraged to pick a time instant where a given method performs best \citep{henderson2018deep}. Considering the Top-K seed averaging of the last step performance of the moving average of the learning curves gives our metric - the \textit{Top-K seeds moving average return}.

\paragraph{Comparison with SOTA:} Table~\ref{table-results_1} depicts our results in various environments upon using different SOTA replay sampler mechanisms (\nameu, \namep~and \nameh). Our proposed sampler outperforms all other baselines in most tasks and compares favorably in others. Our experiments on various environments across various classes, such as classic control, and Atari, amongst many others, show that our proposed methodology consistently outperforms all other baselines in most environments, as summarized in Table~\ref{tab:initial_results}. Furthermore, our proposed methodology is robust across various environments, as highlighted in Table~\ref{table-results_1}. The learning curves for our experiments have been depicted in the Appendix. (see Appendix~\ref{additional_results})

\begin{table}[hbt!]
  \centering
  \caption{\textit{Top-K seeds Moving Average Return} results across various environments. From our experiments, we note that \namel~outperforms previous SOTA baselines in most environments. Appendix~\ref{appendix-hyperparam} depicts the hyperparameters used for the experiment.}
  \vspace*{1mm}
  \label{table-results_1}
  \resizebox{\columnwidth}{!}{
  \begin{tabular}{lcccc}  
    \toprule
    \textbf{Dataset} & {\textit{\textbf{UER}}} & {\textit{\textbf{PER}}}  & {\textit{\textbf{HER}}}  & {\textit{\textbf{IER}}} \\ \toprule
    {\text{CartPole}}  & 153.14 {\tiny $\pm$ 32.82} & 198.06 {\tiny $\pm$ 3.68} & 173.84 {\tiny $\pm$ 26.11} & \textbf{199.83} {\tiny $\pm$ 0.31}\\  \midrule
    
    {\text{Acrobot}} & -257.93 {\tiny $\pm$ 184.28} & -233.45 {\tiny $\pm$ 127.99} & -389.42 {\tiny $\pm$ 113.22} & \textbf{-193.90} {\tiny $\pm$ 57.56} \\  \midrule
    
    {\text{Inverted Pendulum}} & -161.93 {\tiny $\pm$ 10.55} & -155.05 {\tiny $\pm$ 9.95} & -629.42 {\tiny $\pm$ 815.24} & \textbf{-150.27} {\tiny $\pm$ 9.63} \\  \midrule
    
    {\text{LunarLander}}  & -4.42 {\tiny $\pm$ 20.06} & 11.21 {\tiny $\pm$ 15.51} & 6.00 {\tiny $\pm$ 10.02} & \textbf{12.32} {\tiny $\pm$ 27.55}\\  \midrule
    
    {\text{HalfCheetah}}  & 10808.93 {\tiny $\pm$ 1094.32} & 99.75 {\tiny $\pm$ 1124.46} & \textbf{11072.54} {\tiny $\pm$ 297.12} & 10544.88 {\tiny $\pm$ 342.01}\\  \midrule
    
    {\text{Ant}}  & 3932.85 {\tiny $\pm$ 1024.86} & -2699.84 {\tiny $\pm$ 1.34} & 3803.17 {\tiny $\pm$ 996.81}  & \textbf{4203.21} {\tiny $\pm$ 345.22}\\  \midrule
    
    {\text{Reacher}}  & -4.97 {\tiny $\pm$ 0.31} & -5.42 {\tiny $\pm$ 0.61} & -5.30 {\tiny $\pm$ 0.50}  & \textbf{-4.92} {\tiny $\pm$ 0.27}\\  \midrule
    
    {\text{Walker}}  & 3597.03 {\tiny $\pm$ 1203.79} & 1709.48 {\tiny $\pm$ 1635.90} & 889.82 {\tiny $\pm$ 1427.92}  & \textbf{4349.29} {\tiny $\pm$ 680.35}\\  \midrule
    
    {\text{Hopper}}  & 3072.65 {\tiny $\pm$ 621.40} & 784.00 {\tiny $\pm$ 1536.00} & 2685.72 {\tiny $\pm$ 1221.81} & \textbf{3205.05} {\tiny $\pm$ 406.35} \\  \midrule
    
    {\text{Inverted Double Pendulum}} & 8489.54 {\tiny $\pm$ 927.69} & 8654.78  {\tiny $\pm$ 981.67} & 9002.05 {\tiny $\pm$ 464.20} & \textbf{9241.61} {\tiny $\pm$ 91.68}\\  \midrule
    
    {\text{Fetch-Reach}}  & -1.84 {\tiny $\pm$ 0.57} & -49.90 {\tiny $\pm$ 0.10} & -2.92 {\tiny $\pm$ 1.79} &  \textbf{-1.74} {\tiny $\pm$ 0.24}\\  \midrule
    
    {\text{Pong}} & \textbf{19.15} {\tiny $\pm$ 1.32} & 17.02 {\tiny $\pm$ 3.27} & 18.70 {\tiny $\pm$ 1.02} & 19.10 {\tiny $\pm$ 1.20}\\  \midrule
    
    {\text{Enduro}} & 227.01 {\tiny $\pm$ 319.15} & 565.23 {\tiny $\pm$ 116.36} & 514.32 {\tiny $\pm$ 132.39} & \textbf{586.32} {\tiny $\pm$ 111.44} \\
    \bottomrule
  \end{tabular}
 }
\end{table}

\paragraph{Forward vs. Reverse:} The intuitive limitation to the "looking forward" approach is that in many RL problems, the objective for the agent is to reach a final goal state, where the non-trivial reward is obtained. Since non-trivial rewards are only offered in this goal state, it is informative to look back from here to learn about the states that \emph{lead} to this. When the goals are sparse, the TD error is more likely to be large upon reaching the goal state. Our algorithm selects these as pivots, and IER (F) might select batches overflowing into the next episode. Our studies on many environments (see Figure~\ref{fig:scatterplot_ant} and Appendix~\ref{sparse_surprise}) show that the pivot points selected based on importance indeed have large (positive or negative) rewards. Our experiments depicted in Table~\ref{table-temporl-results} show that \namel~outperforms IER (F) in most environments.

\begin{table}[h!]
  \centering
  \caption{\textit{Top-K seeds Moving Average Return} results across various environments for Temporal Ablation study between IER (F) and IER (R; default). Note that we use the base setting of \namel~in this section to avoid spurious comparisons (i.e., with $p = 0$ and no hindsight).}
  \vspace*{1mm}
  \label{table-temporl-results}
  \resizebox{0.7\columnwidth}{!}{
  \begin{tabular}{lcc}
    \toprule
    \textbf{Dataset} & {\textit{\textbf{Forward}}} & {\textit{\textbf{Reverse}}} \\ \toprule
    {\text{CartPole}}  & 196.51 {\tiny $\pm$ 6.26} & \textbf{199.83} {\tiny $\pm$ 0.31}\\  \midrule
    
    {\text{Acrobot}} & -423.15 {\tiny $\pm$ 108.08} & \textbf{-313.03} {\tiny $\pm$ 190.27}\\  \midrule
    
    {\text{Inverted Pendulum}} & \textbf{-882.13} {\tiny $\pm$ 521.77} & -1111.51 {\tiny $\pm$ 559.83}\\  \midrule
    
    {\text{LunarLander}}  & -22.75 {\tiny $\pm$ 30.19} & \textbf{12.32} {\tiny $\pm$ 27.55}\\  \midrule
    
    {\text{HalfCheetah}}  & 9369.30 {\tiny $\pm$ 454.40} & \textbf{10108.15} {\tiny $\pm$ 919.27}\\  \midrule
    
    {\text{Ant}}  & 2963.71 {\tiny $\pm$ 828.50} & \textbf{4203.21} {\tiny $\pm$ 345.22}\\  \midrule
    
    {\text{Reacher}}  & -5.25 {\tiny $\pm$ 0.33} & \textbf{-4.92} {\tiny $\pm$ 0.27}\\  \midrule
    
    {\text{Walker}}  & 2213.89 {\tiny $\pm$ 1684.03} & \textbf{2830.03} {\tiny $\pm$ 881.51}\\  \midrule
    
    {\text{Hopper}}  & 393.64 {\tiny $\pm$ 181.89} & \textbf{505.58} {\tiny $\pm$ 266.52}\\  \midrule
    
    {\text{Inverted Double Pendulum}} & \textbf{9260.27} {\tiny $\pm$ 95.59} & \textbf{9241.61} {\tiny $\pm$ 91.68} \\ \midrule
    
    {\text{FetchReach}}  & -17.73 {\tiny $\pm$ 27.91} & \textbf{-2.28} {\tiny $\pm$ 1.11}\\  \midrule
    
    {\text{Pong}}  & 17.92 {\tiny $\pm$ 2.60} & \textbf{19.10} {\tiny $\pm$ 1.20}\\  \midrule

    {\text{Enduro}}  & \textbf{600.87} {\tiny $\pm$ 149.99} & 525.84 {\tiny $\pm$ 146.39} \\
    \bottomrule
  \end{tabular}
 }
\end{table}

\paragraph{Whole vs. Component Parts:} Our approach is an amalgamation of \nameo~and \namelo. Here we compare these individual parts with \namel. Table~\ref{table:ablation} describes the \textit{Top-K seeds Moving Average Return} across various environments in this domain. As demonstrated, \namel~outperforms its component parts \nameo~nor \namelo~. Furthermore, we also motivate each of our design choices such as the pivot point selection (see Appendix~\ref{pivot_ablation} where we compare our proposed approach with it's variant where the pivot points are randomly selected), and temporal structure (see Appendix~\ref{lookingback_ablation} where we compare our proposed approach with it's variant where the points are randomly sampled instead of temporally looking backward after selecting a pivot point, and also buffer batch size sensitivity (see Appendix~\ref{buffer_size_ablation}).

\begin{table}[hbt!]
  \centering
  \caption{\textit{Top-K seeds Moving Average Return} results across various environments for ablation study between \namelo, \nameo, and \namel.}
  \vspace*{1mm}
  \resizebox{0.8\columnwidth}{!}{
  \begin{tabular}{lccc}  
    \toprule
    \textbf{Dataset} &  {\textit{\textbf{RER}}}  & {\textit{\textbf{OER}}}  & {\textit{\textbf{IER}}} \\ \toprule
    {\text{CartPole}}  & 163.93 {\tiny $\pm$ 40.03} & 162.36 {\tiny $\pm$ 34.89} & \textbf{199.83} {\tiny $\pm$ 0.31}\\  \midrule
    
    {\text{Acrobot}} & -320.70 {\tiny $\pm$ 144.05} & -472.63 {\tiny $\pm$ 31.21} & \textbf{-193.90} {\tiny $\pm$ 57.56} \\  \midrule
    
    {\text{Inverted Pendulum}} & -735.38 {\tiny $\pm$ 613.5} & -166.57 {\tiny $\pm$ 17.23} & \textbf{-150.27} {\tiny $\pm$ 9.63} \\  \midrule
    
    {\text{LunarLander}}  & 9.84 {\tiny $\pm$ 13.23} & -16.08 {\tiny $\pm$ 15.38} & \textbf{12.32} {\tiny $\pm$ 27.55}
    \\\midrule
    
    {\text{HalfCheetah}}  & 9449.39 {\tiny $\pm$ 648.60} & 2237.91 {\tiny $\pm$ 2824.72} & \textbf{10544.88} {\tiny $\pm$ 342.01}\\  \midrule
    
    {\text{Ant}}  & 2168.47 {\tiny $\pm$ 415.53} & -47.93 {\tiny $\pm$ 20.47} & \textbf{4203.21} {\tiny $\pm$ 345.22}\\  \midrule
    
    {\text{Reacher}}  & -5.91 {\tiny $\pm$ 0.37} &  -5.28 {\tiny $\pm$ 0.58} & \textbf{-4.92} {\tiny $\pm$ 0.27} \\  \midrule
    
    {\text{Walker}}  & 1578.33 {\tiny $\pm$ 1313.11} & 207.51 {\tiny $\pm$ 193.06} & \textbf{4349.29} {\tiny $\pm$ 680.35} \\  \midrule
    
    {\text{Hopper}}  & 206.23 {\tiny $\pm$ 318.49} & 660.24 {\tiny $\pm$ 580.77} & \textbf{3205.05} {\tiny $\pm$ 406.35} \\  \midrule
    
    {\text{Inverted Double Pendulum}} & 8953.44 {\tiny $\pm$ 456.95} & 7724.99 {\tiny $\pm$ 1726.58} & \textbf{9241.61} {\tiny $\pm$ 91.68}\\ \midrule
    
    {\text{Fetch-Reach}}  & -49.94 {\tiny $\pm$ 0.07} & -47.72 {\tiny $\pm$ 3.33} & \textbf{-1.74} {\tiny $\pm$ 0.24}\\  \midrule
    
    {\text{Pong}} & 18.58 {\tiny $\pm$ 1.75} & 3.52 {\tiny $\pm$ 21.33} & \textbf{19.10} {\tiny $\pm$ 1.20}\\  \midrule
    
    {\text{Enduro}} & 483.98 {\tiny $\pm$ 75.45} & 361.21 {\tiny $\pm$ 86.30} & \textbf{586.32} {\tiny $\pm$ 111.44} \\
    \bottomrule
  \end{tabular}
 }
\label{table:ablation}
\end{table}



\section{Discussion}
\label{disc}
We summarize our results and discuss possible future steps.
\paragraph{Speedup:} 
\namel~shows a significant speedup in terms of time complexity over \namep~as depicted in Table~\ref{tab:speedup}. On average \namel~achieves a speedup improvement of $26.20\%$ over \namep~across a large umbrella of environment classes.
As the network becomes more extensive or complicated, our approach does have a higher overhead (especially computing TD error). Future work can investigate how to further reduce the computational complexity of our method by computing the TD error fewer times at the cost of operating with an older TD error. 
We also notice a speedup of convergence toward a local-optimal policy of our proposed approach, as shown in a few environments. Furthermore, the lack of speedup in some of the other experiments (even if they offer an overall performance improvement) could be since the "surprised" pivot cannot be successfully utilized to teach the agent rapidly in the initial stages. We refer to Appendix~\ref{additional_results} for the learning curves.

\begin{table}[hbt!]
  \centering
  \caption{Average Speedup in terms of time complexity over \namep~across various environment classes.}
    \vspace*{1mm}
    
  \resizebox{0.4\columnwidth}{!}{
  \begin{tabular}{lc}
\toprule
\textbf{Environment} & {\textit{\textbf{Average Speedup}}}\\ \midrule
\midrule

\text{Classic Control}  & 32.66\% $\color{green}{\uparrow}$ \\ \midrule
\text{Box-2D} & 54.32\% $\color{green}{\uparrow}$ \\ \midrule
\text{Mujoco} & 18.09\% $\color{red}{\downarrow}$ \\  \midrule
\text{Robotics} & 55.56\% $\color{green}{\uparrow}$ \\ \midrule
\text{Atari} & 6.53\% $\color{green}{\uparrow}$ \\ 
\bottomrule
\end{tabular}
 }
\label{tab:speedup}
\end{table}

\paragraph{Issues with stability and consistency} 

Picking pivot points by looking at the TD error might cause us to sample rare events much more often and hence cause instability compared to \nameu~as seen in some environments like HalfCheetah, LunarLander, and Ant, where there is a sudden drop in performance for some episodes (see Appendix~\ref{additional_results}). We observe that our strategy \namel~corrects itself quickly, unlike \namelo, which cannot do this (see Figure~\ref{fig:lr_hopper}). Increasing the number of pivot points per episode (the parameter $G$) and the uniform mixing probability $p$ usually mitigates this. In this work, we do not focus on studying the exact effects of $p$ and $G$ since our objective was to obtain methods that require minimal hyper-parameter tuning. However, future work can systematically investigate the significance of these parameters in various environments.



\paragraph{Why does \namel~outperform the traditional \namelo?}
The instability\footnote{i.e., sudden drop in performance without recovery like catastrophic forgetting} and unreliability of pure \namelo~with neural approximation has been noted in various works \citep{rotinov2019reverse,lee2019sample}, where \namelo~is stabilized by mixing it with \nameu~. \cite{hong2022topological} stabilizes reverse sweep by mixing it with \namep. This is an interesting phenomenon since RER is near-optimal in the tabular and linear approximation settings \citep{agarwal2021online}. Two explanations of this are i) The loss function used to train the neural network is highly non-convex, which hinders the working of \namelo~and ii) The proof given in \cite{agarwal2021online} relies extensively on `coverage' of the entire state-action space - that is, the entire state-action space is visited enough number of times - which might not hold, as shown in the toy example in Section~\ref{sec:toy_example}.


\section{Conclusion}

In conclusion, our proposed approach, Introspective Experience Replay (\namel), has shown significant promise as a solution for improving the convergence and robustness of reinforcement learning algorithms. We have demonstrated through extensive experiments that \namel~outperforms state-of-the-art techniques such as uniform experience replay (\nameu), prioritized experience replay (\namep), and hindsight experience replay (\nameh) on a wide range of tasks, and also shows a significant average speedup improvement in terms of time complexity over \namep. One of the key strengths of our proposed approach is its ability to selectively sample batches of data points prior to surprising events. This allows for the removal of bias and spurious correlations that can impede the convergence of RL algorithms. Additionally, our method is based on the theoretically rigorous reverse experience replay (\namelo) technique, which adds a further level of rigor to our approach.

Overall, our proposed approach, Introspective Experience Replay (\namel), is a promising solution that offers significant improvements over existing methods and has the potential to advance the field of reinforcement learning. We believe that our proposed approach could be widely adopted in various RL applications and have a positive impact on the field of AI.

\section*{Reproducibility Statement}

In this paper, we work with thirteen datasets, all of which are open-sourced in gym (\url{https://github.com/openai/gym}). More information about the environments is available in Appendix~\ref{appendix-env}. We predominantly use DQN, DDPG and TD3 algorithms in our research, both of which have been adapted from their open-source code. We also experimented with seven different replay buffer methodologies, all of which have been adapted from their source code\footnote{\url{https://github.com/rlworkgroup/garage}}. More details about the models and hyperparameters are described in Appendix~\ref{appendix-hyperparam}. All runs have been run using the A100-SXM4-40GB, TITAN RTX, and V100 GPUs. 
Our source code is made available for additional reference \footnote{\url{https://github.com/google-research/look-back-when-surprised}}.

\bibliography{main}
\bibliographystyle{icml2023}

\newpage
\appendix
\onecolumn

\section{Environments}
\label{appendix-env}
For all OpenAI environments, data is summarized from \url{https://github.com/openai/gym}, and more information is provided in the wiki \url{https://github.com/openai/gym/wiki}. Below we briefly describe some of the tasks we experimented on in this paper.

\subsection{CartPole-v0}

CartPole, as introduced in \cite{barto1983neuronlike}, is a task of balancing a pole on top of the cart. The cart has access to position and velocity as its state vector. Furthermore, it can go either left or right for each action. The task is over when the agent achieves 200 timesteps without a positive reward (balancing the pole) which is the goal state or has failed, either when (i) the cart goes out of boundaries ($\pm$ 2.4 units off the center), or (ii)  the pole falls over (less than $\pm$ 12 deg). The agent is given a continuous 4-dimensional space describing the environment and can respond by returning one of two values, pushing the cart either right or left.

\subsection{Acrobot-v1}

Acrobot, as introduced in \cite{sutton1995generalization}, is a task where the agent is given rewards for swinging a double-jointed pendulum up from a stationary position. The agent can actuate the second joint by one of three actions: left, right, or no torque. The agent is given a six-dimensional vector comprising the environment's angles and velocities. The episode terminates when the end of the second pole is over the base. Each timestep that the agent does not reach this state gives a -1 reward, and the episode length is 500 timesteps.

\subsection{Pendulum-v0}

The inverted pendulum swingup problem, as introduced in \cite{lillicrap2015continuous} is based on the classic problem in control theory. The system consists of a pendulum attached fixed at one end, and free at the other end.
The pendulum starts in a random position and the goal is to apply torque on the free end to swing it into an upright position, with its center of gravity right above the fixed point. The episode length is 200 timesteps, and the maximum reward possible is 0, when no torque is being applied, and the object has 0 velocity remaining at an upright configuration.

\subsection{LunarLander-v2}

The LunarLander environment introduced in \cite{brockman2016openai} is a classic rocket trajectory optimization problem. The environment has four discrete actions - do nothing, fire the left orientation engine, fire the right orientation engine, and fire the main engine. This scenario is per Pontryagin's maximum principle, as it is optimal to fire the engine at full throttle or turn it off. The landing coordinates (goal) is always at $(0,0)$. The coordinates are the first two numbers in the state vector. There are a total of 8 features in the state vector. The episode terminates if (i) the lander crashes, (ii) the lander gets outside the window, or (iii) the lander does not move nor collide with any other body.

\subsection{HalfCheetah-v2} 

HalfCheetah is an environment based on the work by \cite{wawrzynski2009cat} adapted by \cite{todorov2012mujoco}. The HalfCheetah is a 2-dimensional robot with nine links and eight joints connecting them (including two paws). The goal is to apply torque on the joints to make the cheetah run forward (right) as fast as possible, with a positive reward allocated based on the distance moved forward and a negative reward is given for moving backward. The torso and head of the cheetah are fixed, and the torque can only be applied to the other six joints over the front and back thighs (connecting to the torso), shins (connecting to the thighs), and feet (connecting to the shins). The reward obtained by the agent is calculated as follows:

\begin{equation*}
    r_t = \dot{x_t} - 0.1*\left \| a_t \right \|_2^2
\end{equation*}

\subsection{Ant-v2}

Ant is an environment based on the work by \cite{schulman2015high} and adapted by \cite{todorov2012mujoco}. The ant is a 3D robot with one torso, a free rotational body, and four legs. The task is to coordinate the four legs to move in the forward direction by applying torques on the eight hinges connecting the two links of each leg and the torso. Observations consist of positional values of different body parts of the ant, followed by the velocities of those individual parts (their derivatives), with all the positions ordered before all the velocities.
The reward obtained by the agent is calculated as follows:

\begin{equation*}
    r_t = \dot{x_t} - 0.5*\left \| a_t \right \|_2^2 - 0.0005*\left \| s_{t}^{\text{contact}}\right \|_2^2 + 1
\end{equation*}

\subsection{Reacher-v2}

The Reacher environment, as introduced in \cite{todorov2012mujoco}, is a two-jointed robot arm. The goal is to move the robot's end effector (called *fingertip*) close to a target that is spawned at a random positions. The action space is a two-dimensional vector representing the torque to be applied at the two joints.
The state space consists of angular positions (in terms of cosine and sine of the angle formed by the two moving arms), coordinates, and velocity states for different body parts followed by the distance from target for the whole object.

\subsection{Hopper-v2}

The Hopper environment, as introduced in \cite{todorov2012mujoco}, sets out to increase the number of independent state and control variables compared to classic control environments. The hopper is a two-dimensional figure with one leg that consists of four main body parts - the torso at the top, the thigh in the middle, the leg at the bottom, and a single foot on which the entire body rests. The goal of the environment is to make hops that move in the forward (right) direction by applying torques on the three hinges connecting the body parts. The action space is a three-dimensional element vector. The state space consists of positional values for different body parts followed by the velocity states of individual parts.

\subsection{Walker-v2}

The Walker environment, as builds on top of the Hopper environment introduced in \cite{todorov2012mujoco}, by adding another set of legs making it possible for the robot to walker forward instead of hop. The hopper is a two-dimensional figure with two legs that consists of four main body parts - the torso at the top, two thighs in the middle, two legs at the bottom, and two feet on which the entire body rests. The goal of the environment is to coordinate both feel and move in the forward (right) direction by applying torques on the six hinges connecting the body parts. The action space is a six-dimensional element vector. The state space consists of positional values for different body parts followed by the velocity states of individual parts.

\subsection{Inverted Double-Pendulum-v2}

Inverted Double-Pendulum as introduced in \cite{todorov2012mujoco} is built upon the CartPole environment as introduced in \cite{barto1983neuronlike}, with the infusion of Mujoco. This environment involves a cart that can be moved linearly, with a pole fixed and a second pole on the other end of the first one (leaving the second pole as the only one with one free end). The cart can be pushed either left or right. The goal is to balance the second pole on top of the first pole, which is on top of the cart, by applying continuous forces on the cart.
The agent takes a one-dimensional continuous action space in the range [-1,1], denoting the force applied to the cart and the sign depicting the direction of the force. 
The state space consists of positional values of different body parts of the pendulum system, followed by the velocities of those individual parts (their derivatives) with all the positions ordered before all the velocities. The goal is to balance the double-inverted pendulum on the cart while maximizing its height off the ground and having minimum disturbance in its velocity.

\subsection{FetchReach-v1}

The FetchReach environment introduced in \cite{plappert2018multi} was released as part of \textit{OpenAI Gym} and used the Mujoco physics engine for fast and accurate simulation. The goal is 3-dimensional and describes the desired position of the object. Rewards in this environment are sparse and binary. The agent obtains a reward of $0$ if the target location is at the target location (within a tolerance of $5$ cm) and $-1$ otherwise. Actions are four-dimensional, where 3 specifies desired gripper movement, and the last dimension controls the opening and closing of the gripper. The FetchReach aims to move the gripper to a target position.

\subsection{Pong-v0}

Pong, also introduced in \cite{mnih2013playing}, is comparatively more accessible than other Atari games such as Enduro. Pong is a two-dimensional sports game that simulates table tennis. The player controls an in-game paddle by moving vertically across the left and right sides of the screen. Players use this paddle to hit the ball back and forth. The goal is for each player to reach eleven points before the opponent, where the point is earned for each time the agent returns the ball and the opponent misses.

\subsection{Enduro-v0}

Enduro, introduced in \cite{mnih2013playing}, is a hard environment involving maneuvering a race car in the National Enduro, a long-distance endurance race. The goal of the race is to pass a certain number of cars each day. The agent must pass 200 cars on the first day and 300 cars on all subsequent days. Furthermore, as time passes, the visibility changes as well. At night in the game, the player can only see the oncoming cars' taillights. As the days' progress, cars will become more challenging to avoid. Weather and time of day are factors in how to play. During the day, the player may drive through an icy patch on the road, which would limit control of the vehicle, or a patch of fog may reduce visibility.

\section{Model and Hyperparameters}
\label{appendix-hyperparam}

In this paper, we work with two classes of algorithms: DQN, DDPG and TD3. The hyperparameters used for training our DQN algorithms in various environments are described in Table~\ref{dqn_hyperparam}. The hyperparameters used for training our DDPG algorithms in various environments are described in Table~\ref{ddpg_hyperparam}. The hyperparameters used for training DDPG are described in Table~\ref{ddpg_hyperparam}. Furthermore, the hyperparameters used for training TD3 are described in Table~\ref{td3_hyperparam}.

\begin{table}[h!]
    \centering
    \caption{Hyperparameters used for training DQN on various environments.}
    \vspace*{1mm}
    \label{dqn_hyperparam}
    \resizebox{0.8\columnwidth}{!}{
    \begin{tabular}{lcccccl}
        \toprule
        \textbf{Description}     & \textbf{CartPole} & \textbf{Acrobot} & \textbf{LunarLander} & \textbf{Pong}& \textbf{Enduro} & \texttt{argument\_name}     \\ 
        \midrule
        \multicolumn{6}{l}{\textit{General Settings}} \\
        \midrule
        Discount & $0.9$ & $0.9$ & $0.9$ & $0.99$ & $0.99$ & \texttt{discount}\\
        Batch size & $512$ & $512$ & $512$ & $32$ & $32$ & \texttt{batch\_size}\\
        Number of epochs & $100$ & $100$ & $200$ & $150$ & $800$ & \texttt{n\_epochs}\\
        Steps per epochs & $10$ & $10$ & $10$ & $20$ & $20$ & \texttt{steps\_per\_epoch}\\
        Number of train steps & $500$ & $500$ & $500$ & $125$ & $125$ & \texttt{num\_train\_steps}\\
        Target update frequency & $30$ & $30$ & $10$ & $2$ & $2$ & \texttt{target\_update\_frequency}\\
        Replay Buffer size & $1e^{6}$ & $1e^{6}$ & $1e^{6}$ & $1e^{4}$& $1e^{4}$& \texttt{buffer\_size} \\
        \midrule
        \multicolumn{6}{l}{\textit{Algorithm Settings}} \\
        \midrule
        CNN Policy Channels & - & - & - & $(32,64,64)$ & $(32,64,64)$ & \texttt{cnn\_channel}\\
        CNN Policy Kernels & - & - & - & $(8,4,3)$ & $(8,4,3)$ & \texttt{cnn\_kernel}\\
        CNN Policy Strides & - & - & - & $(4,2,1)$ & $(4,2,1)$ & \texttt{cnn\_stride}\\
        Policy hidden sizes (MLP) & $(8,5)$ & $(8,5)$ & $(8,5)$ & $(512, )$ & $(512, )$ & \texttt{pol\_hidden\_sizes} \\
        Buffer batch size & $64$ & $128$ & $128$ & $32$ & $32$ & \texttt{batch\_size} \\
        \midrule
        \multicolumn{6}{l}{\textit{Exploration Settings}} \\
        \midrule
        Max epsilon & 1.0 & 1.0 & 1.0 & 1.0 & 1.0 & \texttt{max\_epsilon} \\
        Min epsilon & 0.01 & 0.1 & 0.1 & 0.01 & 0.01 & \texttt{min\_epsilon} \\
        Decay ratio & 0.4 & 0.4 & 0.12 & 0.1 & 0.1 & \texttt{decay\_ratio} \\
        \midrule
        \multicolumn{6}{l}{\textit{Optimizer Settings}} \\
        \midrule
        Learning rate & $5e^{-5}$ & $5e^{-5}$ & $5e^{-5}$ & $1e^{-4}$ & $1e^{-4}$ & \texttt{lr} \\
        \midrule
        \multicolumn{6}{l}{\textit{PER Specific Settings}}\\
        \midrule
        Prioritization Exponent & $0.4$ & $0.6$ & $0.8$ & $0.4$ & $0.4$ & \texttt{$\alpha$} \\
        Bias Annealing Parameter & $0.6$ & $0.6$ & $0.8$ & $0.6$ & $0.6$ & \texttt{$\beta$} \\
        \midrule
        \multicolumn{6}{l}{\textit{IER Specific Settings}}\\
        \midrule
       Use Hindsight for storing states & $-$ & $\checkmark$ & $-$ & $-$ & $\checkmark$ & \texttt{use\_hindsight} \\
        Mixing Factor (p) & $0$ & $0$ & $0$ & $0$ & $0$ & \texttt{p} \\
        \bottomrule
    \end{tabular}
    }
\end{table}

\begin{table}[h!]
    \centering
    \caption{Hyperparameters used for training DDPG on Pendulum environment.}
    \vspace*{1mm}
    \label{ddpg_hyperparam}
    \resizebox{0.8\columnwidth}{!}{
    \begin{tabular}{lcl}
        \toprule
        \textbf{Description}     & \textbf{Pendulum} & \texttt{argument\_name}     \\ 
        \midrule
        \multicolumn{2}{l}{\textit{General Settings}} \\
        \midrule
        Discount & $0.95$  & \texttt{discount}\\
        Batch size & $256$ & \texttt{batch\_size}\\
        Number of epochs & $50$ & \texttt{n\_epochs}\\
        Steps per epochs & $50$ & \texttt{steps\_per\_epoch}\\
        Number of train steps & $40$ & \texttt{num\_train\_steps}\\
        Target update Tau & $0.01$ & \texttt{target\_update\_frequency}\\
        Replay Buffer size & $1e^{6}$ & \texttt{buffer\_size} \\
        \midrule
        \multicolumn{2}{l}{\textit{Algorithm Settings}} \\
        \midrule
        Policy hidden sizes (MLP) & $(400,300)$ & \texttt{pol\_hidden\_sizes} \\
        QF hidden sizes (MLP) & $(400,300)$ & \texttt{qf\_hidden\_sizes} \\
        Buffer batch size & $256$ & \texttt{batch\_size} \\
        \midrule
        \multicolumn{2}{l}{\textit{Exploration Settings}} \\
        \midrule
        Exploration Policy & Ornstein Uhlenbeck Noise & \texttt{exp\_policy} \\
        Sigma & 0.2 & \texttt{sigma} \\
        \midrule
        \multicolumn{2}{l}{\textit{Optimizer Settings}} \\
        \midrule
        Policy Learning rate & $1e^{-4}$ & \texttt{pol\_lr} \\
        QF Learning rate & $1e^{-3}$ & \texttt{qf\_lr} \\
        \midrule
        \multicolumn{2}{l}{\textit{PER Specific Settings}}\\
        \midrule
        Prioritization Exponent & $0.4$ & \texttt{$\alpha$} \\
        Bias Annealing Parameter & $0.8$ & \texttt{$\beta$} \\
        \midrule
        \multicolumn{2}{l}{\textit{IER Specific Settings}}\\
        \midrule
       Use Hindsight for storing states & $-$ & \texttt{use\_hindsight} \\
        Mixing Factor (p) & $0$ & \texttt{p} \\
        \bottomrule
    \end{tabular}
    }
\end{table}

\begin{table}[h!]
    \centering
    \caption{Hyperparameters used for training TD3 on various environments.}
    \vspace*{1mm}
    \label{td3_hyperparam}
    \resizebox{0.9\columnwidth}{!}{
    \begin{tabular}{lcccccccl}
        \toprule
        \textbf{Description}     & \textbf{Ant} & \textbf{Reacher} & \textbf{Walker}    & \textbf{Double-Pendulum} & \footnotesize{\textbf{HalfCheetah}} & \textbf{Hopper} & \textbf{FetchReach} & \texttt{argument\_name}     \\ 
        \midrule
        \multicolumn{9}{l}{\textit{General Settings}} \\
        \midrule
        Discount & $0.99$ & $0.99$ & $0.99$ & $0.99$ & $0.99$ & $0.99$ & $0.95$ & \texttt{discount}\\
        Batch size & $250$ & $250$ & $250$ & $100$ & $100$ & $100$ & $256$ & \texttt{batch\_size}\\
        Number of epochs & $500$ & $500$ & $500$ & $750$ & $500$ & $500$ & $100$ & \texttt{n\_epochs}\\
        Steps per epochs & $40$  & $40$  & $40$ & $40$ & $20$ & $40$ & $50$ & \texttt{steps\_per\_epoch}\\
        Number of train steps & $50$ & $50$ & $50$ & $1$ & $50$ & $100$ & $100$ & \texttt{num\_train\_steps}\\
        Replay Buffer size & $1e^{6}$ & $1e^{6}$ & $1e^{6}$ & $1e^{6}$ & $1e^{6}$ & $1e^{6}$ & $1e^{6}$ & \texttt{buffer\_size} \\
        \midrule
        \multicolumn{9}{l}{\textit{Algorithm Settings}} \\
        \midrule
        Policy hidden sizes (MLP) & $(256,256)$ & $(256,256)$ & $(256,256)$  & $(256,256)$ & $(256, 256)$ & $(256, 256)$ & $(256, 256)$ & \texttt{pol\_hidden\_sizes} \\
        Policy noise clip & $0.5$ & $0.5$ & $0.5$ & $0.5$ & $0.5$ & $0.5$ & $0.5$ & \texttt{pol\_noise\_clip} \\
        Policy noise & $0.2$ & $0.2$ & $0.2$ & $0.2$ & $0.2$ & $0.2$ & $0.2$ & \texttt{pol\_noise} \\
        Target update tau & $0.005$ & $0.005$ & $0.005$ & $0.005$ & $0.005$ & $0.005$ & $0.01$ & \texttt{tau} \\
        Buffer batch size & $100$ & $100$ & $100$ & $100$ & $100$ & $100$ & $256$ & \texttt{batch\_size} \\
        \midrule
       \multicolumn{9}{l}{\textit{Gaussian noise Exploration Settings}}\\
        \midrule
        Max sigma & 0.1 & 0.1 & 0.1 & 0.1 & 0.1 & 0.1 & 0.1 & \texttt{max\_sigma} \\
        Min sigma & 0.1 & 0.1 & 0.1 & 0.1 & 0.1 & 0.1 & 0.1 & \texttt{min\_sigma} \\
        \midrule
        \multicolumn{7}{l}{\textit{Optimizer Settings}}\\
        \midrule
        Policy Learning rate & $1e^{-3}$ & $1e^{-3}$ & $1e^{-4}$ & $3e^{-4}$ & $1e^{-3}$ & $3e^{-4}$ & $1e^{-3}$ & \texttt{pol\_lr} \\
        QF Learning rate & $1e^{-3}$ & $1e^{-3}$ & $1e^{-3}$ & $1e^{-3}$ & $1e^{-3}$ & $1e^{-3}$ & $1e^{-3}$ & \texttt{qf\_lr} \\
        \midrule
        \multicolumn{9}{l}{\textit{PER Specific Settings}}\\
        \midrule
        Prioritization Exponent & $0.4$ & $0.4$ & $0.4$ & $0.8$ & $0.4$ & $0.4$ & $0.4$ & \texttt{$\alpha$} \\
        Bias Annealing Parameter & $0.6$ & $0.6$ & $0.6$ & $0.6$ & $0.6$ & $0.2$ & $0.6$ & \texttt{$\beta$} \\
        \midrule
        \multicolumn{9}{l}{\textit{IER Specific Settings}}\\
        \midrule
       Use Hindsight for storing states & $-$ & $-$ & $-$ & $-$ & $-$ & $-$ & $-$ & \texttt{use\_hindsight} \\
        Mixing Factor (p) & $0$ & $0$ & $0.4$ & $0$ & $0.3$ & $0.8$ & $0$ & \texttt{p} \\
        \bottomrule
    \end{tabular}
    }
\end{table}

Additionally, we use Tabular MDPs for learning a policy in our toy example. Since the environment is fairly simpler, and has very few states, function approximation is unnecessary. For all the agents trained on GridWorld, we use a common setting as described in Table~\ref{gridworld_hyperparam}. 

\begin{table}[h!]
    \centering
    \caption{Hyperparameters used for training Tabular MDP on GridWorld-1D environment.}
    \vspace*{1mm}
    \label{gridworld_hyperparam}
    \resizebox{0.4\columnwidth}{!}{
    \begin{tabular}{lcl}
        \toprule
        \textbf{Description}     & \textbf{GridWorld} &  \texttt{argument\_name}     \\ 
        \midrule
        Discount & $0.99$ & \texttt{discount}\\
        Batch size & $1$ & \texttt{batch\_size}\\
        Number of epochs & $100$ & \texttt{n\_epochs}\\
        Replay Buffer size & $3e^{4}$ & \texttt{buffer\_size} \\
        Buffer batch size & $64$ & \texttt{batch\_size} \\
        Exploration factor & 0.3 & \texttt{max\_epsilon} \\
        Learning rate & $0.1$ & \texttt{lr} \\
        \bottomrule
    \end{tabular}
    }
\end{table}

\section{Additional Results}
\label{additional_results}

\subsection{Analysis of Fault Tolerance in Reinforcement Learning}
\label{appendix:metrics}

Selecting an appropriate metric for comparison in RL is crucial due to the high variance in results attributed to the environment stochasticity and the stochasticity in the learning process. As such, using the average performance metric with small number of seeds can give statistically different distributions in results. This has been highlighted by the works of \cite{henderson2018deep}. Reliable reinforcement learning still remains an open problem. As such, using the average over all seeds might not give us a stable comparison of the algorithms considered, unless we can increase our number of seeds to large numbers such as > 20. This is computationally very expensive and such high number of trials are seldom used in practice. Instead, other approaches use the maximum of N runs to report performance. This however is not ideal since it overestimates your performance by a large amount. We can see this with the PER hyper-parameter search where we deploy the best performing hyper-parameter from the hyper-parameter grid search into an independent experiment with Top-K out of n metric. We notice that this under-performs the grid search result by a large margin. Thus, the top-1 metric is not desirable. 

Instead, we use the Top-K performance metric which we show below to be robust to noise and is closer to providing the right analysis in comparison to average metric approach. Similar to work such as \cite{mnih2016asynchronous} which used a top-5/50 experiments, we use a top-3/5 experiments to limit the overestimation of results further.



To explain our decision choice of Top-K metric, we consider a fault-based model for RL algorithm output and show via a simulation of this in a simple toy example why Top-K seeds can give us a better way of inferring the best algorithm than the average, especially when the number of independent trials is low. RL algorithms usually perform well sometimes, and some other times they fail badly. This has been noted in the literature \cite{henderson2018deep} and can also be seen in our experiments. Since the deviations are so large between seeds, this can be modeled as a `fault' rather than a mean + gaussian noise. To give a better understanding, let us elaborate with a simplified scenario: Consider the following setting where we have 20 different environments. There are two algorithms: A and B. Algorithm A gives a moving average return of 0.9 50\% of the time and a moving average return of 0 the remaining times (on all 20 environments). Algorithm B, on the other hand, gives a moving average return of 1 50\% of the time and a moving average return of 0 the remaining times (on all 20 environments). In reality, Algorithm B performs better than Algorithm A in all 20 environments. We can conclude this with the empirical average if we extend the number of experiments to very large numbers, such as 50 seeds per environment or larger. We further extend our above analysis by adding another Algorithm C that gives a moving average return of 0.8 50\% of the time and a moving average return of 0 the remaining times (on all 20 environments). 

\begin{figure}[hbt!]
\centering
\subfigure[Without Gaussian Noise]{%
\includegraphics[trim=0 0 0 20,clip,width=0.45\linewidth]{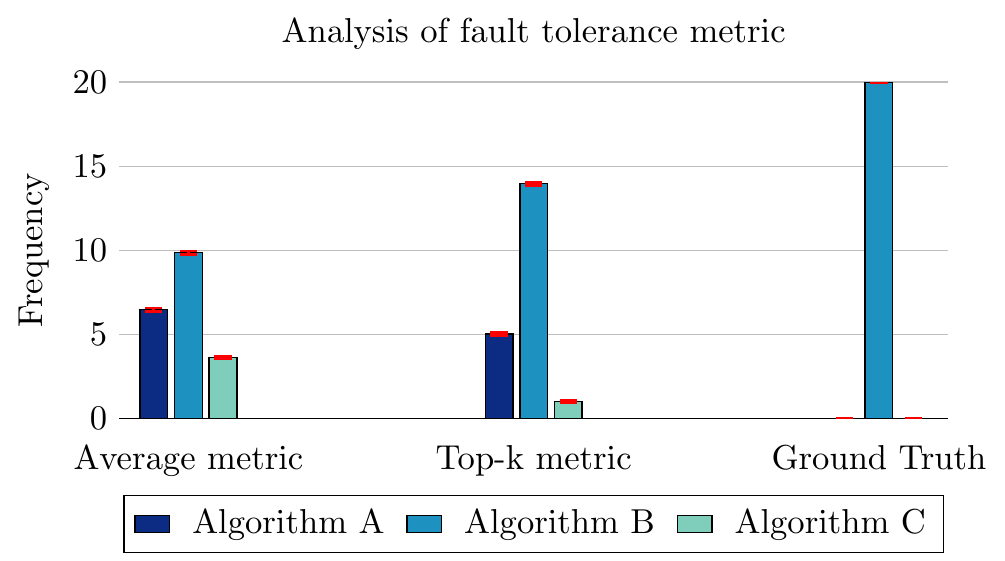}
\label{fig:just_fault}}
\quad
\subfigure[With Gaussian Noise ($\sigma = 0.2$)]{%
\includegraphics[trim=0 0 0 20,clip,width=0.45\linewidth]{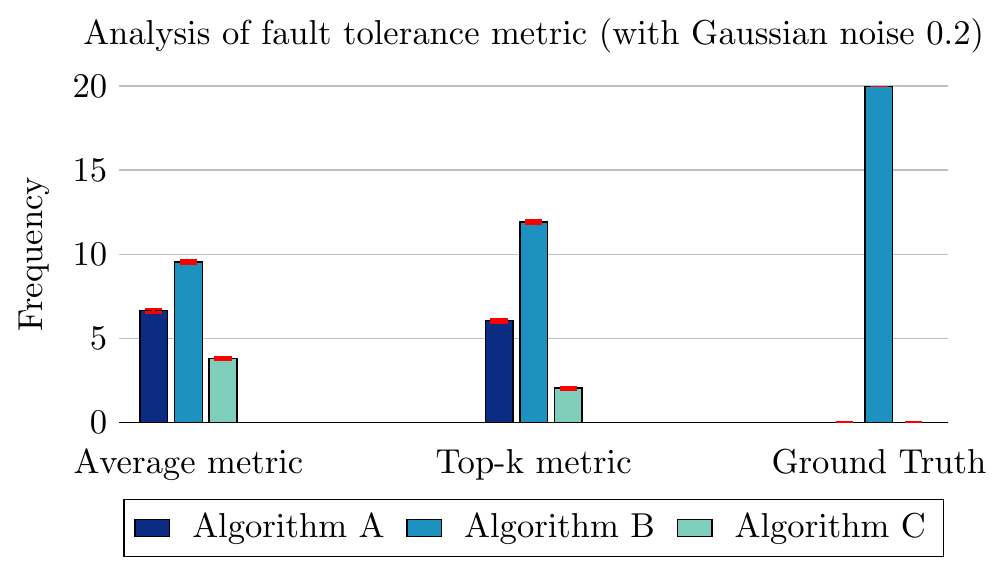}
\label{fig:with_fault_0.2}}
\caption{Analysis of Fault Tolerance Metrics.}
\label{fig:metrics}
\end{figure}

In this simplified model, we test how well the Top-K metric and the average metric perform in recovering the ground truth (i.e, Algorithm B) via Monte-Carlo simulation with 500 trials. In each trial, we generate the return of each algorithm over each of the 20 environments from the model described above with 10 different random seeds. For each environment, we check the best algorithm with the average metric and the best algorithm with the Top-K metric. We compare this with the ground truth (i.e, algorithm B being the best in all 20 environments). Figure~\ref{fig:just_fault} depicts the comparison between the average metric, and Top-K metric with respect to the ground truth. As illustrated, the Top-K metric is more robust to faults and is closer to the ground truth than the other metrics. We further add gaussian noise of mean 0, and standard deviation 0.2, keeping all other parameters constant. This noise is added to fault runs and standard ones. We note little difference in our results and depict the results averaged over 500 runs as depicted in Figure~\ref{fig:with_fault_0.2}
Therefore, our model suggests that the Top-K metric is more robust to faulty runs and can help facilitate the comparison of various learning algorithms with high volatility. 

\subsection{Grid-Search for tuning \namep~hyperparameters}
\label{appendix:per_grid_search}
In this section we present the results from our grid search experiment for tuning the hyperparameters for \namep. We tune  the bias annealing parameter ($\beta$), and the prioritization exponent ($\alpha$). We perform a robust grid search across all environments we have experimented with with the range of the priortization exponent and beta as $[0.2, 0.4, 0.6, 0.8, 1.0]$. Figure~\ref{per_grid_search} illustrates the performance of \namep~with varying hyperparameter

We briefly summarize the findings from our grid-search experiment below:
\begin{itemize}
    \item On environments such as CartPole , Reacher, we notice only a marginal differences between default hyper parameters ($\alpha=0.4$, $\beta=0.6$) and the best hyper parameters tuned for the model.
    \item For other environments such as HalfCheetah, Ant, FetchReach, and Walker, we notice a significant gap between the performance of \namep~and \namel~even after a thorough tuning of \namep~hyperparameters. For instance, \namel outperforms \namep~by more that 8000 in terms of average return on HalfCheetah environment. Furthemore, \namel~ outperforms \namep~by almost 48 in terms of average return on FetchReach. Finally, \namel~outperforms \namep~by more than 6200 in terms of average return on Ant. On Walker, \namel~outperforms \namep~by almost 1500 in terms of average return. For these environments, we report the results with the default setting, i.e. $\alpha=0.4$ and $\beta=0.6$.
    \item Some environments such as Acrobot, Pendulum, LunarLander, Hopper, and Double-Inverted-Pendulum did show significant improvements when tuned for the prioritization exponent, and bias annealing factor. We take the best hyperparameters from the grid-search experiment, and re-run the code to compute Top-K metrics (i.e, we pick top 3 out of 5 runs). However, from our experiments, we show that even the select hyperparameter is not robust across seeds, and overall performs worse than our proposed approach of \namel~across all environments: Acrobot, Pendulum, LunarLander, Hopper, and Double-Inverted-Pendulum.
    \item For Atari environments such as Pong and Enduro, we use the default parameters recommended by \cite{schaul2015prioritized}, and do not perform a grid-search experiment.
\end{itemize}

\begin{figure}[hbt!]
\centering
\subfigure[CartPole-v0]{%
\includegraphics[width=0.30\linewidth]{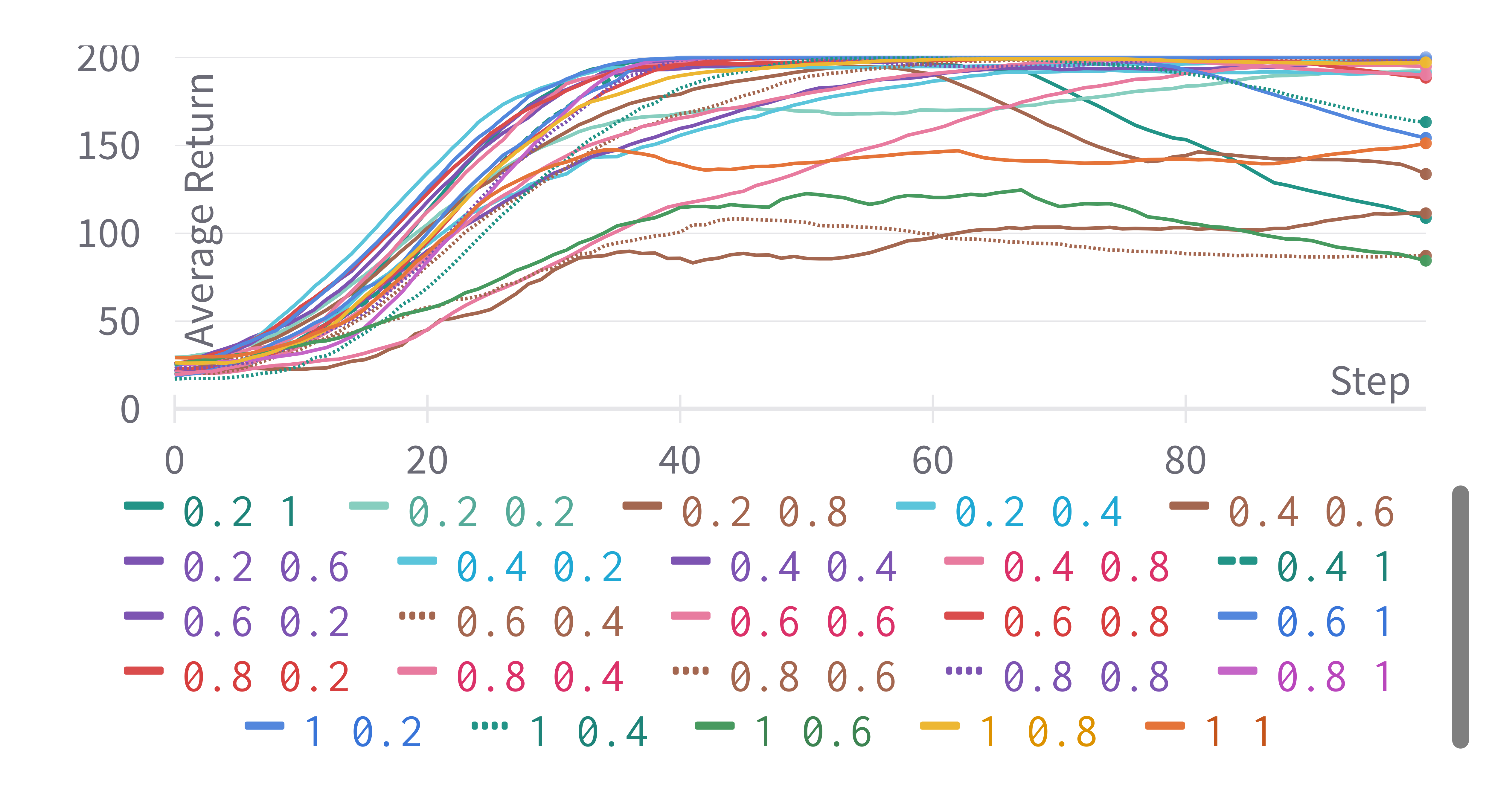}
}
\quad
\subfigure[Acrobot-v1]{%
\includegraphics[width=0.30\linewidth]{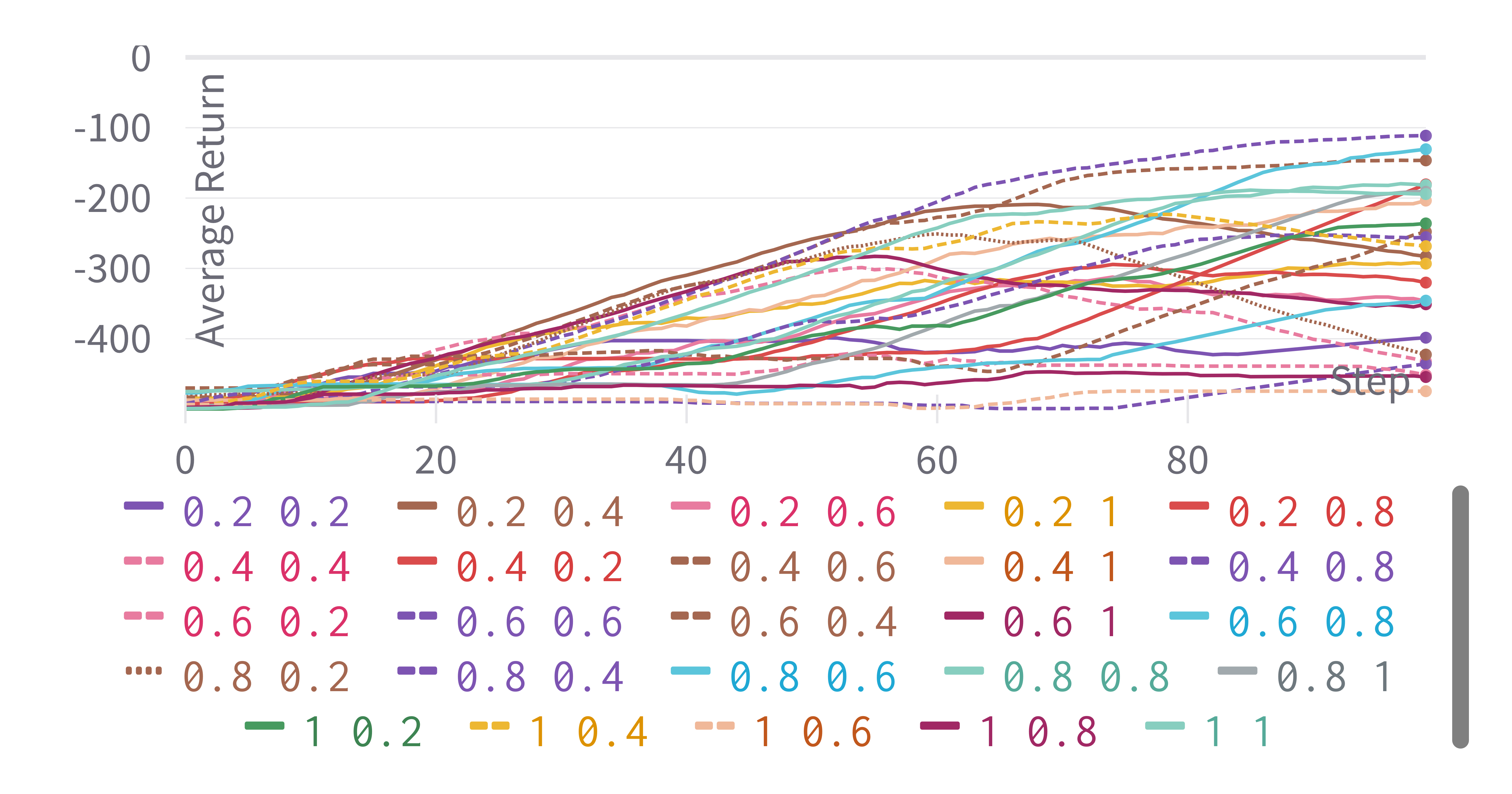}
}
\quad
\subfigure[Pendulum-v0]{%
\includegraphics[width=0.30\linewidth]{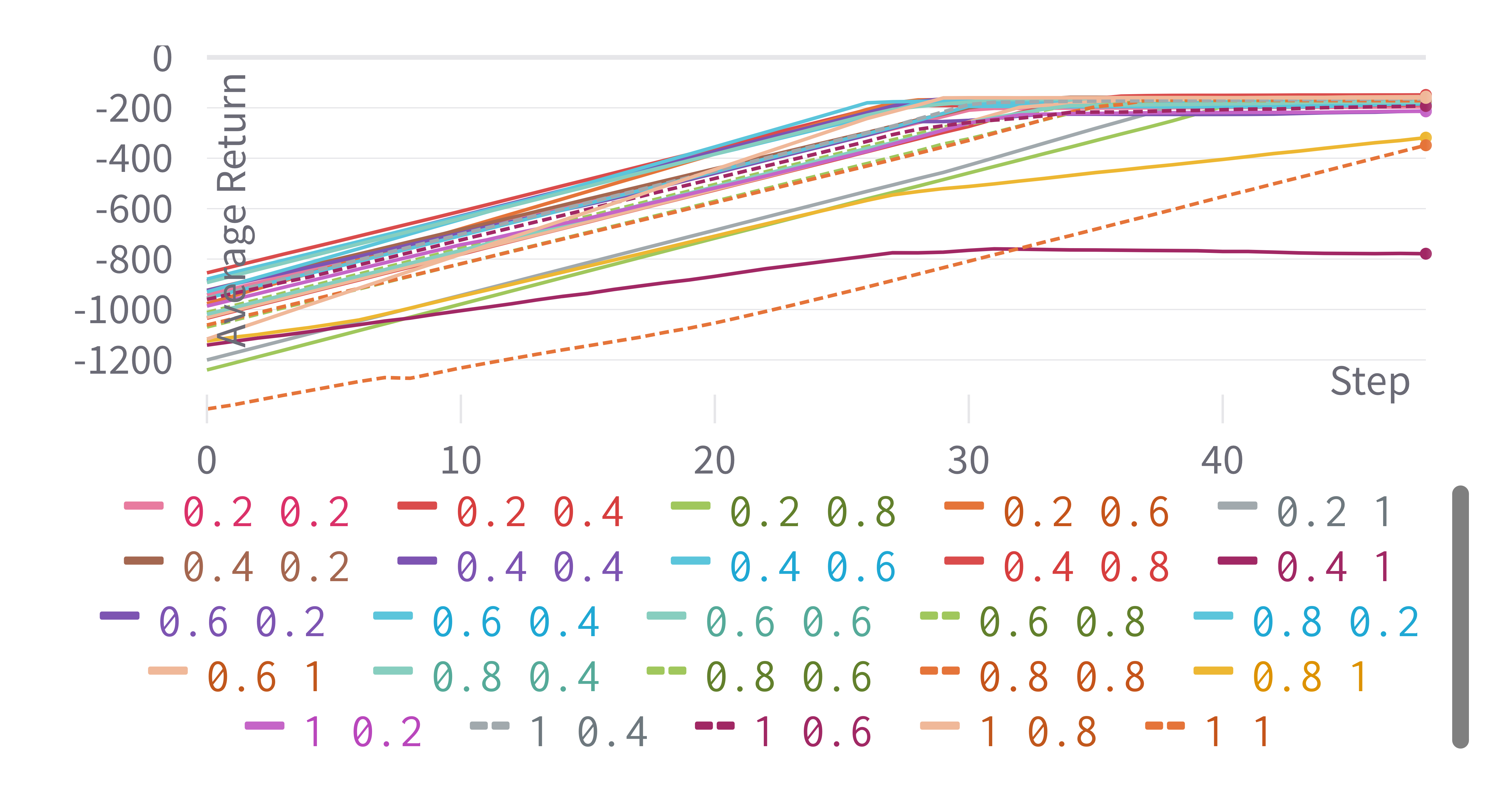}
}

\subfigure[LunarLander-v2]{%
\includegraphics[width=0.30\linewidth]{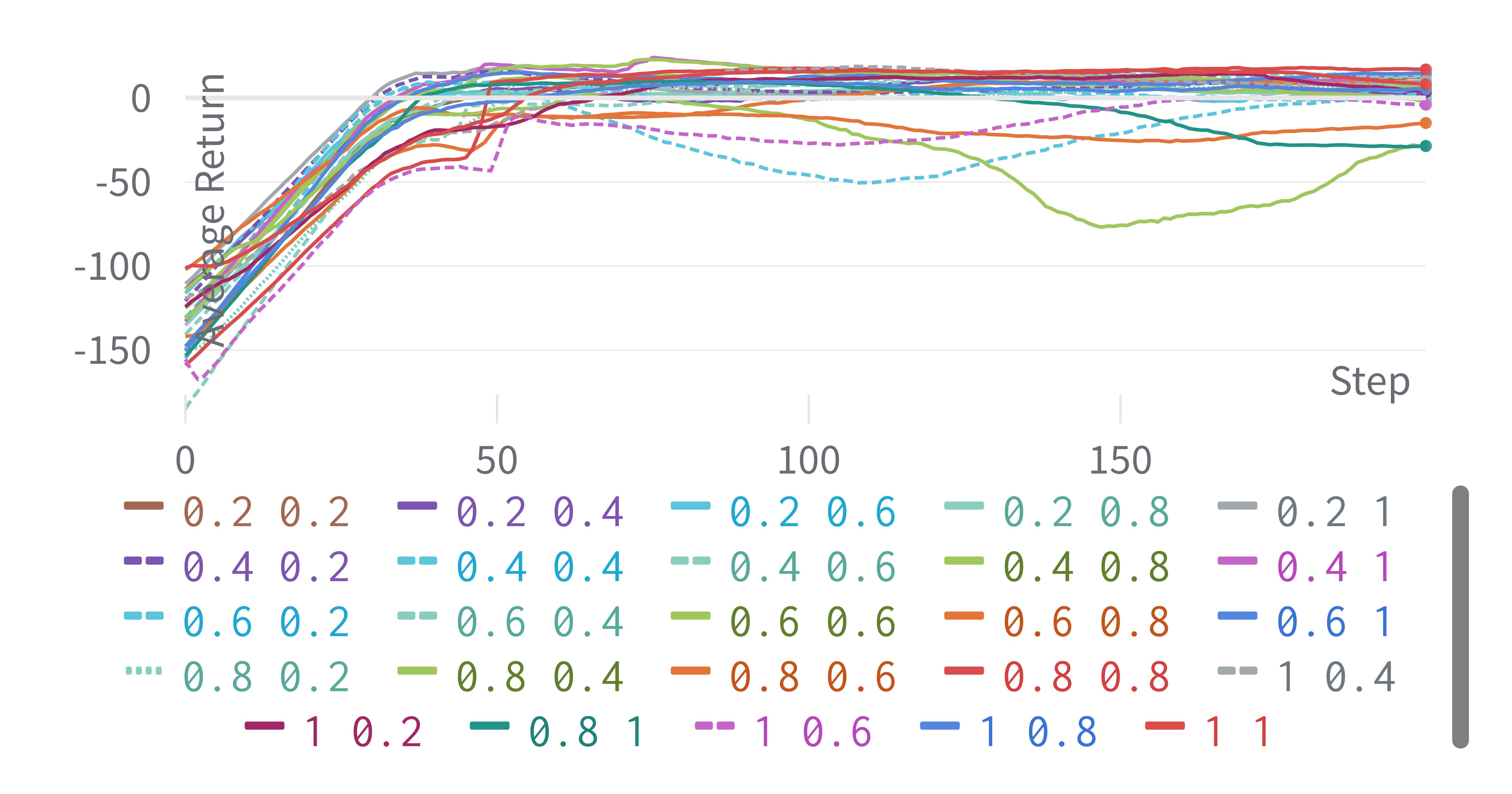}
}
\quad
\subfigure[HalfCheetah-v2]{%
\includegraphics[width=0.30\linewidth]{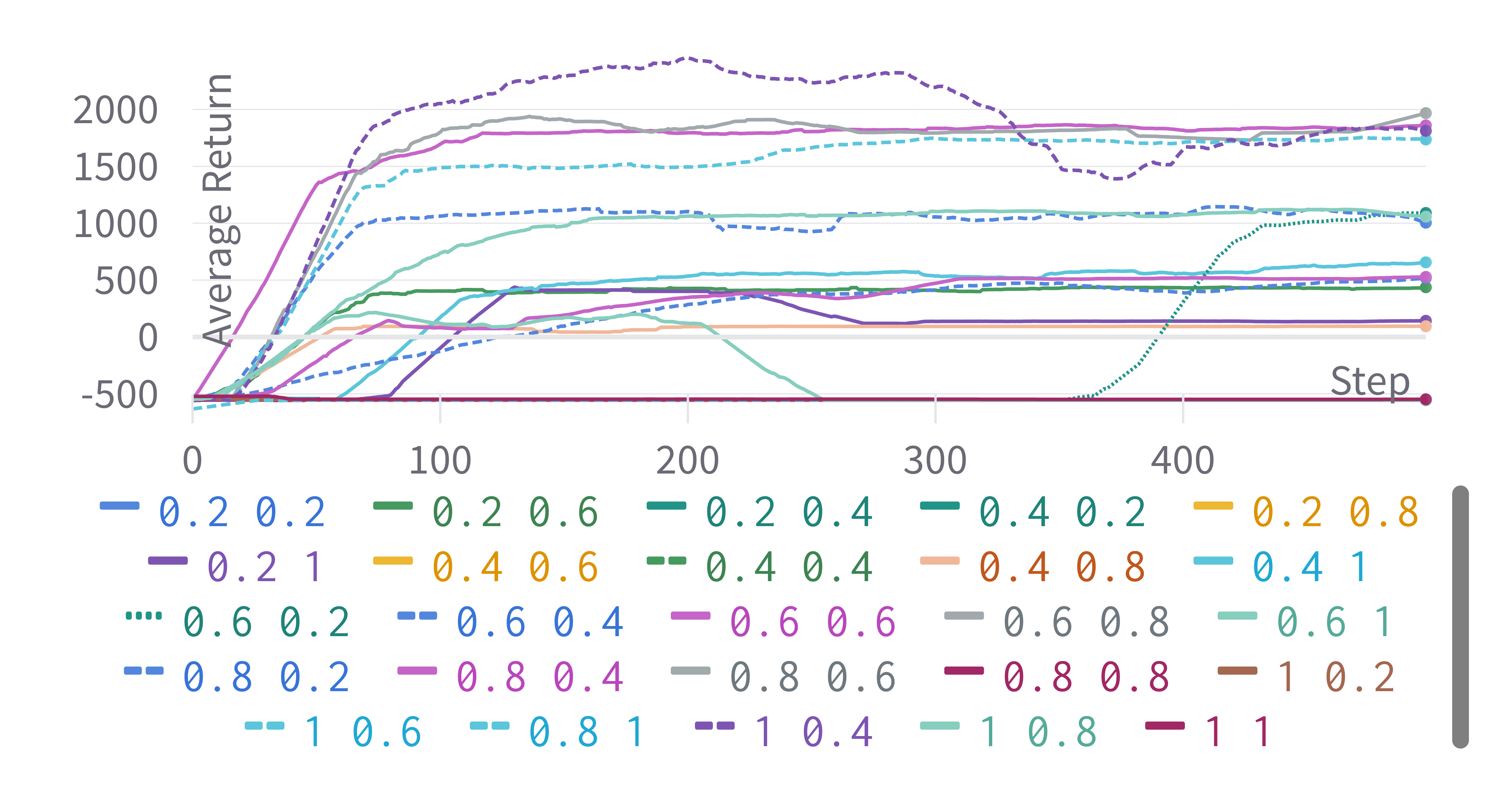}
}
\quad
\subfigure[Ant-v2]{%
\includegraphics[width=0.30\linewidth]{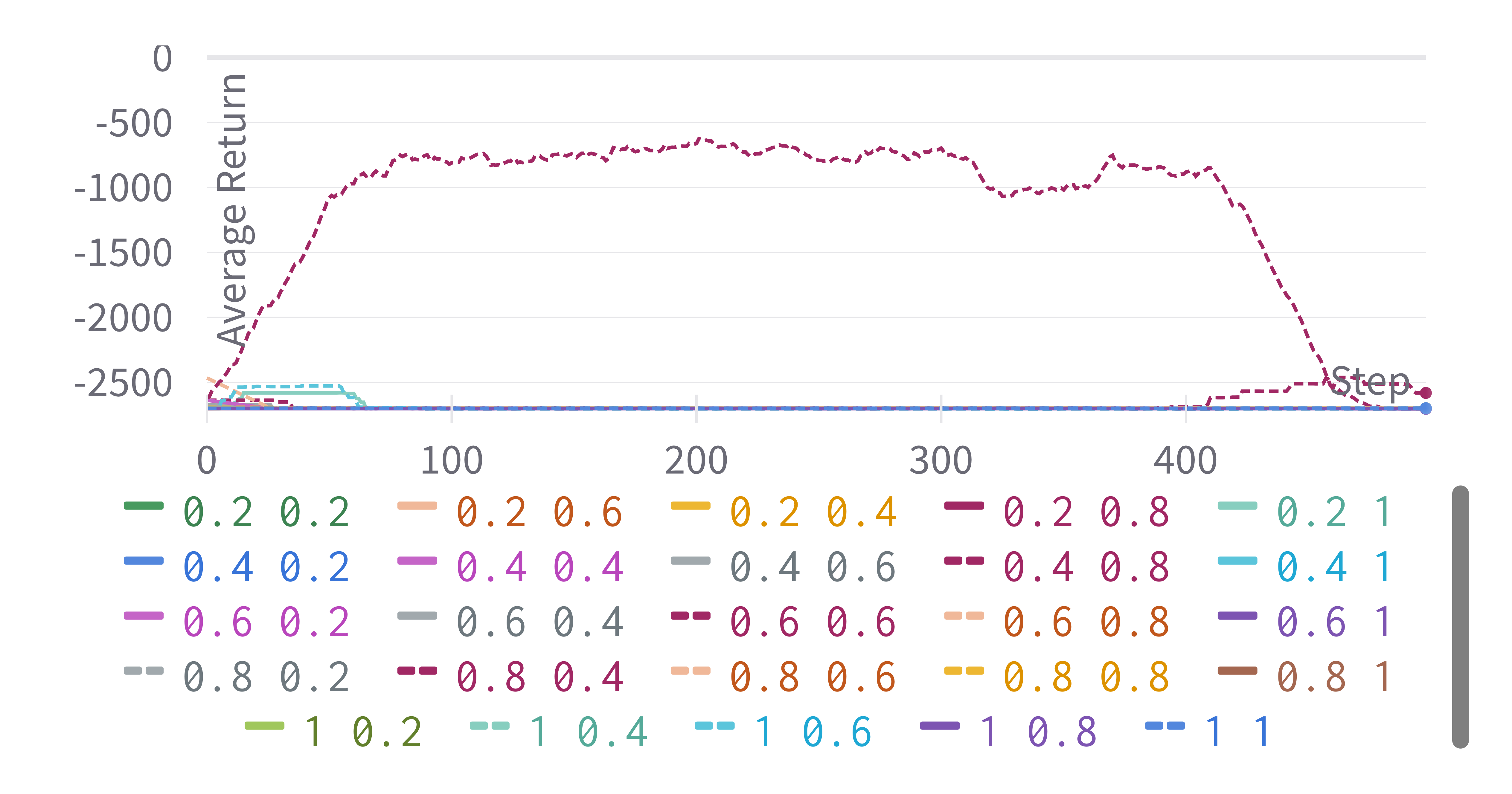}
}

\subfigure[Reacher-v2]{%
\includegraphics[width=0.30\linewidth]{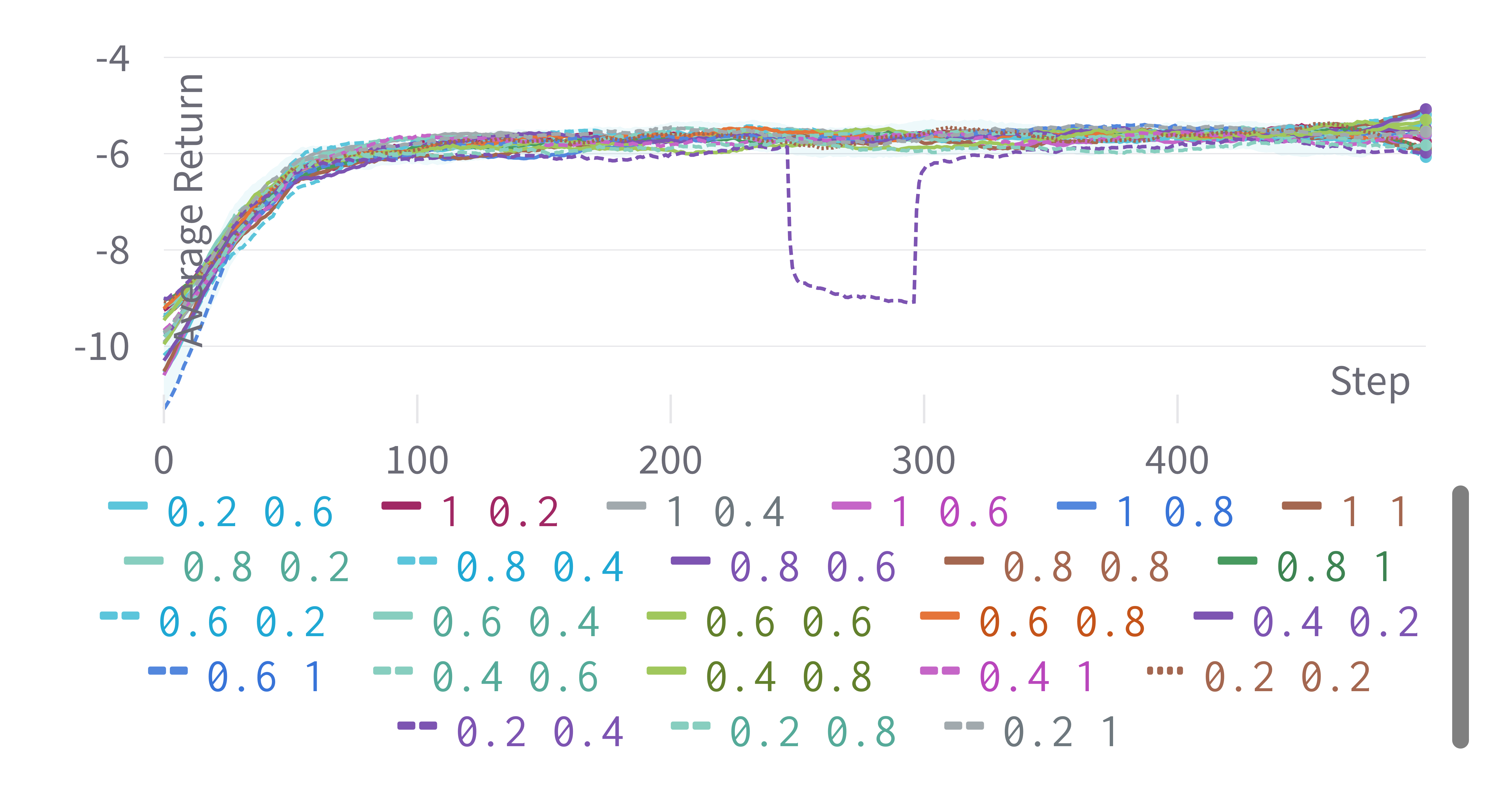}
}
\quad
\subfigure[Walker-v2]{%
\includegraphics[width=0.30\linewidth]{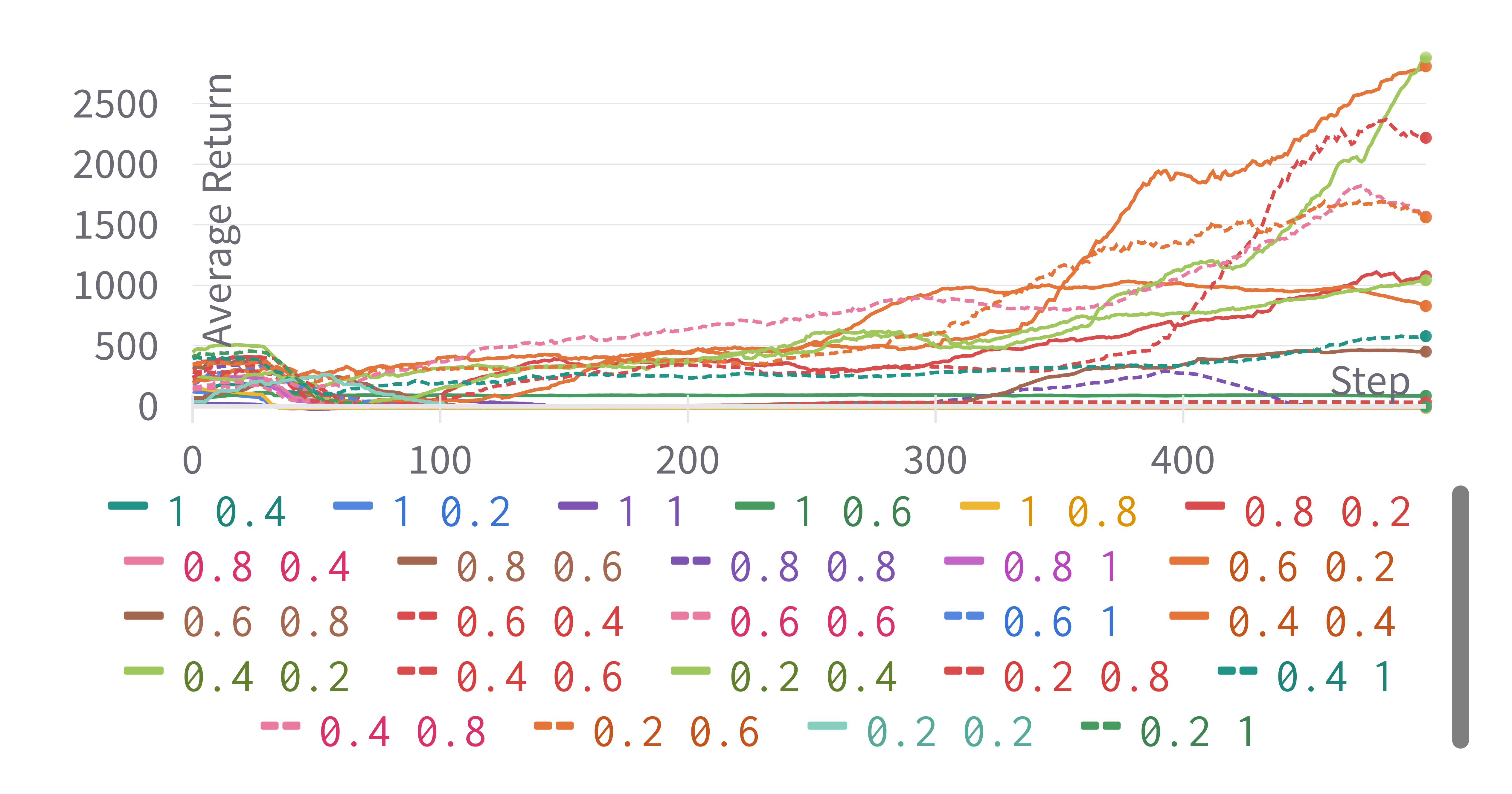}
}
\quad
\subfigure[Hopper-v2]{%
\includegraphics[width=0.30\linewidth]{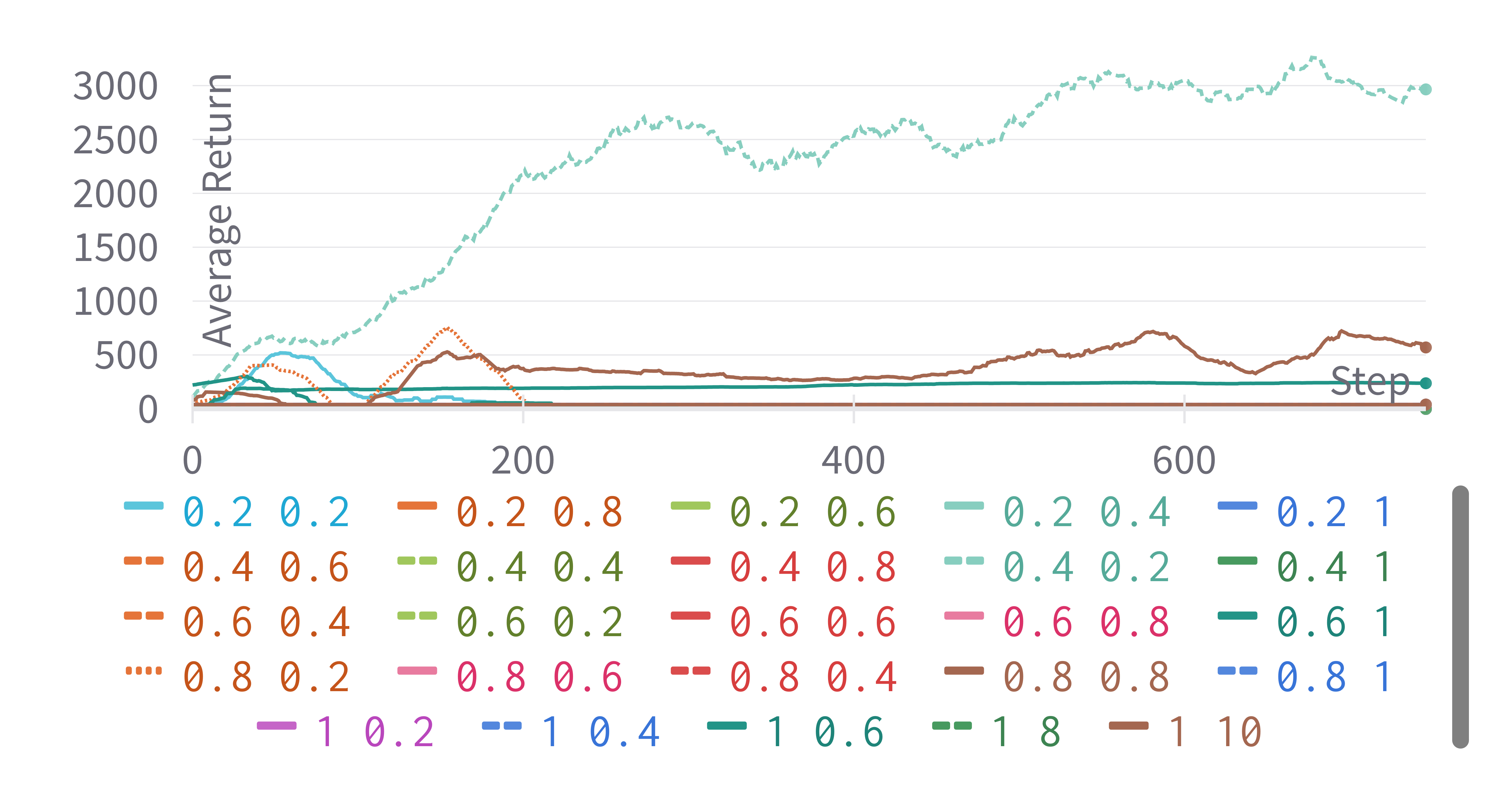}
}

\subfigure[InvertedDoublePendulum-v2]{%
\includegraphics[width=0.30\linewidth]{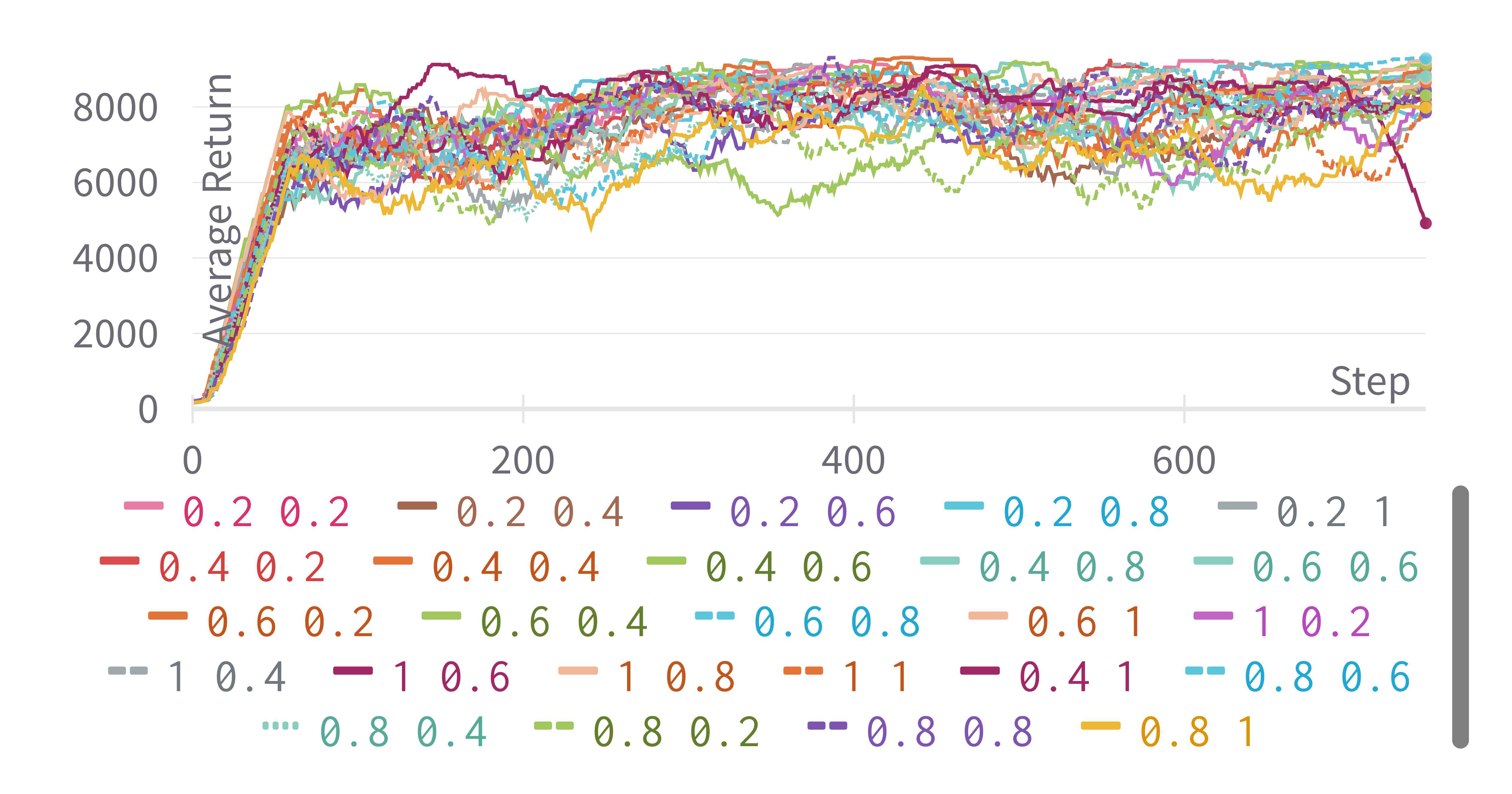}
}
\quad
\subfigure[FetchReach-v1]{%
\includegraphics[width=0.30\linewidth]{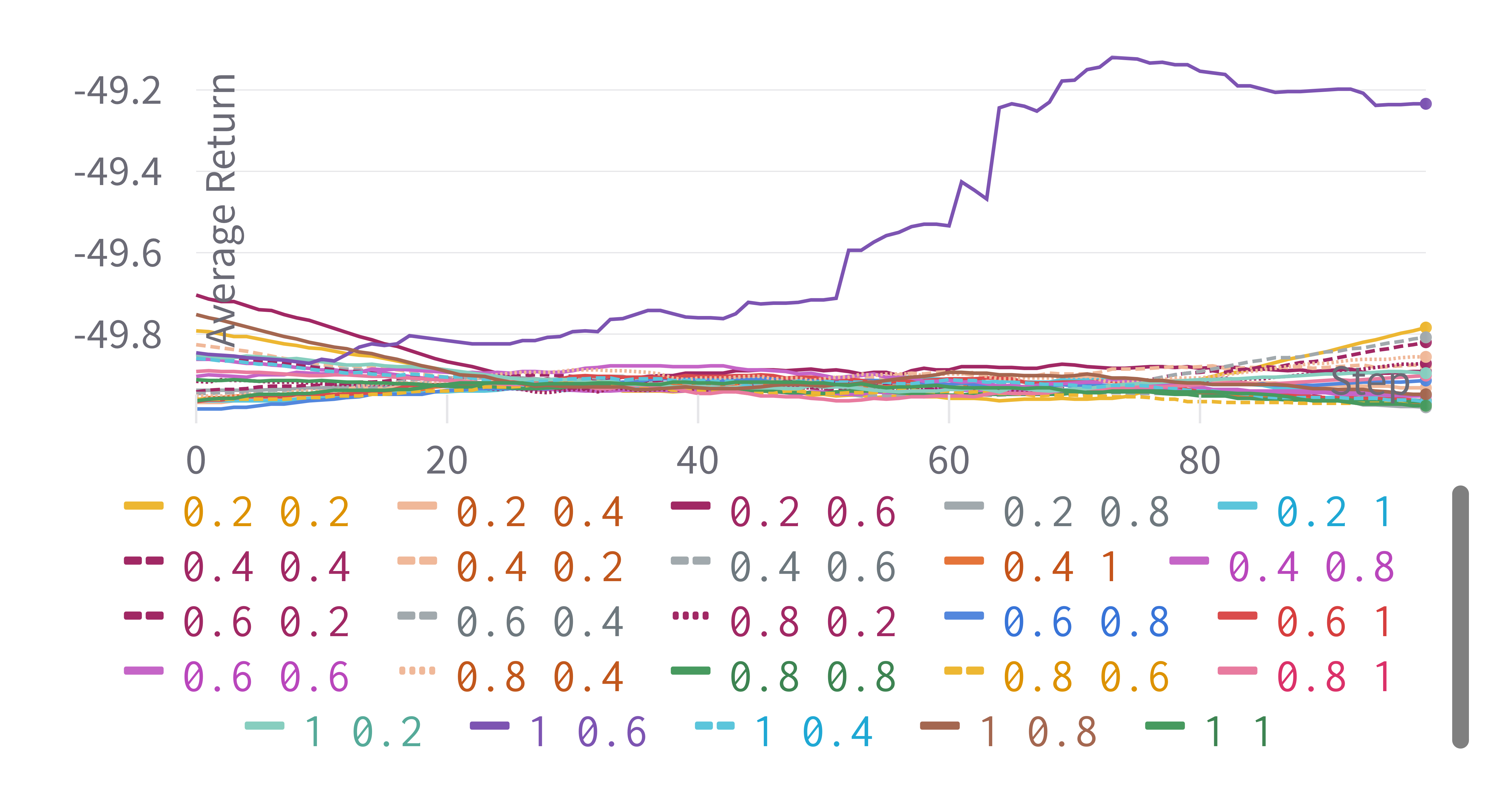}
}
\caption{Grid-Search of Prioritization exponent ($\alpha$), and Bias Annealing parameter ($\beta$) respectively.}
\label{fig:ablation_study}
\end{figure}

\subsection{Performance of \namel~with Low-Dimensional State Space}

\begin{figure}[hbt!]
\centering
\subfigure[CartPole-v0]{%
\includegraphics[width=0.45\linewidth]{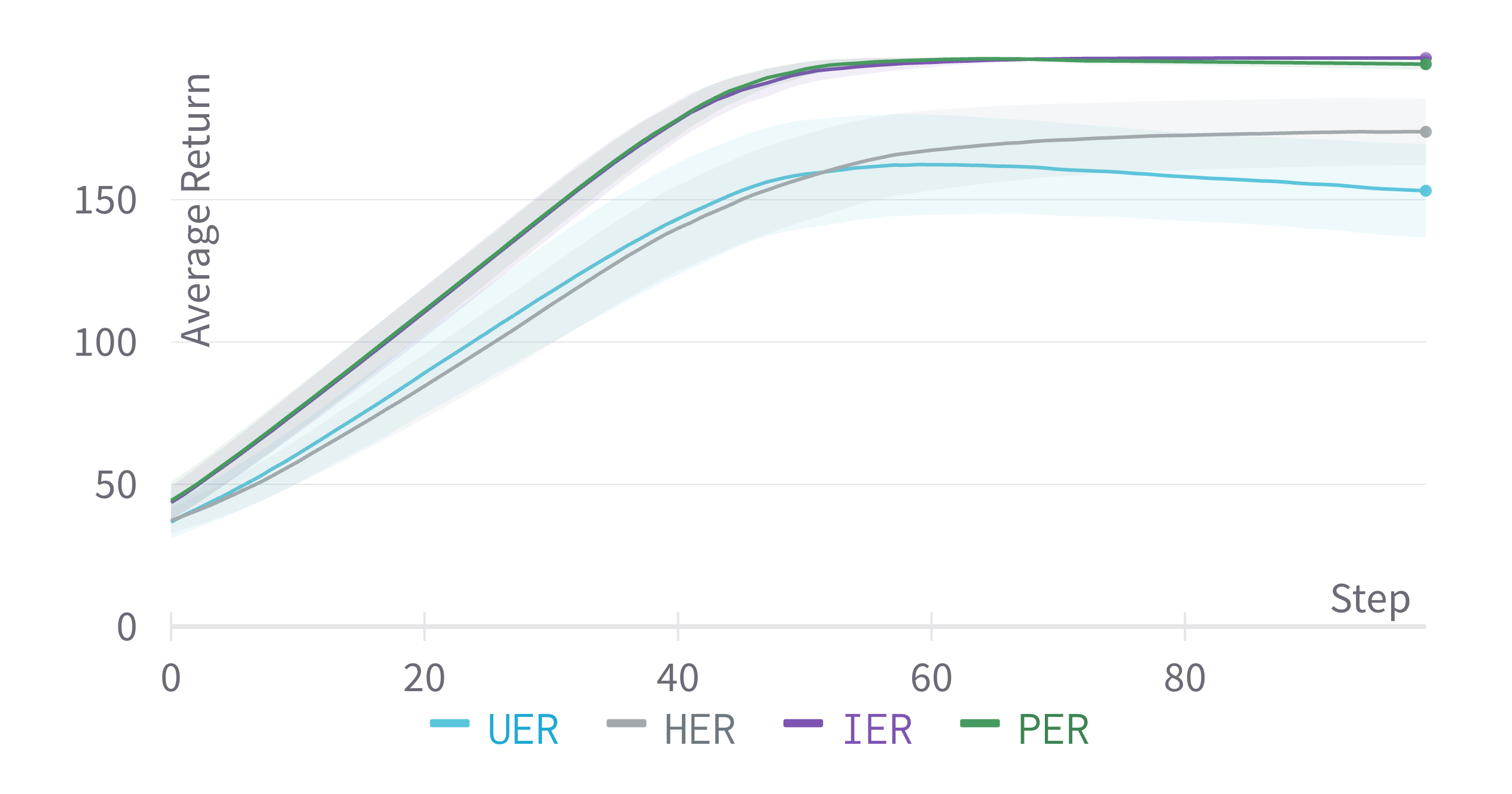}
\label{fig:lr_cartpole}}
\quad
\subfigure[Acrobot-v1]{%
\includegraphics[width=0.45\linewidth]{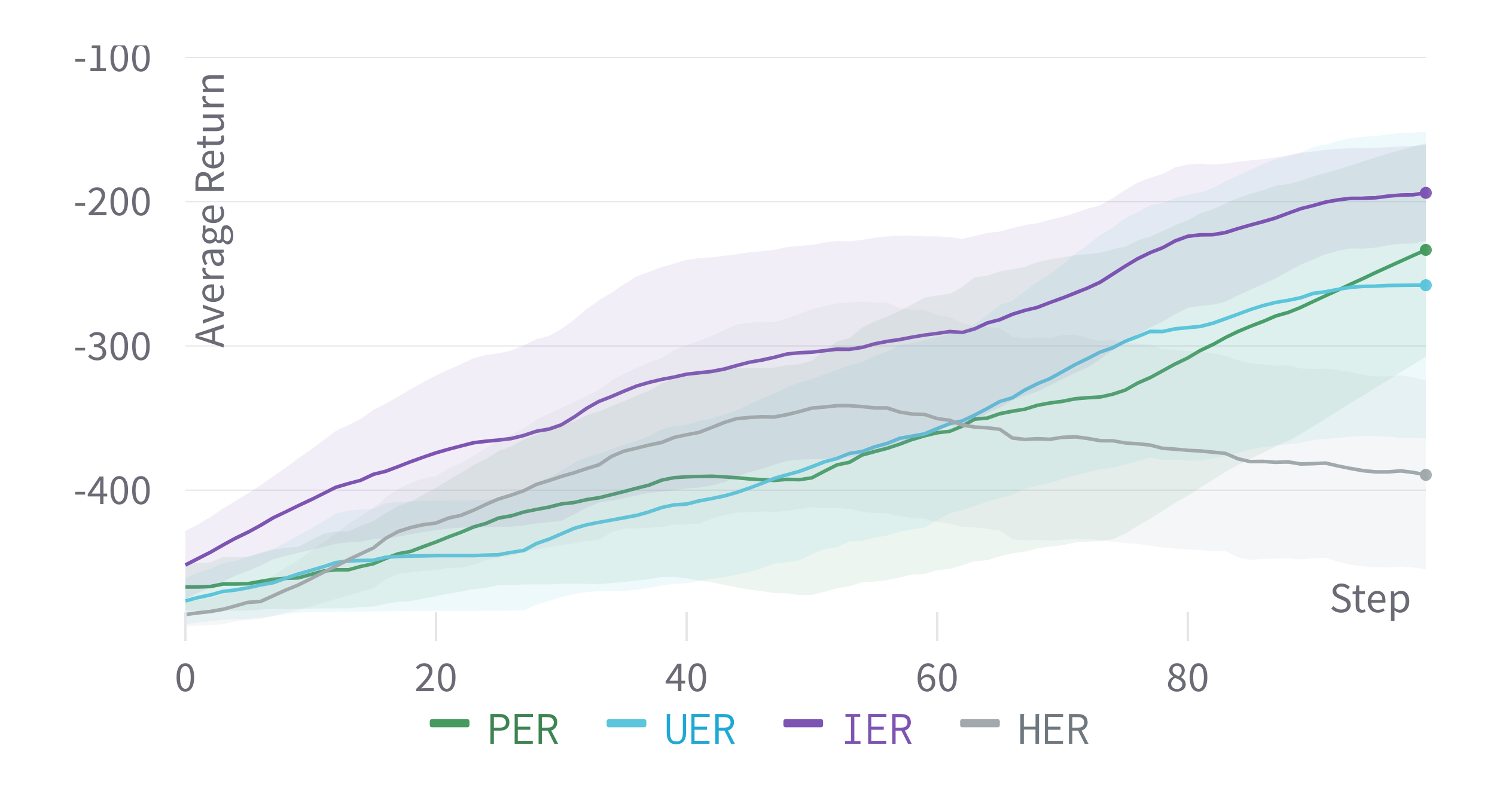}
\label{fig:lr_acrobot}}

\subfigure[Pendulum-v0]{%
\includegraphics[width=0.45\linewidth]{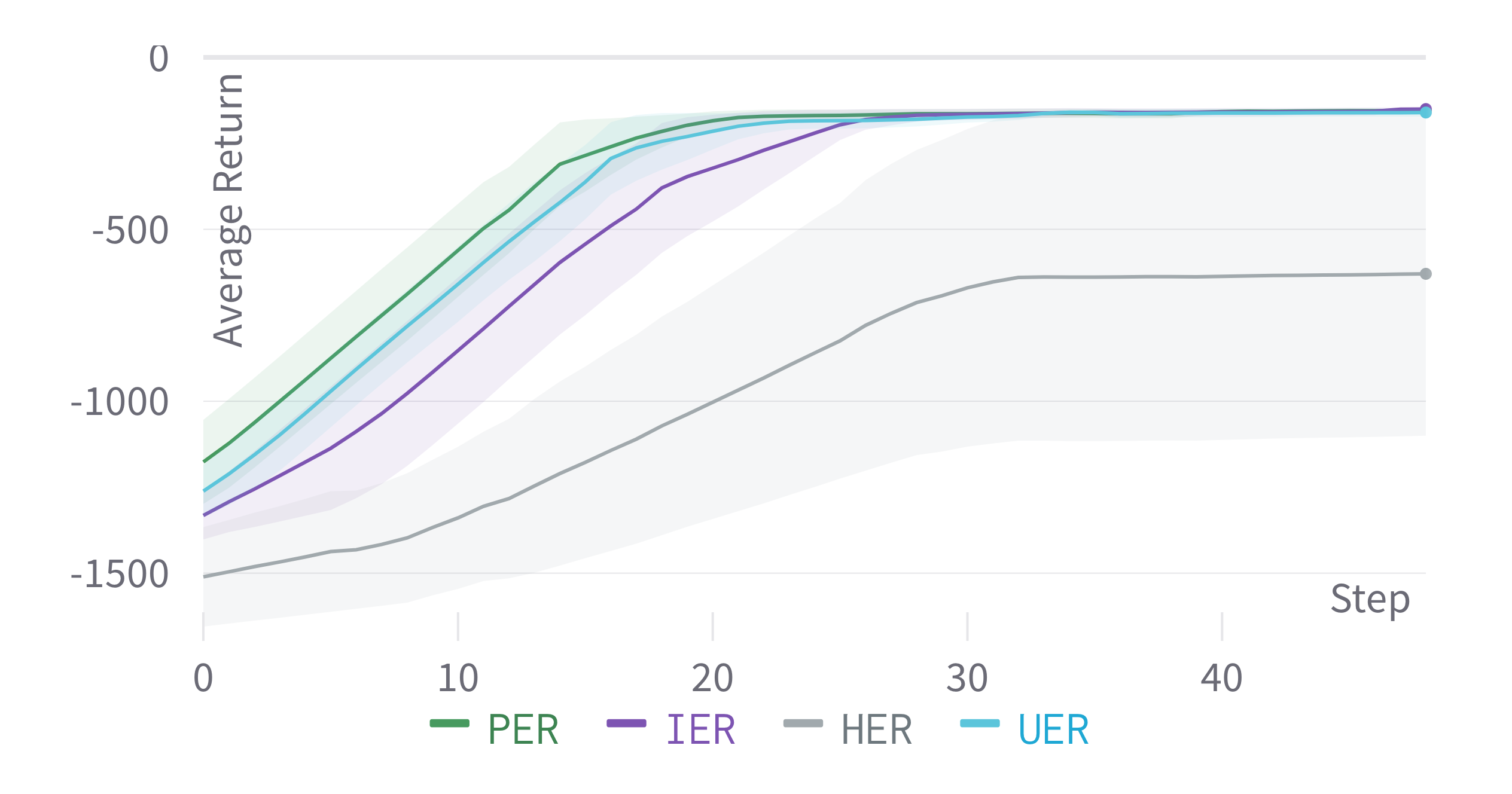}
\label{fig:lr_pendulum}}
\quad
\subfigure[LunarLander-v2]{%
\includegraphics[width=0.45\linewidth]{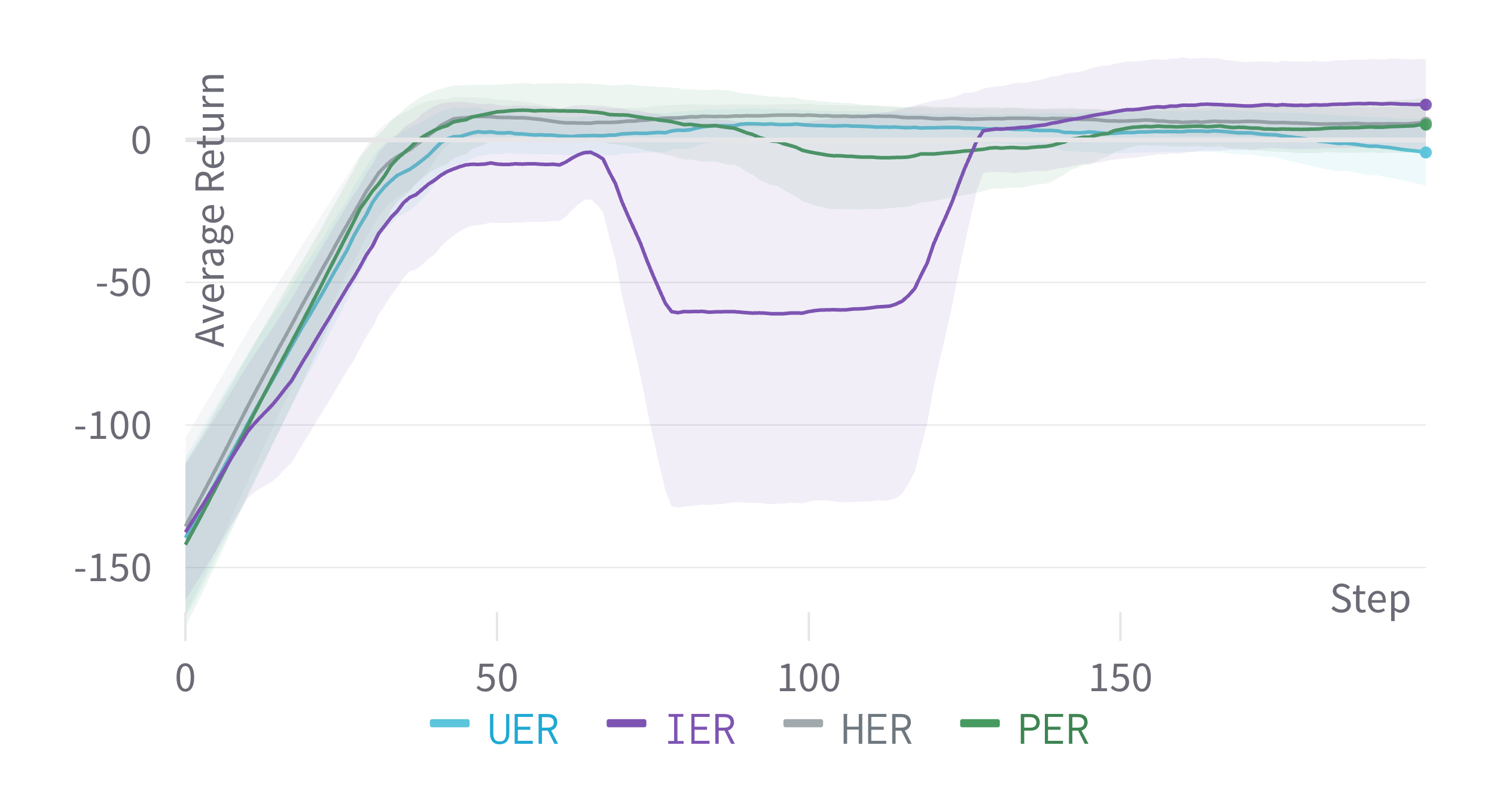}
\label{fig:lr_lunarlander}}
\caption{Learning curves of DQN/DDPG agents on Classic Control and Box-2D environments.}
\label{fig:cc}
\end{figure}

This section briefly discusses our results on environments with a low-dimensional state space, such as classic control environments (CartPole, Acrobot, and Pendulum) and Box-2D environments (LunarLander). Figure~\ref{fig:cc} depicts the learning curves of our DQN/DDPG agents in these environments.

We note that our proposed methodology can significantly outperform other baselines for the classic control algorithms. Furthermore, Figure~\ref{fig:lr_cartpole} shows excellent promise as we achieve a near-perfect score across all seeds in one-tenth of the time it took to train \namep.

\subsection{Performance in Multiple Joint Dynamics Simulation and Robotics Environments}

\begin{figure}[hbt!]
\centering
\subfigure[HalfCheetah-v2]{%
\includegraphics[width=0.45\linewidth]{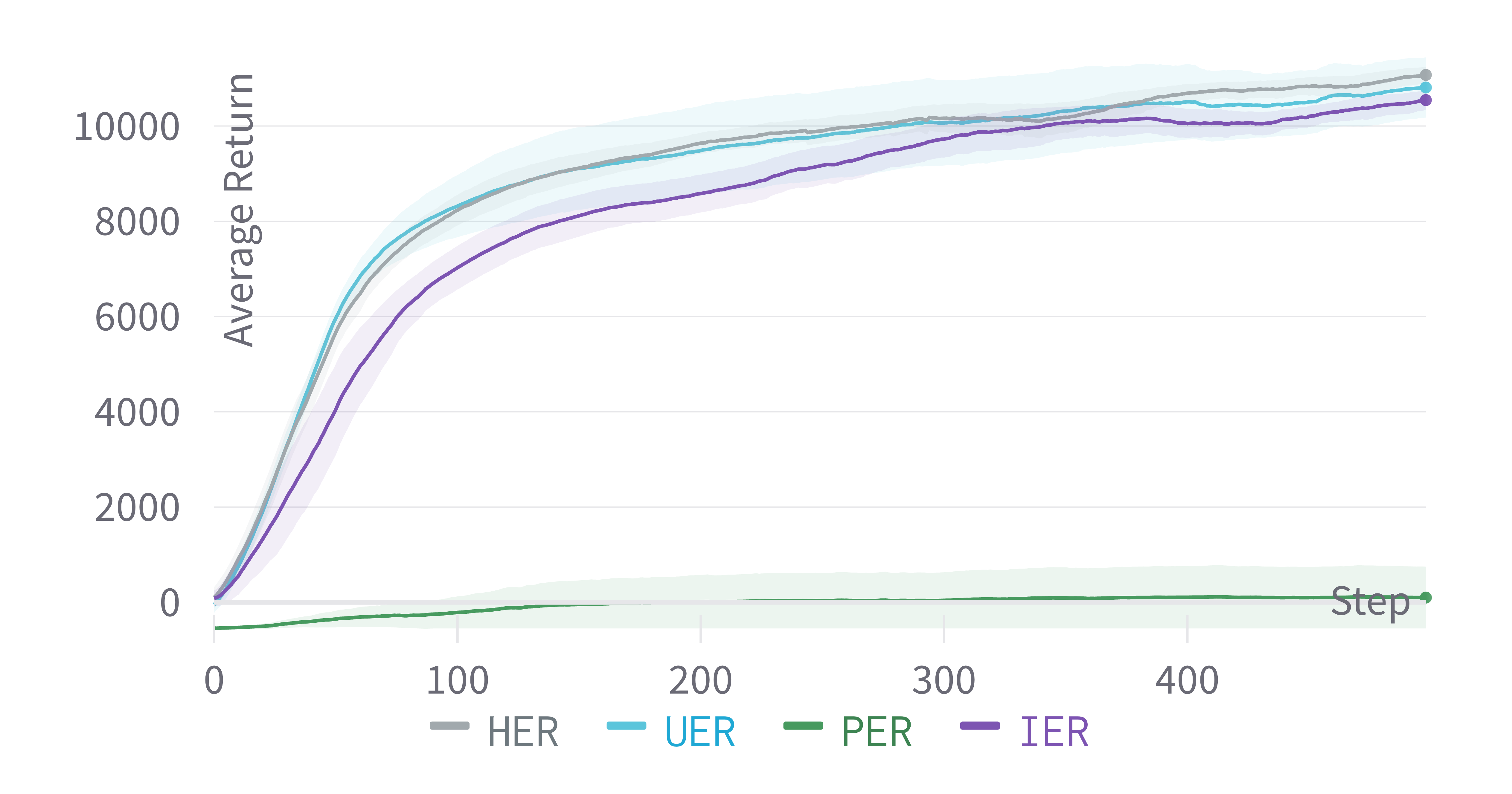}
\label{fig:lr_halfcheetah}}
\quad
\subfigure[Ant-v2]{%
\includegraphics[width=0.45\linewidth]{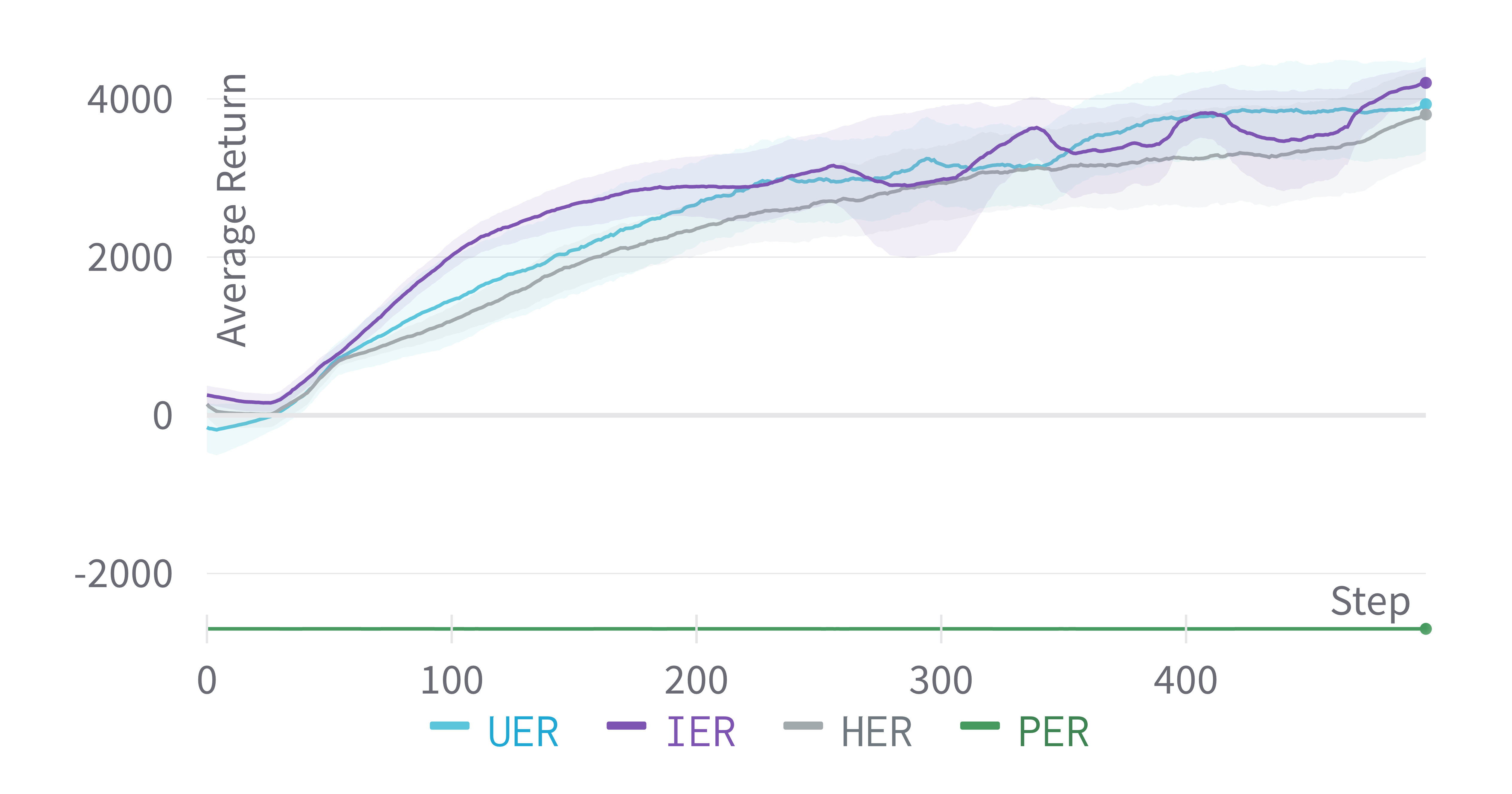}
\label{fig:lr_ant}}

\subfigure[Reacher-v2]{%
\includegraphics[width=0.45\linewidth]{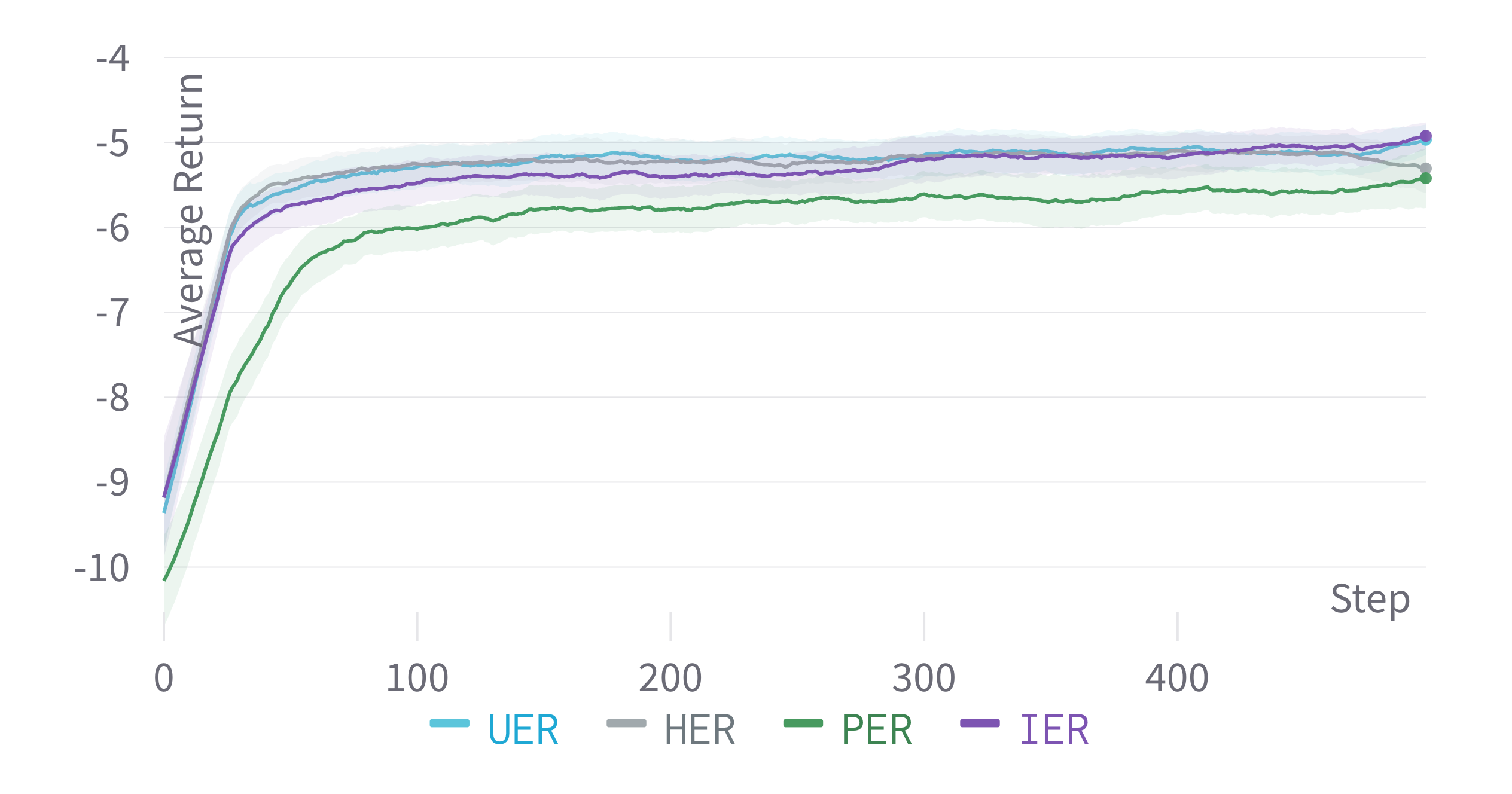}
\label{fig:lr_reacher}}
\quad
\subfigure[Walker-v2]{%
\includegraphics[width=0.45\linewidth]{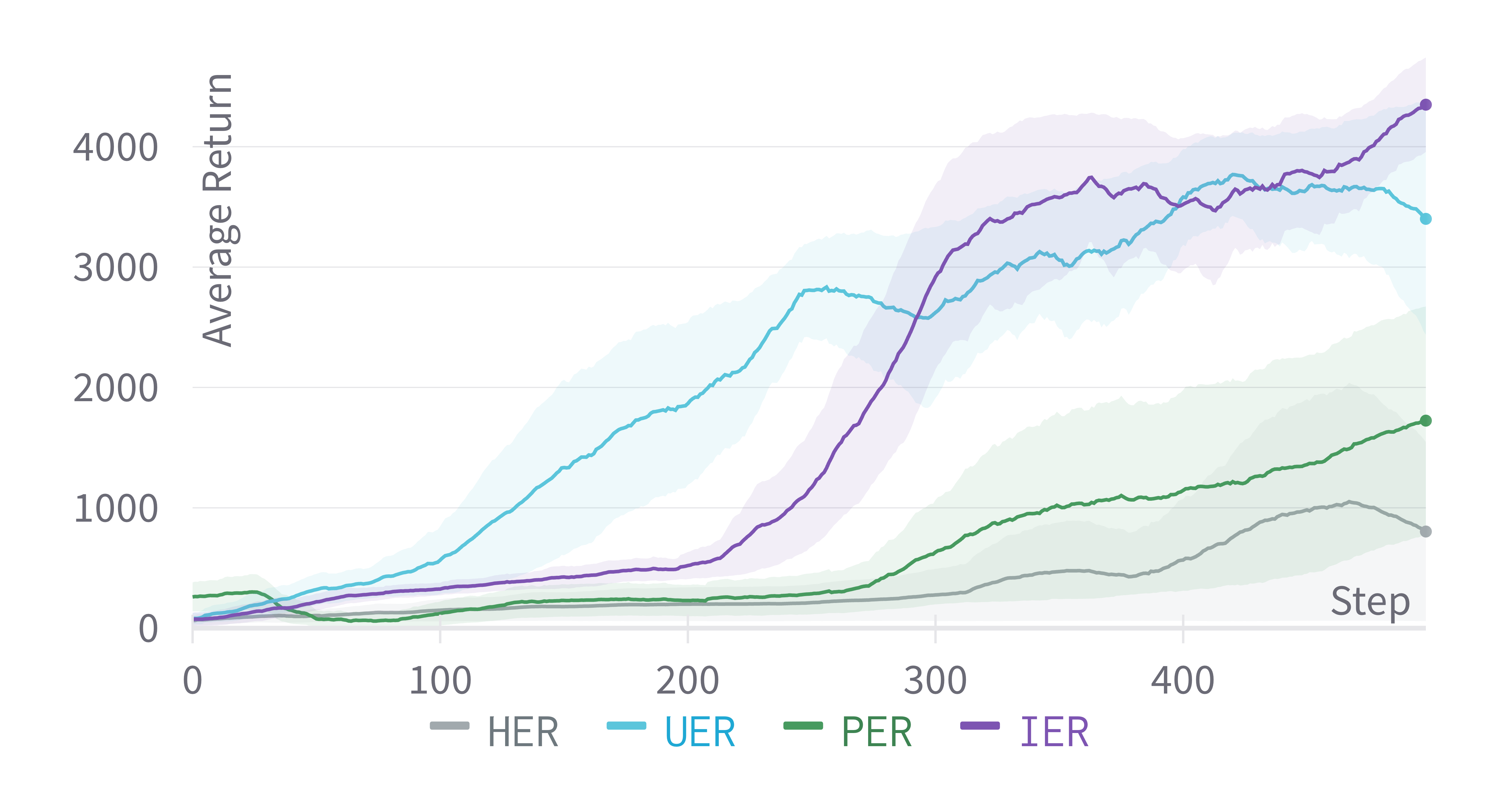}
\label{fig:lr_walker}}

\subfigure[Hopper-v2]{%
\includegraphics[width=0.45\linewidth]{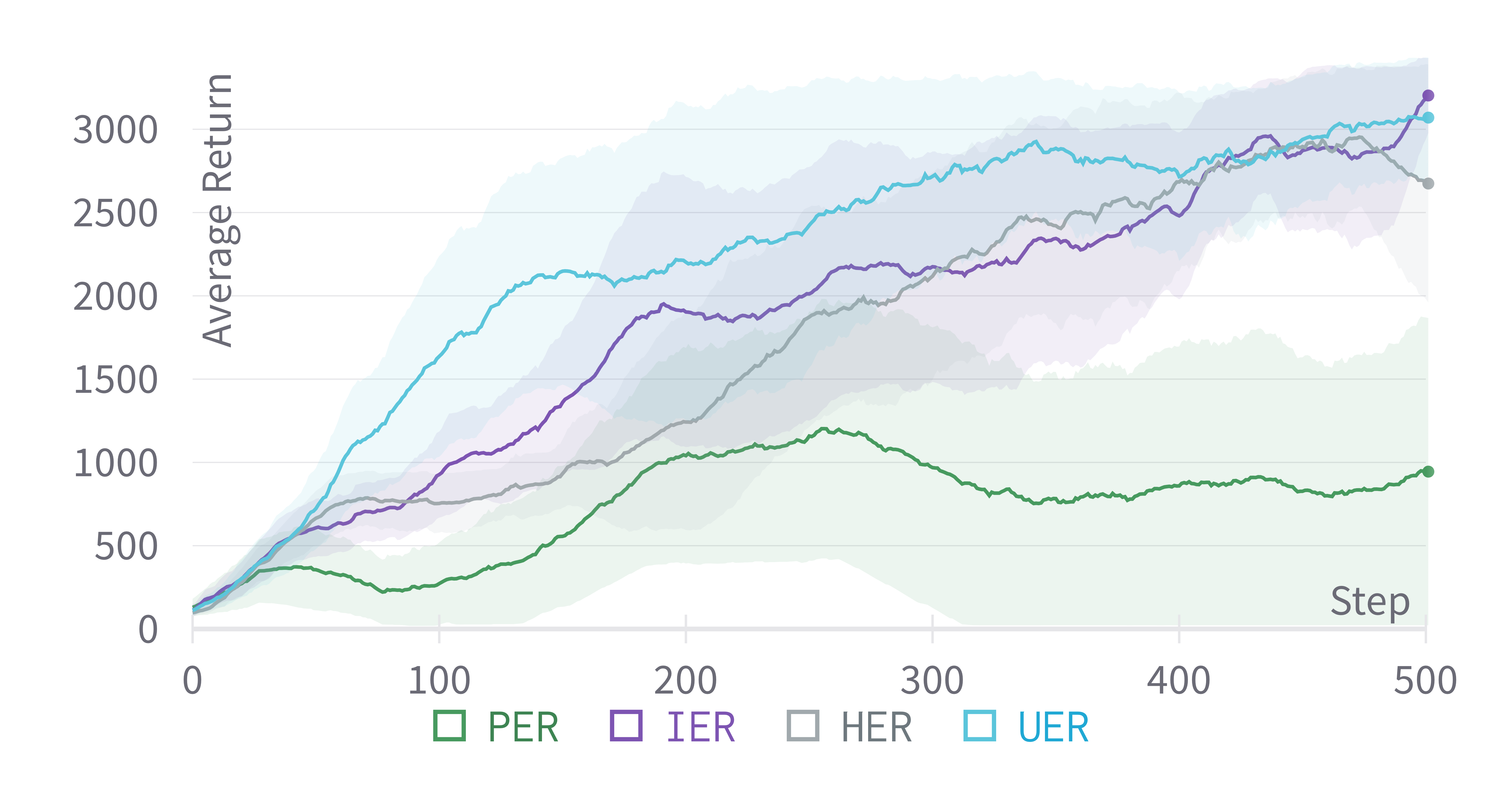}
\label{fig:lr_hopper}}
\quad
\subfigure[InvertedDoublePendulum-v2]{%
\includegraphics[width=0.45\linewidth]{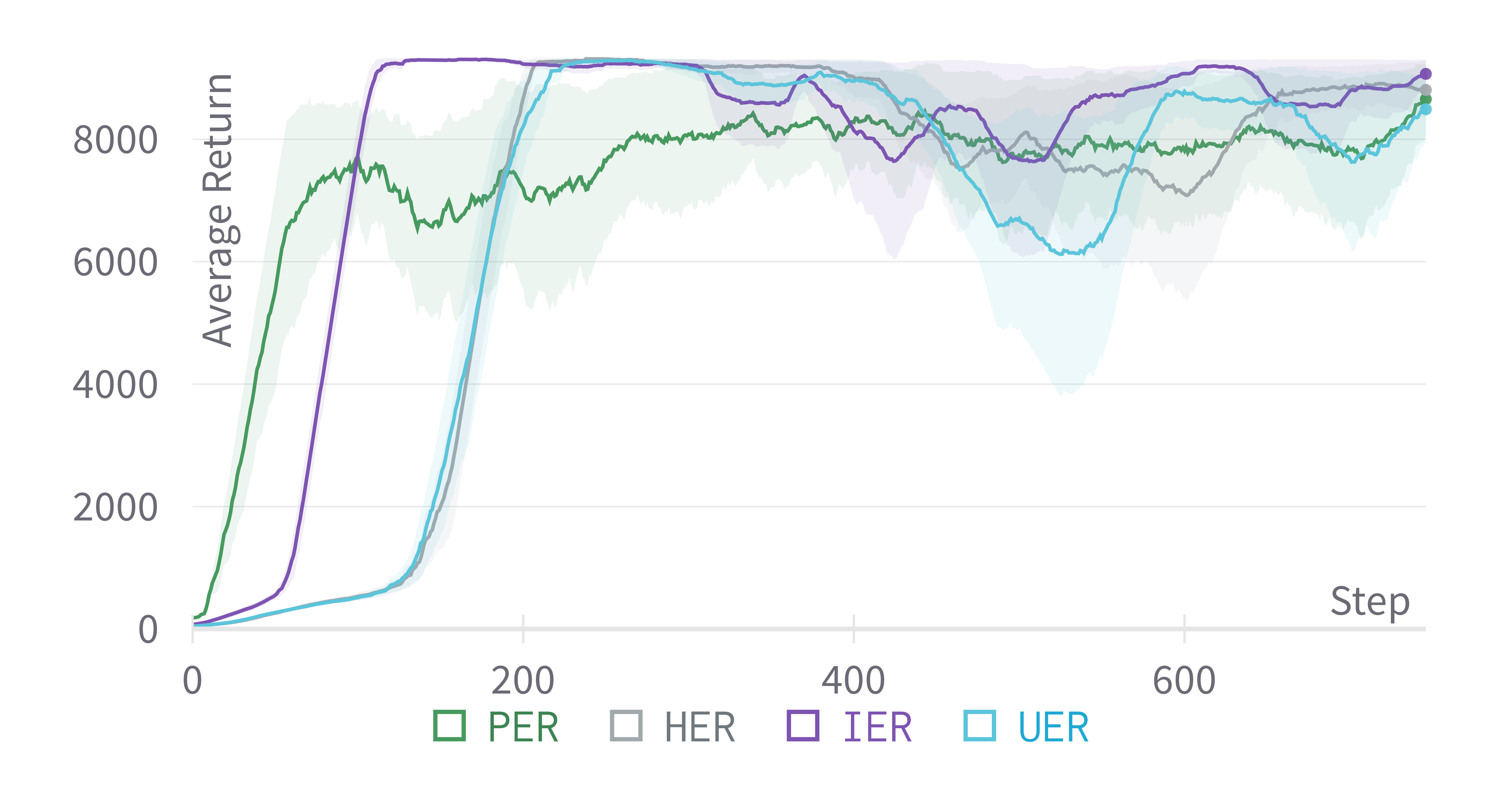}
\label{fig:lr_dpendulum}}

\subfigure[FetchReach-v1]{%
\includegraphics[width=0.45\linewidth]{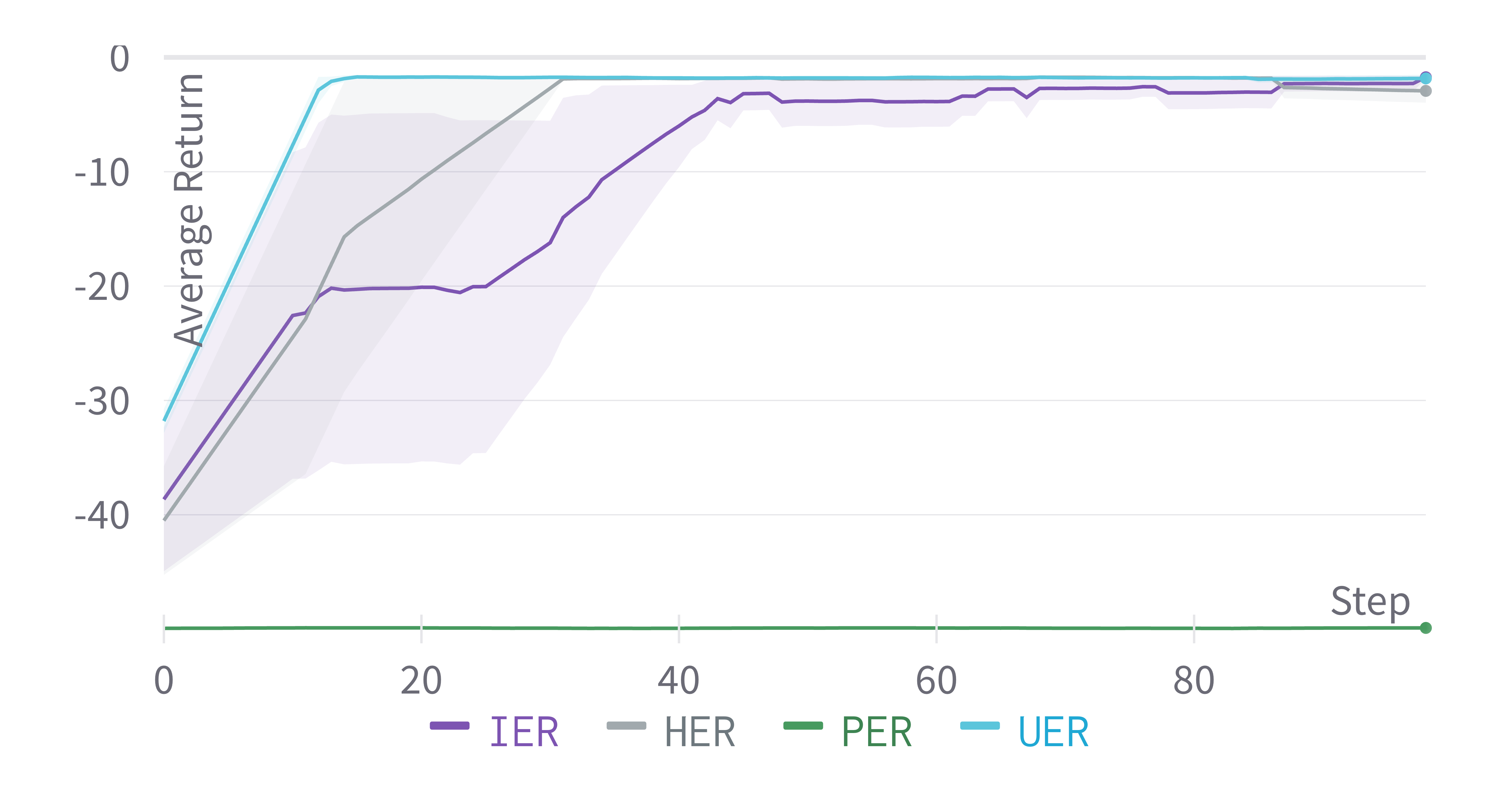}
\label{fig:lr_fetchreach}}

\caption{Learning curves of TD3 agents on Mujoco and Robotics environments.}
\label{fig:mu}
\end{figure}

Multiple joint dynamic simulation environments (mujoco physics environments) and robotics environments such as HalfCheetah, Ant, Inverted Double-Pendulum, and FetchReach (\cite{todorov2012mujoco,plappert2018multi}) are more complex and enable us to study whether the agent can understand the physical phenomenon of real-world environments. Figure~\ref {fig:mu} depicts the learning curves of our TD3 agents in these environments. Again, our proposed methodology outperforms all other baselines significantly in most of the environments studied in this section. Additionally, it is essential to point out that our proposed method shows an impressive speedup of convergence in Inverted Double Pendulum and convergence to a much better policy in Ant. 

\subsection{Performance of \namel~in Human Challenging Environments}
\begin{figure}[hbt!]
\centering
\subfigure[Pong-v0]{%
\includegraphics[width=0.45\linewidth]{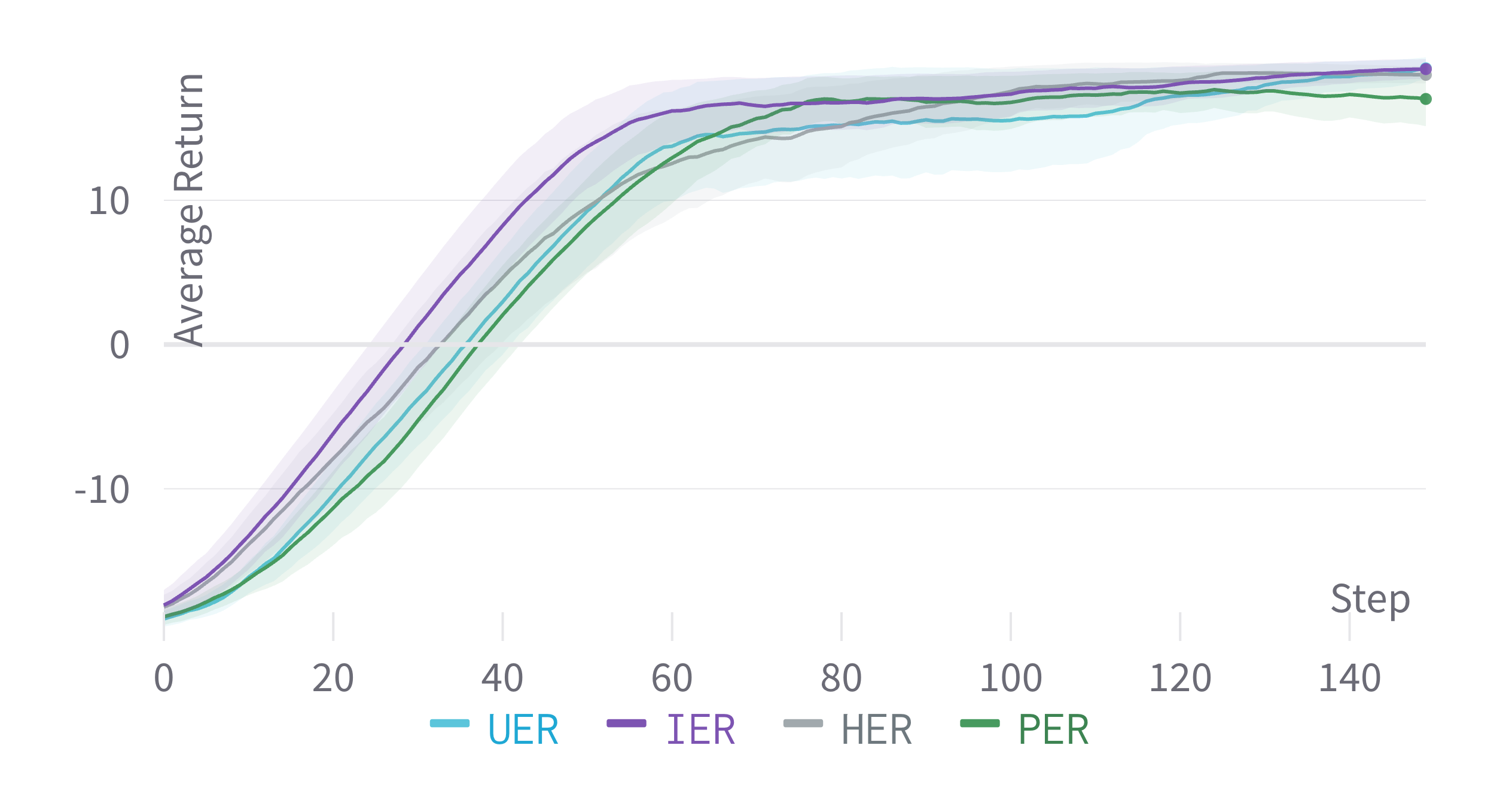}
\label{fig:lr_pong}}
\quad
\subfigure[Enduro-v0]{%
\includegraphics[width=0.45\linewidth]{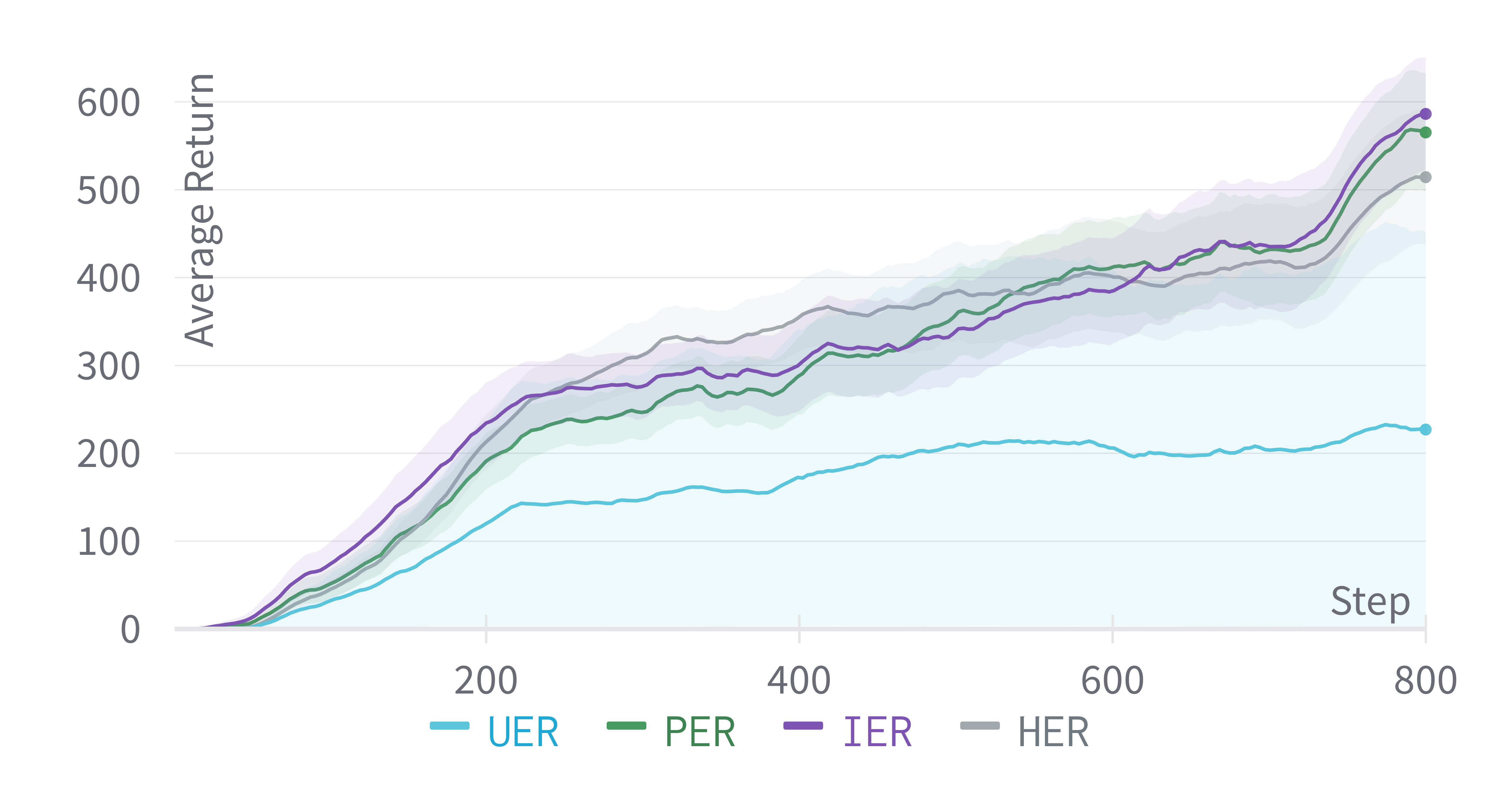}
\label{fig:lr_enduro}}

\caption{Learning curves of DQN agents on Atari environments.}
\label{fig:at}
\end{figure}

This section briefly discusses our results on human-challenging environments such as Atari environments (Pong and Enduro). These environments are highly complex, and our algorithms take millions of steps to converge to a locally optimal policy. Figure~\ref{fig:at} depicts the learning curves of our DQN agents in these environments. We note that our proposed methodology can perform favorably when compared to other baselines for the Atari environments and can reach large reward policies significantly faster than \nameu.

\subsection{Whole vs. Component Parts}

This section briefly presents the learning curves of our models on three different sampling schemes: \namel, \nameo~and \namelo.

\begin{figure}[hbt!]
\centering
\subfigure[CartPole-v0]{%
\includegraphics[width=0.30\linewidth]{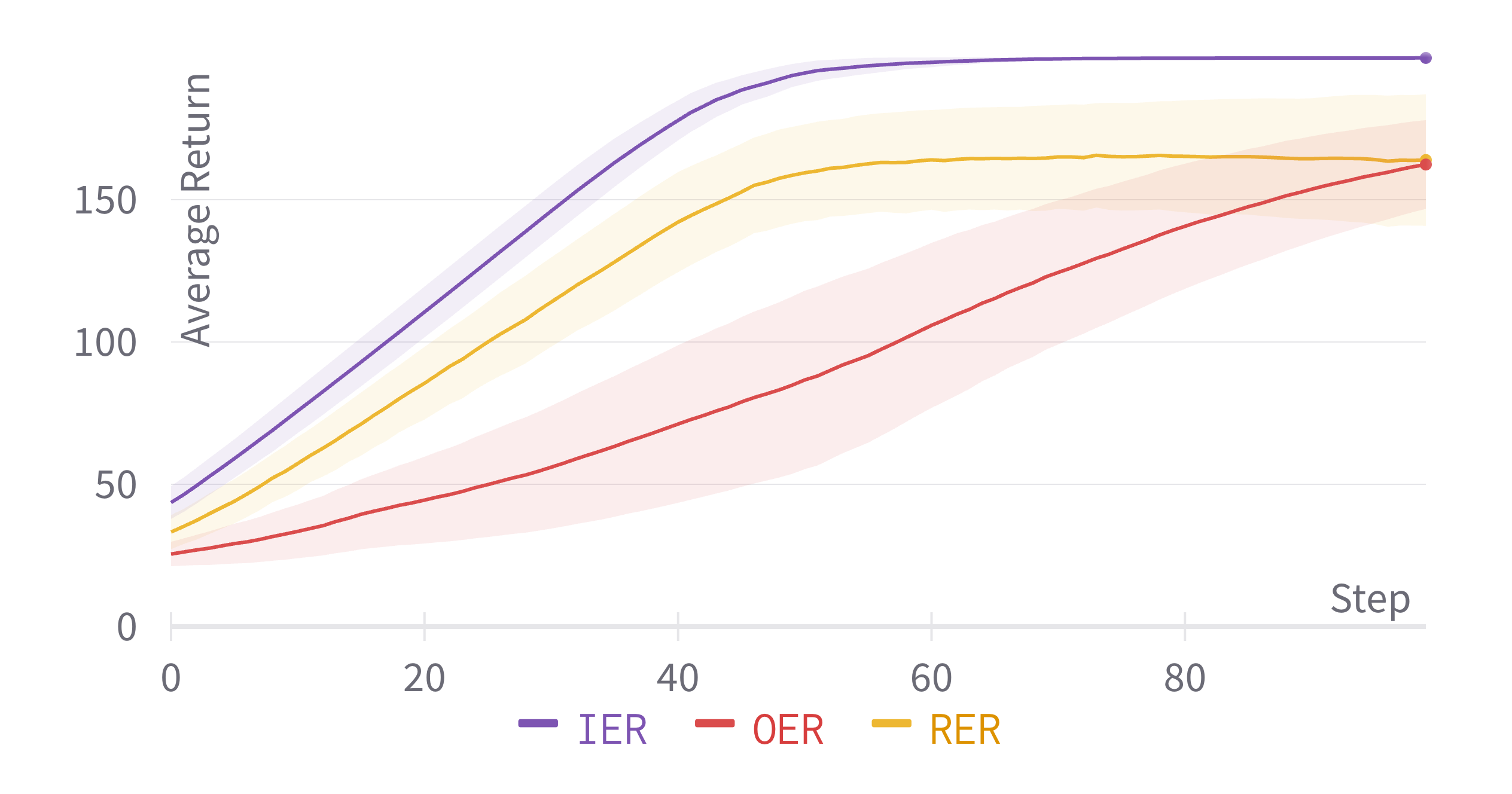}
}
\quad
\subfigure[Acrobot-v1]{%
\includegraphics[width=0.30\linewidth]{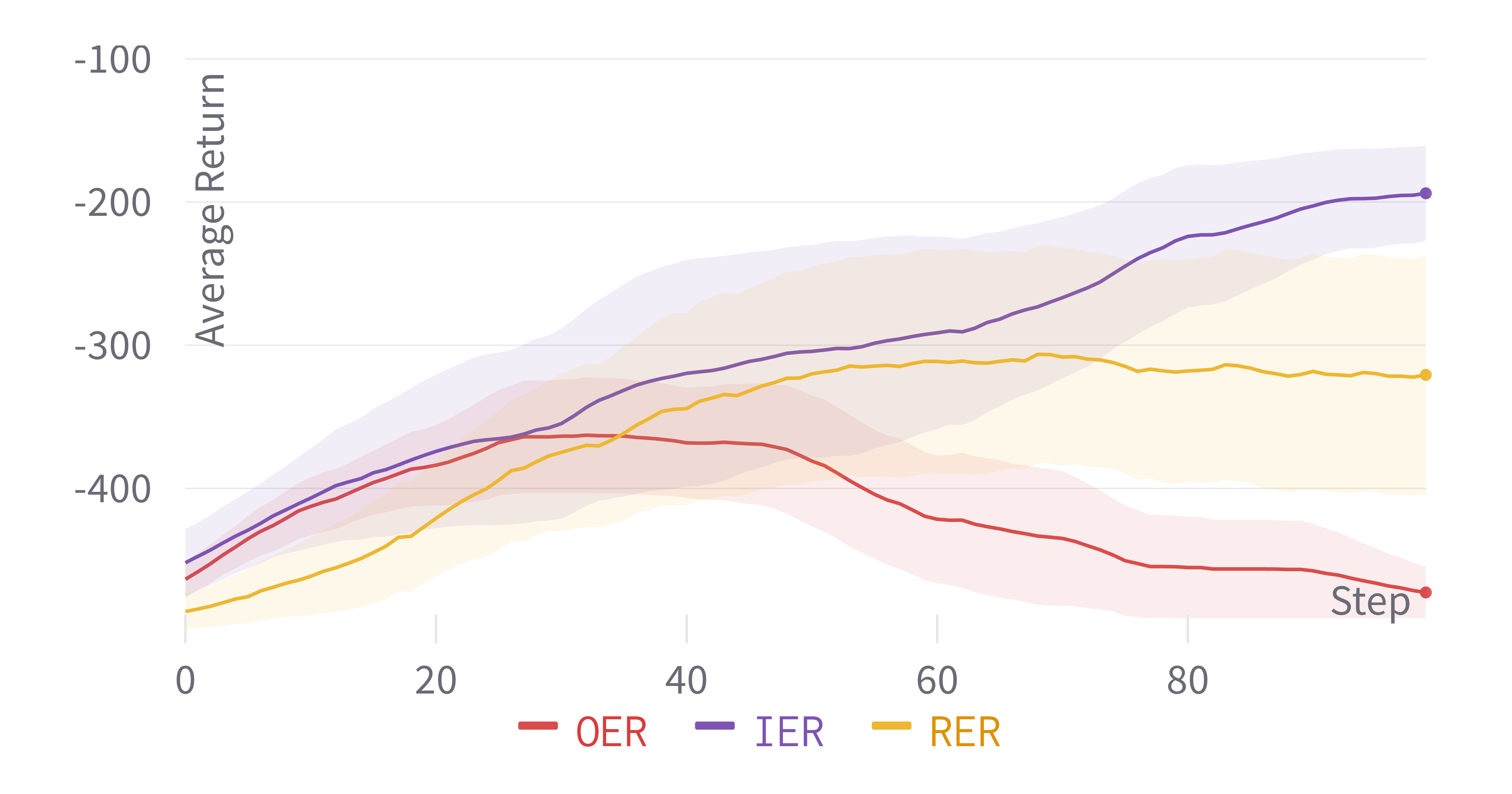}
}
\quad
\subfigure[Pendulum-v0]{%
\includegraphics[width=0.30\linewidth]{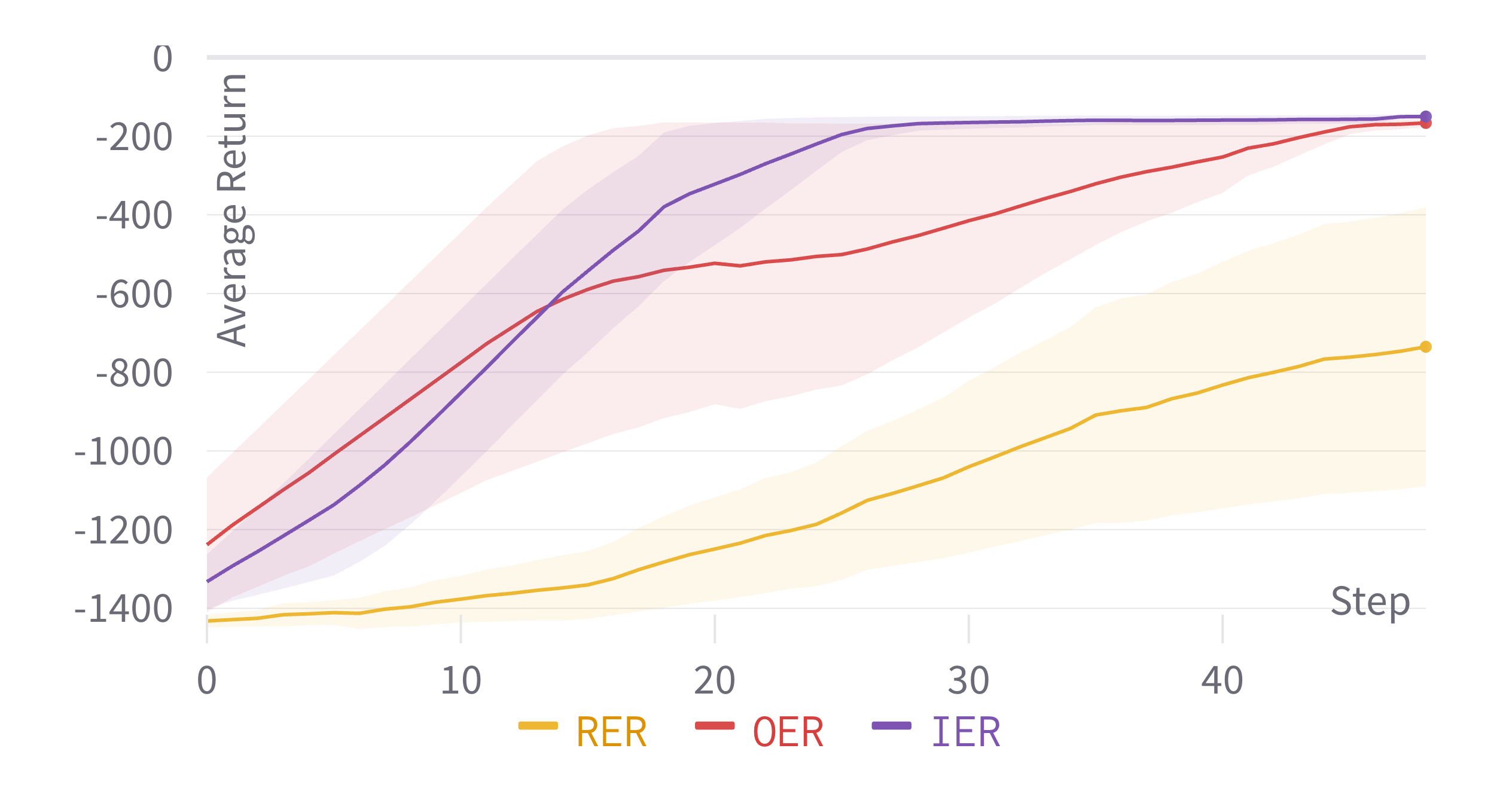}
}

\subfigure[LunarLander-v2]{%
\includegraphics[width=0.30\linewidth]{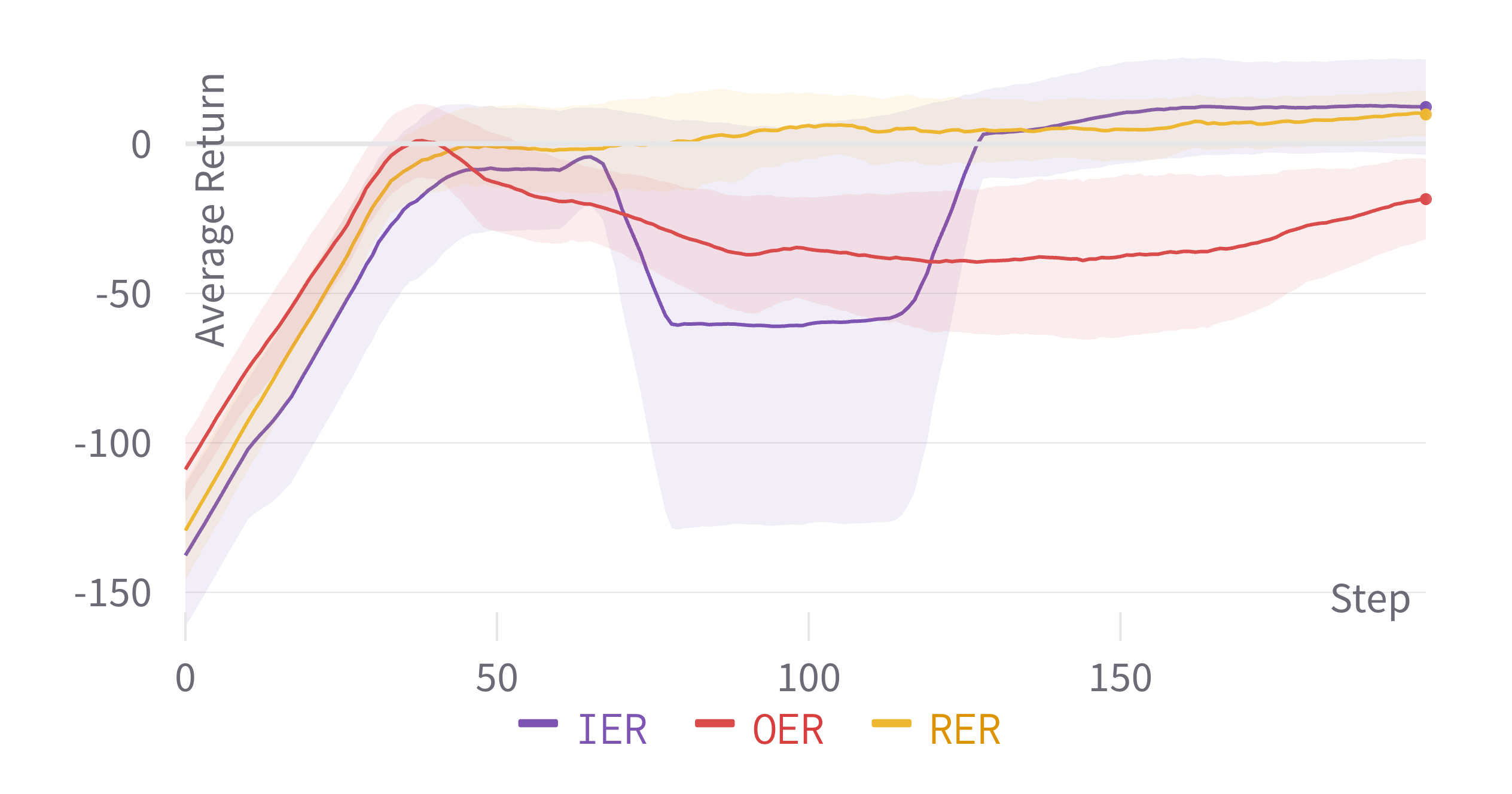}
}
\quad
\subfigure[HalfCheetah-v2]{%
\includegraphics[width=0.30\linewidth]{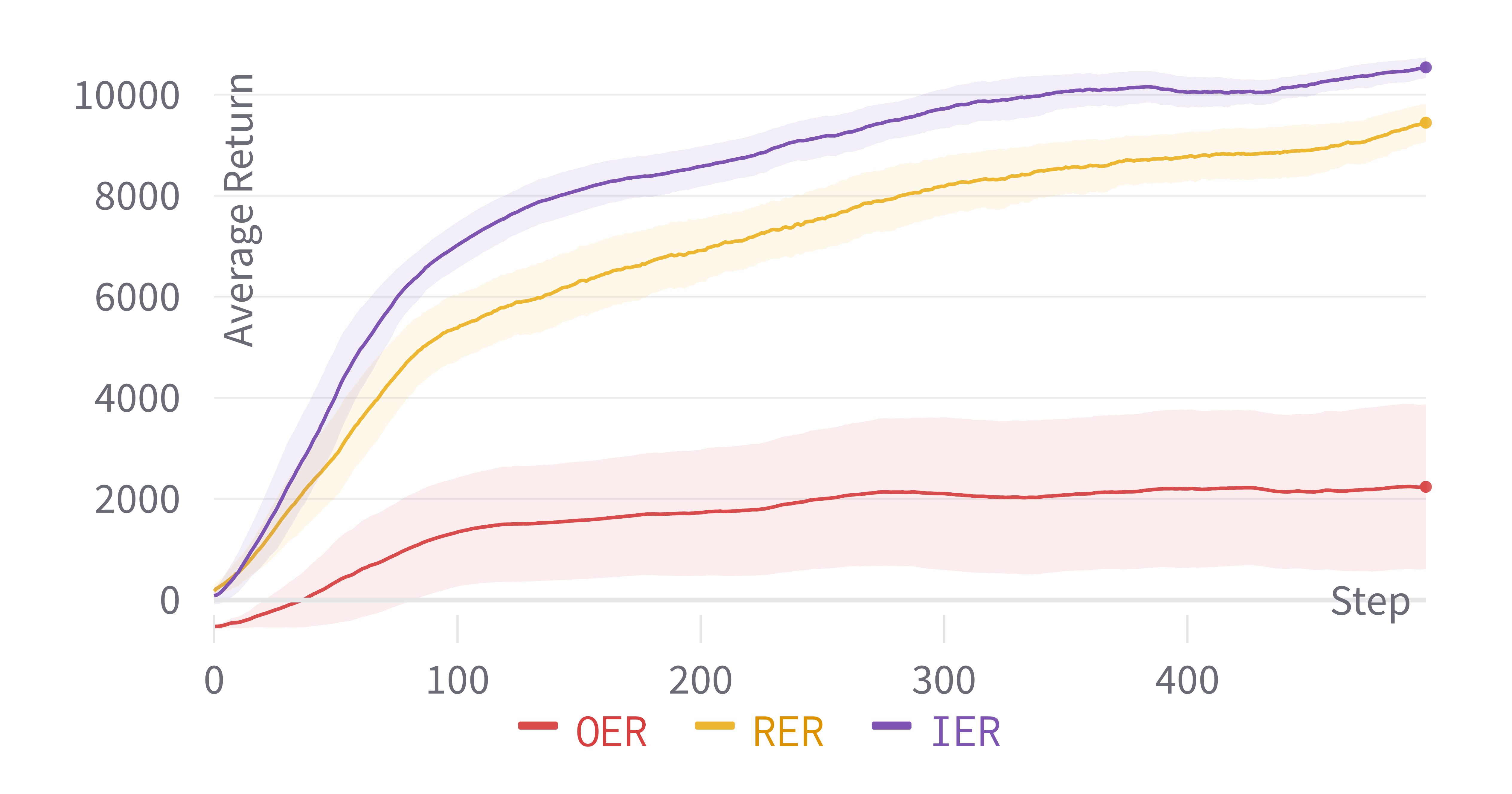}
}
\quad
\subfigure[Ant-v2]{%
\includegraphics[width=0.30\linewidth]{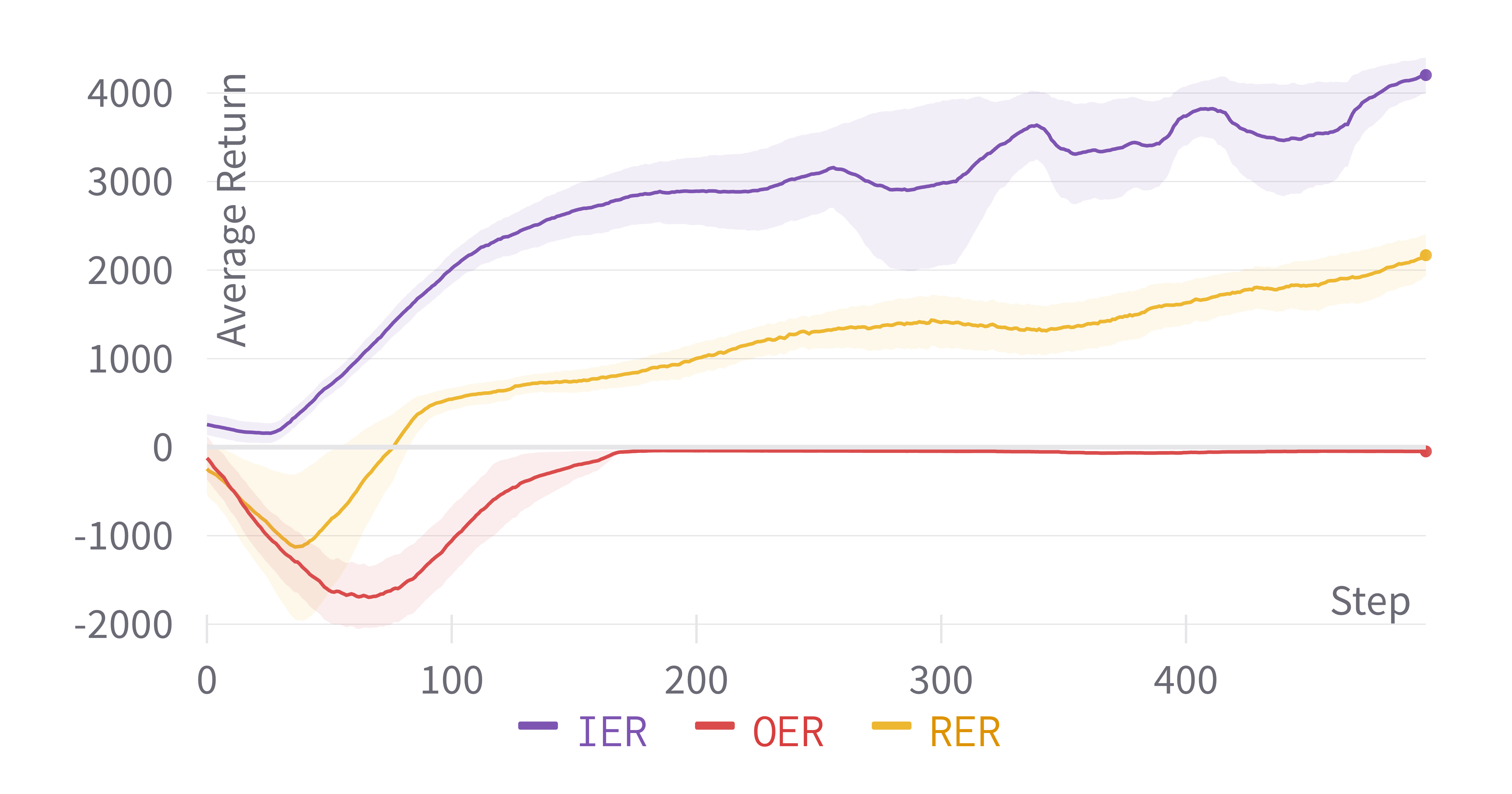}
}

\subfigure[Reacher-v2]{%
\includegraphics[width=0.30\linewidth]{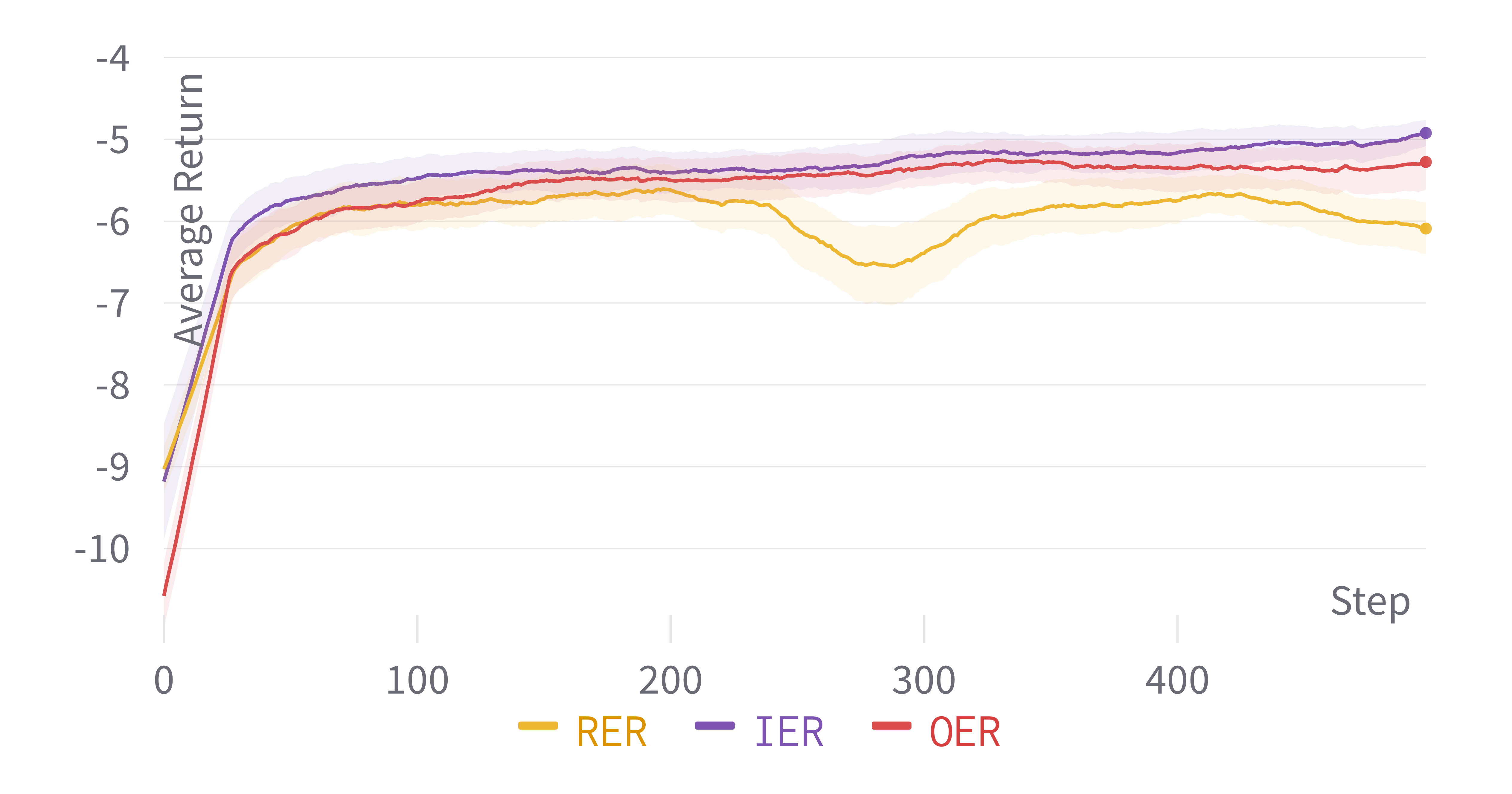}
}
\quad
\subfigure[Walker-v2]{%
\includegraphics[width=0.30\linewidth]{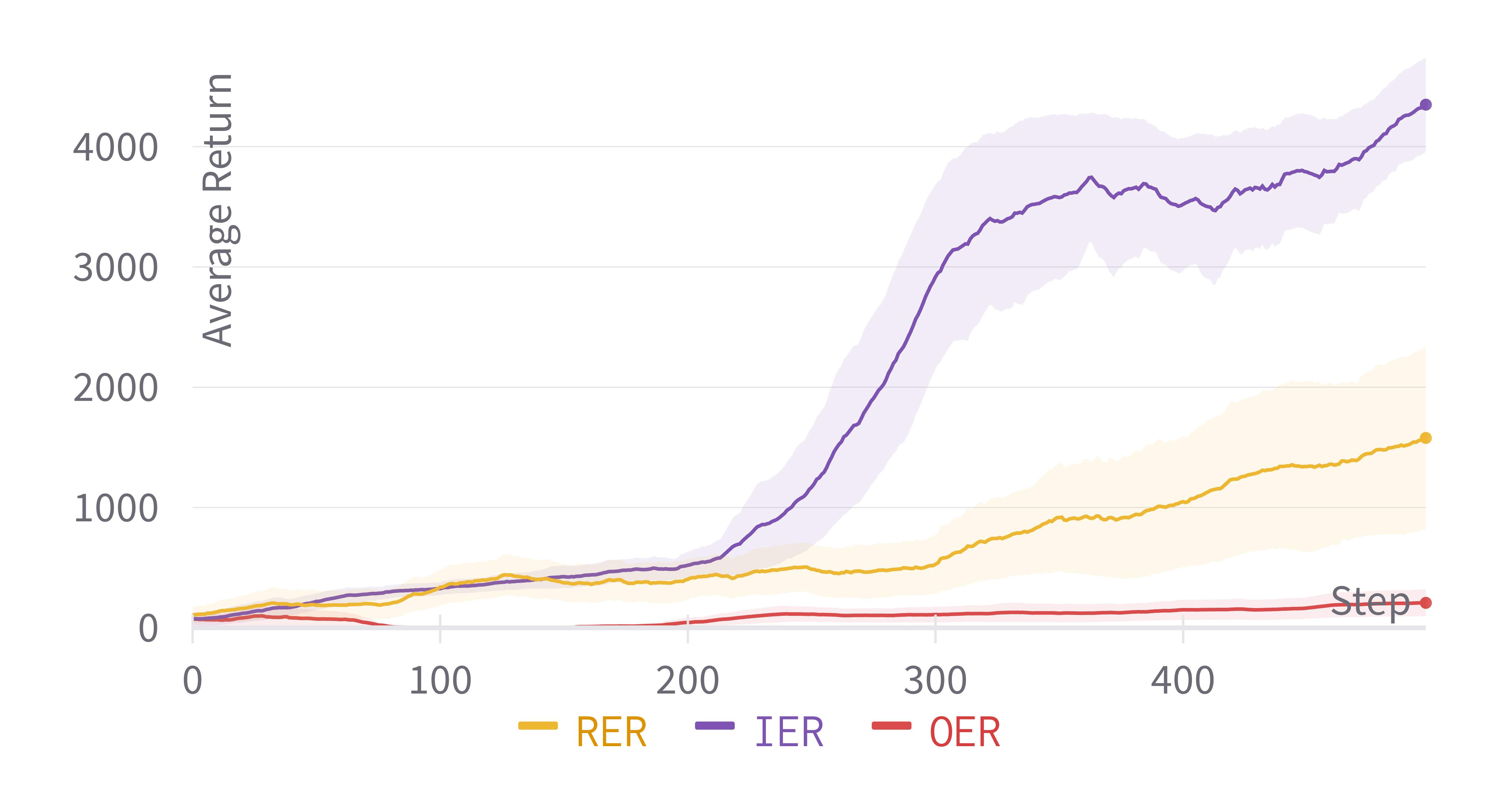}
}
\quad
\subfigure[Hopper-v2]{%
\includegraphics[width=0.30\linewidth]{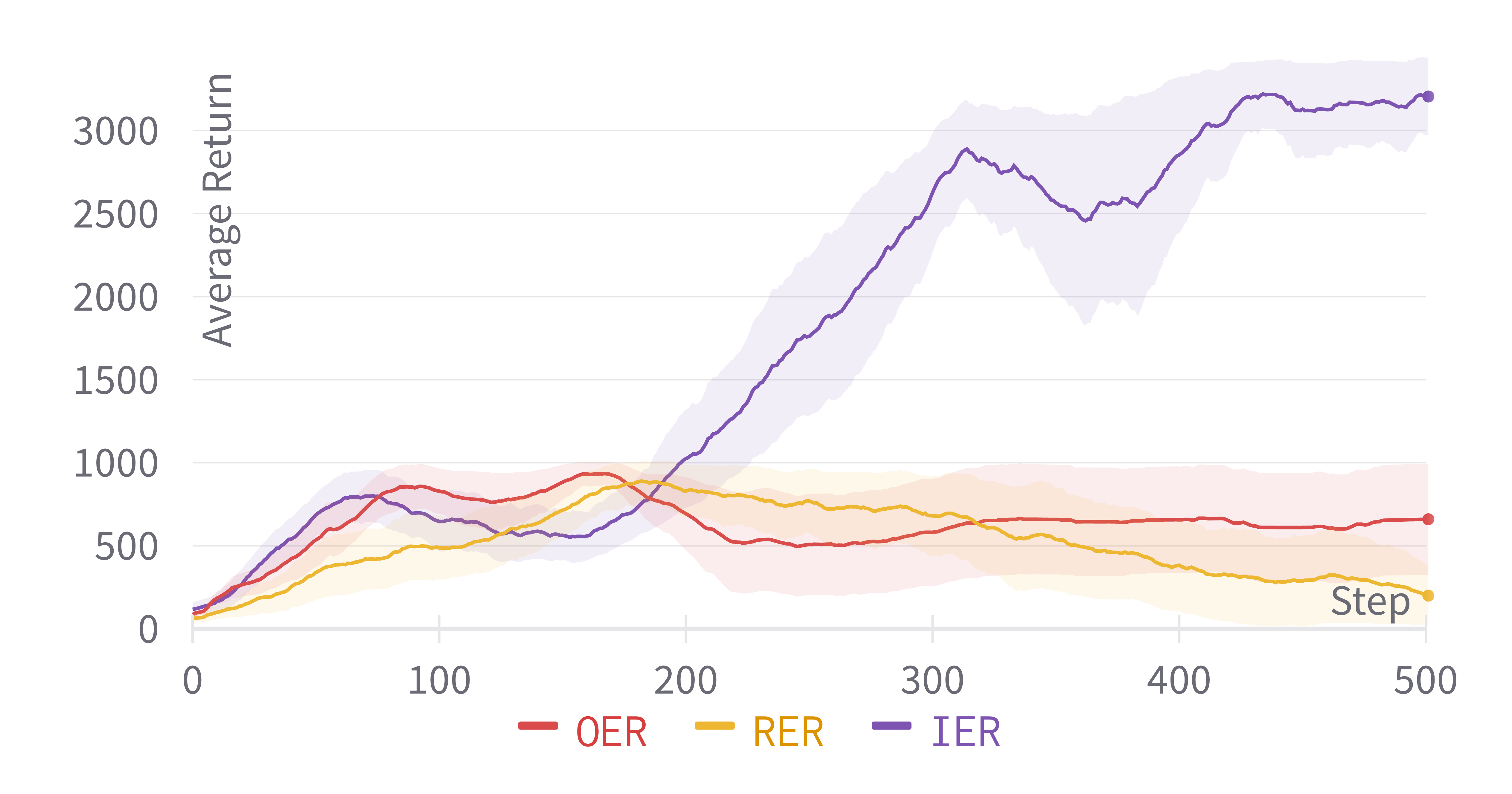}
}

\subfigure[InvertedDoublePendulum-v2]{%
\includegraphics[width=0.30\linewidth]{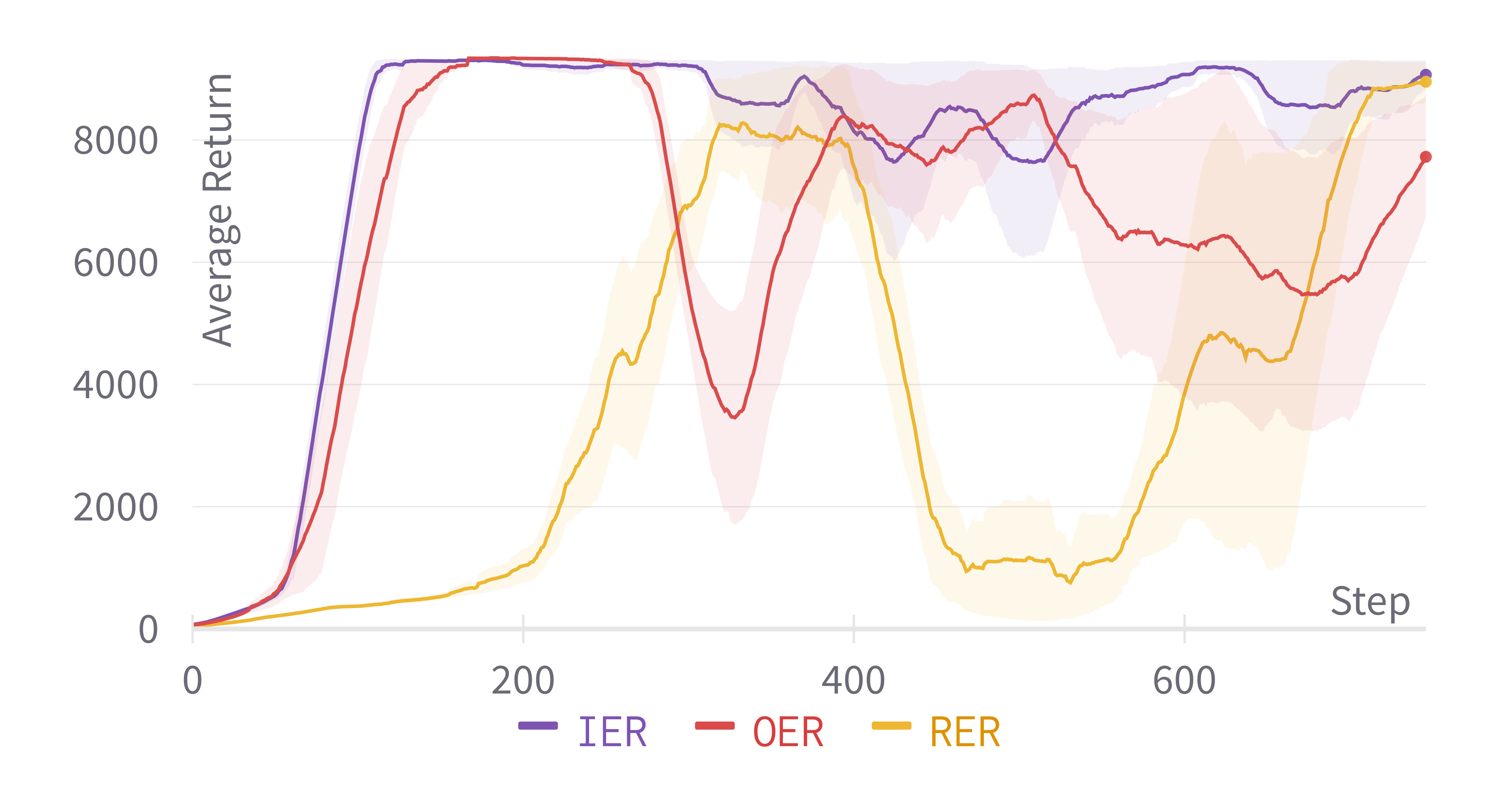}
}
\quad
\subfigure[FetchReach-v1]{%
\includegraphics[width=0.30\linewidth]{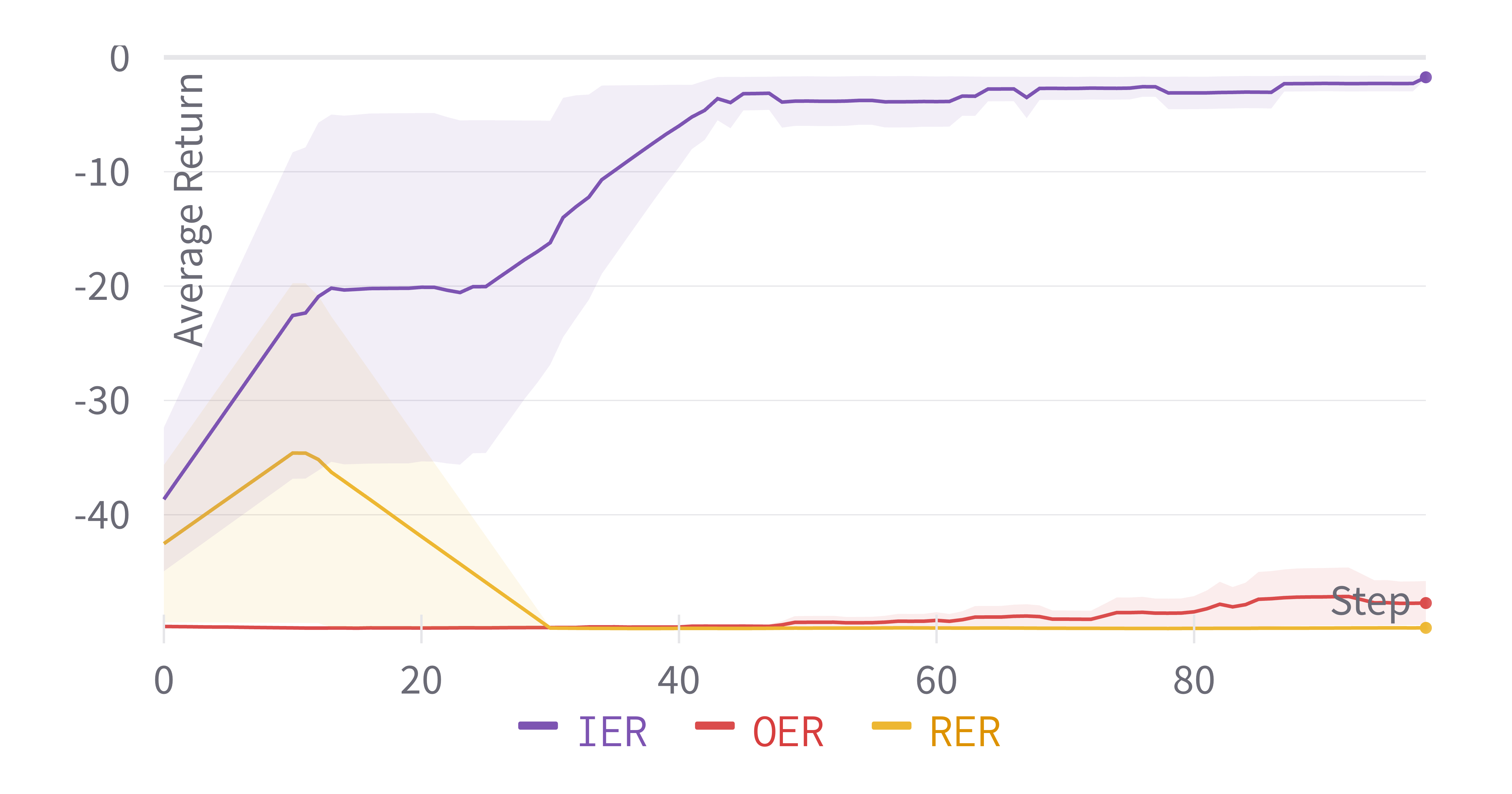}
}
\quad
\subfigure[Pong-v0]{%
\includegraphics[width=0.30\linewidth]{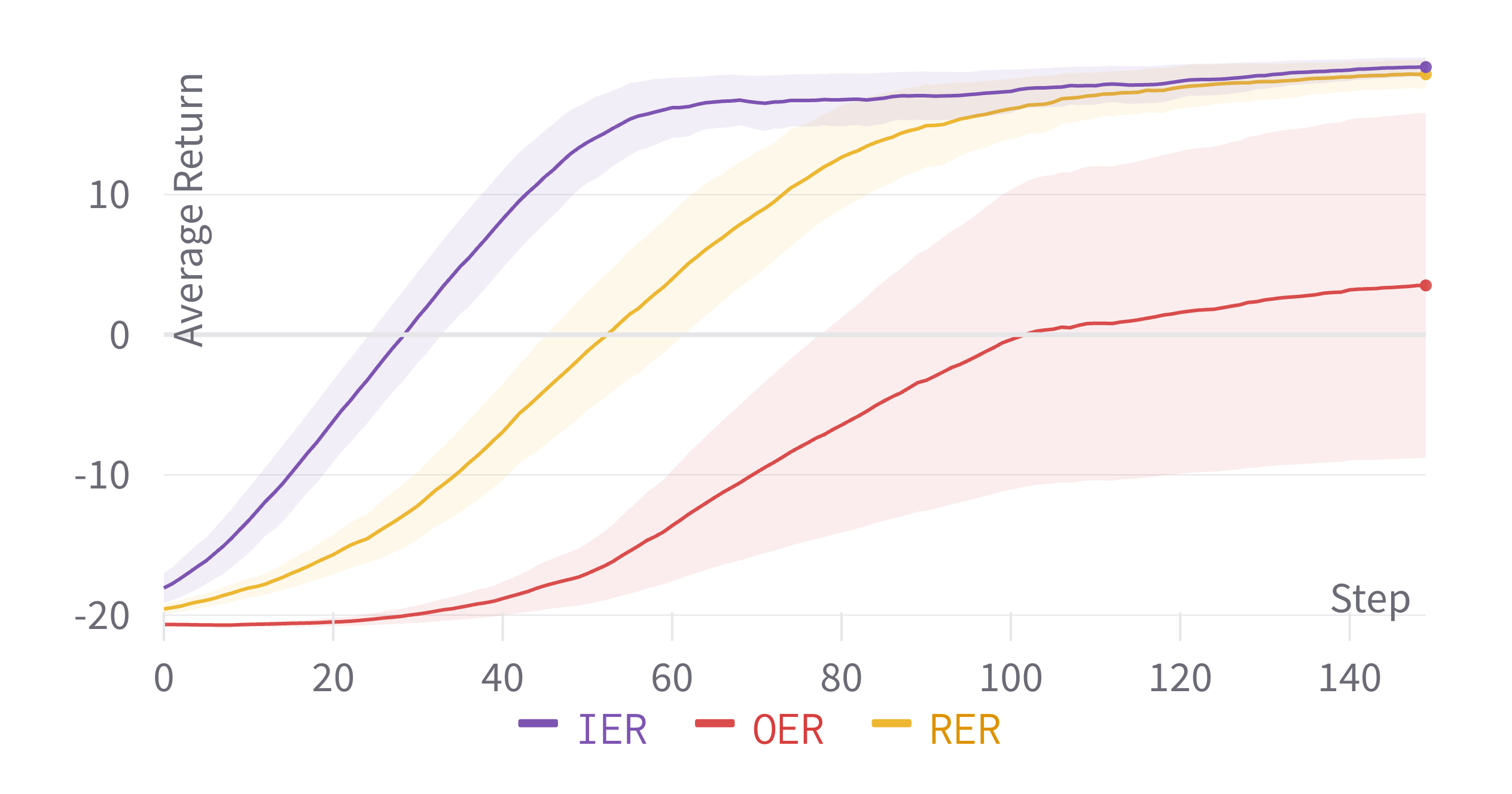}
}

\subfigure[Enduro-v0]{%
\includegraphics[width=0.30\linewidth]{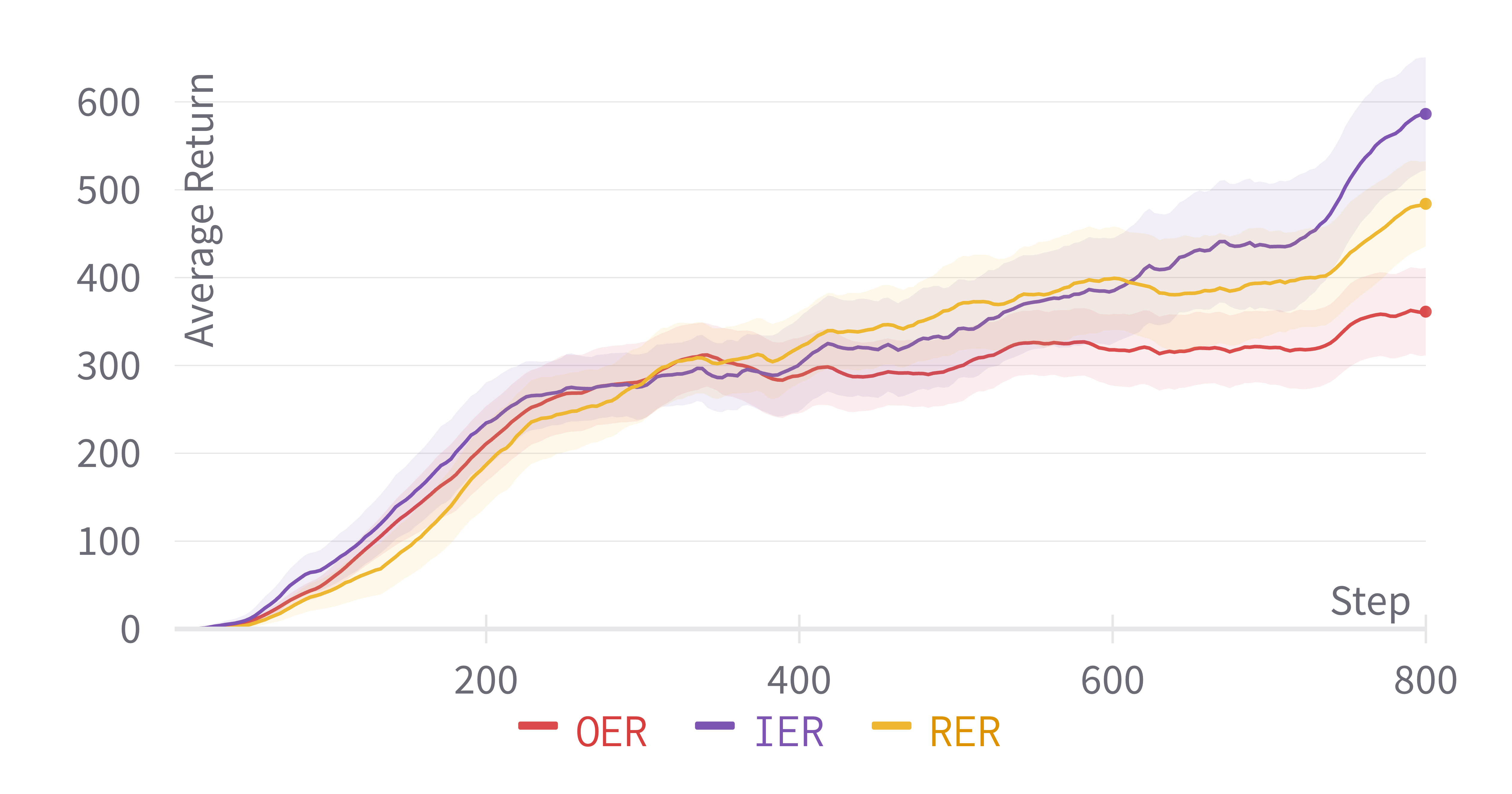}
}
\caption{Ablation study of \nameo, \namelo~and \namel.}
\label{fig:ablation_study}
\end{figure}

\subsection{Buffer Batch size sensitivity of \namel}
\label{buffer_size_ablation}

\begin{figure}[h!]
\centering
\includegraphics[width=0.7\linewidth]{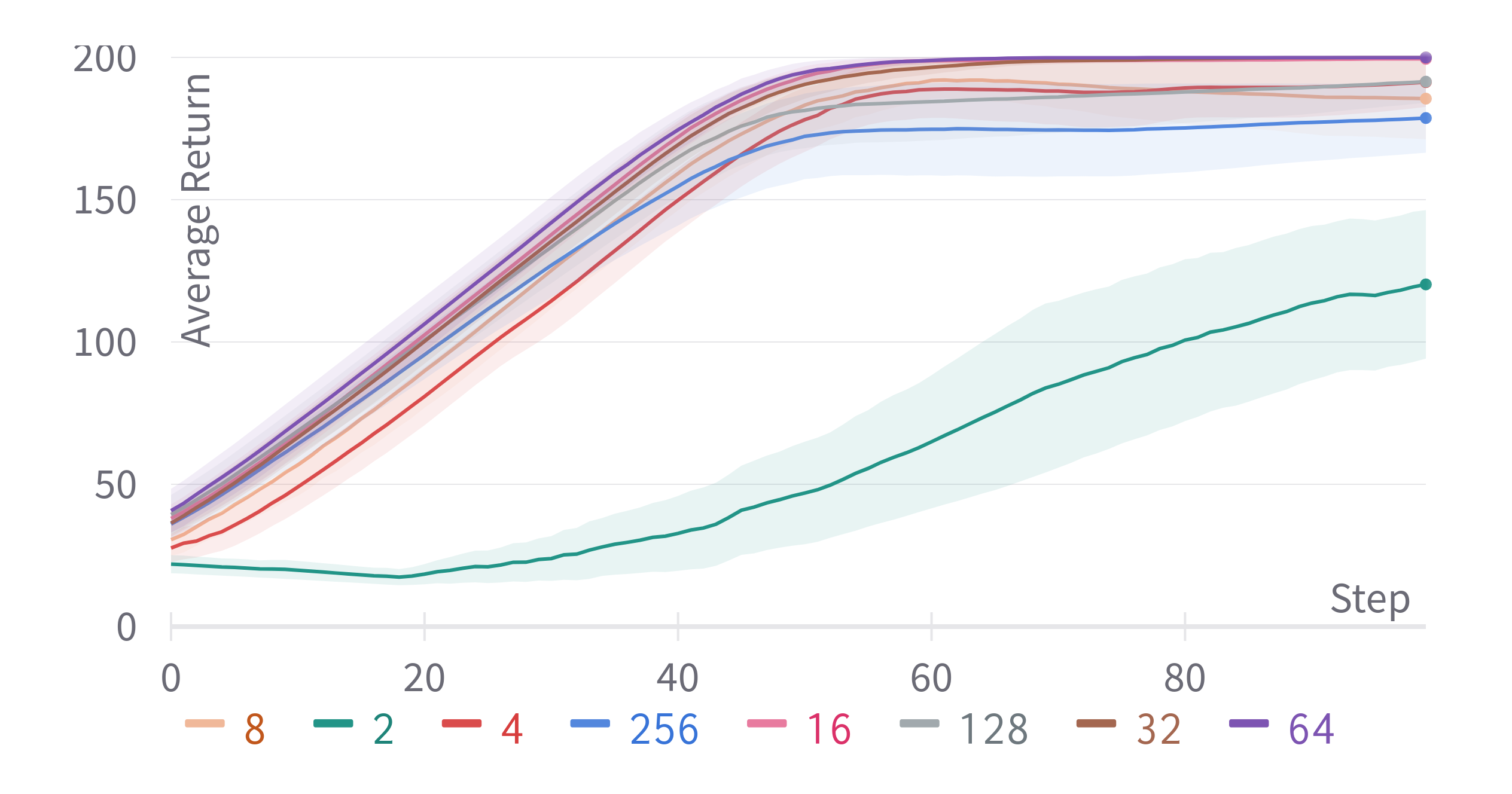}
\caption{Buffer batch size sensitivity of \namel~sampler on the CartPole Environment}
\label{fig:buffer_size_sensitivity}
\end{figure}

This section briefly presents the sensitivity to the buffer batch size hyperparameter for our proposed approach (\namel). To analyze this, we run our experiments on the CartPole environment with varying batch size of the range 2-256. Table~\ref{tab:buffer_size_sensitivity} and Figure~\ref{fig:buffer_size_sensitivity} depict the buffer batch size sensitivity results from our proposed sampler.

\begin{table}[h!]
    \centering
    \caption{Buffer Batch size sensitivity of \namel~on the CartPole environment.}
    \vspace*{1mm}
    \label{tab:buffer_size_sensitivity}
    \resizebox{0.4\columnwidth}{!}{
    \begin{tabular}{lcl}
        \toprule
        \textbf{Buffer Batch Size} & {\textit{\textbf{Average Reward}}}\\ \midrule
\midrule

\text{2}  & 126.69 {\tiny $\pm$ 41.42} \\ \midrule
\text{4} & 192.33 {\tiny $\pm$ 13.29} \\ \midrule
\text{8} & 181.27 {\tiny $\pm$ 32.13} \\  \midrule
\text{16} & 199.24 {\tiny $\pm$ 1.32} \\ \midrule
\text{32} & \textbf{199.99} {\tiny $\pm$ 0.001} \\ \midrule
\text{64} & 199.83 {\tiny $\pm$ 0.31} \\ \midrule
\text{128} & 193.23 {\tiny $\pm$ 10.08} \\ \midrule
\text{256} & 179.95 {\tiny $\pm$ 18.94} \\
\bottomrule
    \end{tabular}
    }
\end{table}

\subsection{How important is sampling pivots?}
\label{pivot_ablation}

\begin{figure}[h!]
\centering
\includegraphics[width=0.7\linewidth]{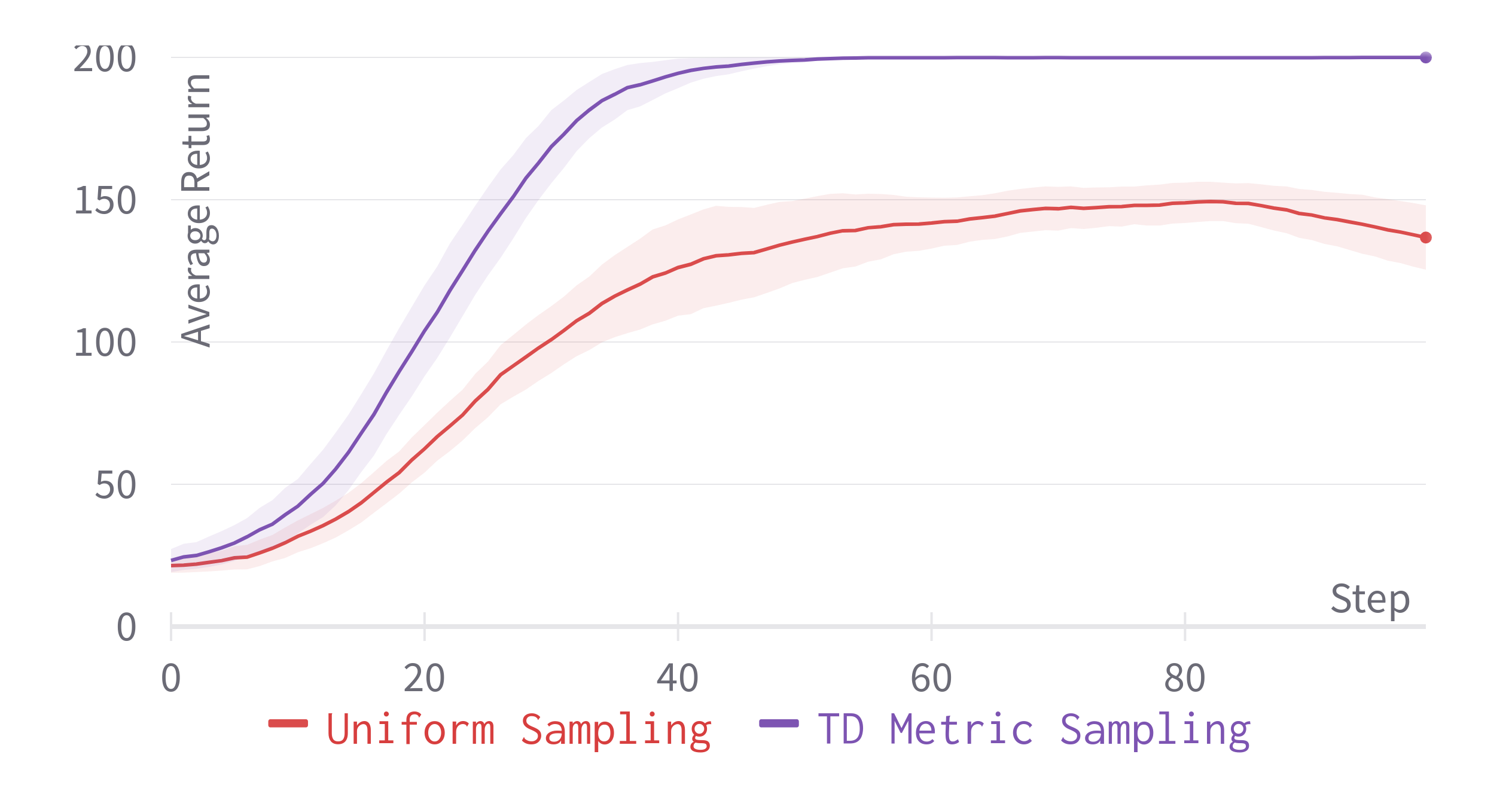}
\caption{Ablation study of Importance Sampling of \namel~sampler on the CartPole Environment. Here ``Uniform Sampling'' denotes the uniformly random sampling of pivots, and ``TD Metric Sampling'' denotes our proposed approach (\namel).}
\label{fig:sampling_pivots}
\end{figure}

This section briefly presents the ablation study to analyze the importance of sampling ``surprising'' states as pivots. As a baseline, we build a experience replay where these pivots are randomly sampled from the buffer. The "looking back" approach is used to create batches of data. For nomenclature, we refer to our proposed approach (\namel) to use the ``TD Metric'' sampling of pivots, and the baseline that uses ``Uniform'' sampling of pivots. 
Table~\ref{tab:sampling_pivots} and Figure~\ref{fig:sampling_pivots} depict the buffer batch size sensitivity results from our proposed sampler.

\begin{table}[h!]
    \centering
    \caption{Importance of sampling pivots in our proposed approach (\namel) on the CartPole environment.}
    \vspace*{1mm}
    \label{tab:sampling_pivots}
    \resizebox{0.4\columnwidth}{!}{
    \begin{tabular}{lcl}
        \toprule
        \textbf{Sampling Scheme} & {\textit{\textbf{Average Reward}}}\\ \midrule
\midrule

\text{TD Metric (\namel)}  & \textbf{199.83}  {\tiny $\pm$ 0.31} \\ \midrule
\text{Uniform (\namel)} & 136.71 {\tiny $\pm$ 19.59} \\ 
\bottomrule
    \end{tabular}
    }
\end{table}

\subsection{How important is ``looking back''?}
\label{lookingback_ablation}

This section briefly presents the ablation study to analyze the importance of ``looking back'' after sampling pivots. As a baseline, we build a experience replay where we sample uniformly instead of looking back. For nomenclature, we refer to our proposed approach (\namel) to use the ``Looking Back'' approach (similar to \namel), and the baseline that uses ``Uniform'' approach. We refer to these two approaches as possible filling schemes, i.e. fill the buffer with states once the pivot state is sampled.

\begin{figure}[h!]
\centering
\includegraphics[width=0.7\linewidth]{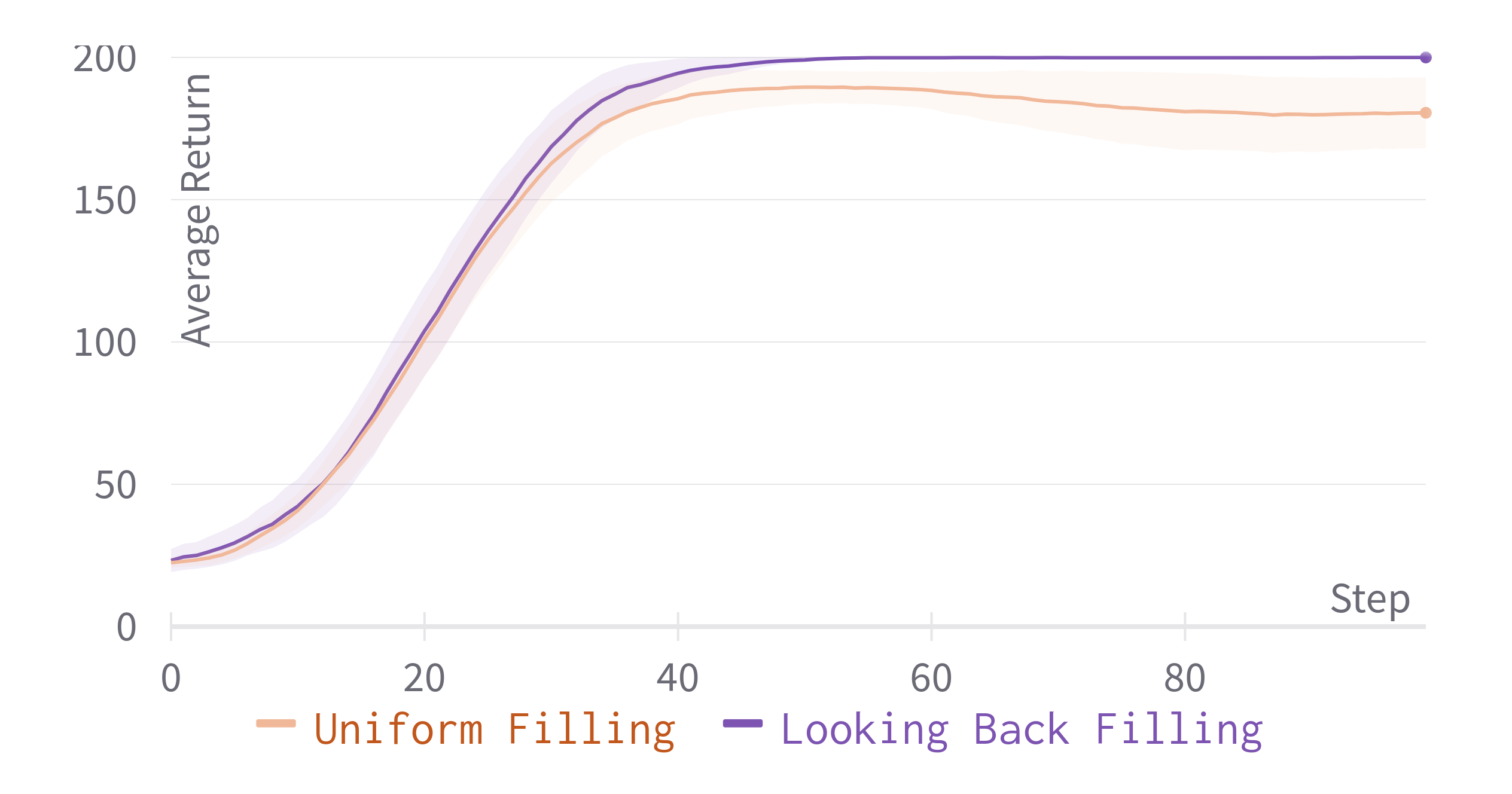}
\caption{Ablation study of Filling Scheme of \namel~sampler on the CartPole Environment. Here ``Uniform Filling'' denotes the uniformly random sampling of states to fill after sampling the pivot state, and ``Looking Back Filling'' denotes our proposed approach (\namel).}
\label{fig:looking_back}
\end{figure}

Table~\ref{tab:looking_back} and Figure~\ref{fig:looking_back} depict the buffer batch size sensitivity results from our proposed sampler.

\begin{table}[h!]
    \centering
    \caption{Importance of looking back in our proposed approach (\namel) on the CartPole environment.}
    \vspace*{1mm}
    \label{tab:looking_back}
    \resizebox{0.4\columnwidth}{!}{
    \begin{tabular}{lcl}
        \toprule
        \textbf{Filling Scheme} & {\textit{\textbf{Average Reward}}}\\ \midrule
\midrule

\text{Looking Back (\namel)}  & \textbf{199.83}  {\tiny $\pm$ 0.31} \\ \midrule
\text{Uniform (\namel)} & 182.5 {\tiny $\pm$ 23.49} \\ 
\bottomrule
    \end{tabular}
    }
\end{table}

\section{Ablation study of Temporal effects}
\label{ablation_temporal}

\begin{figure}[h!]
\centering
\subfigure[CartPole-v0]{%
\includegraphics[width=0.30\linewidth]{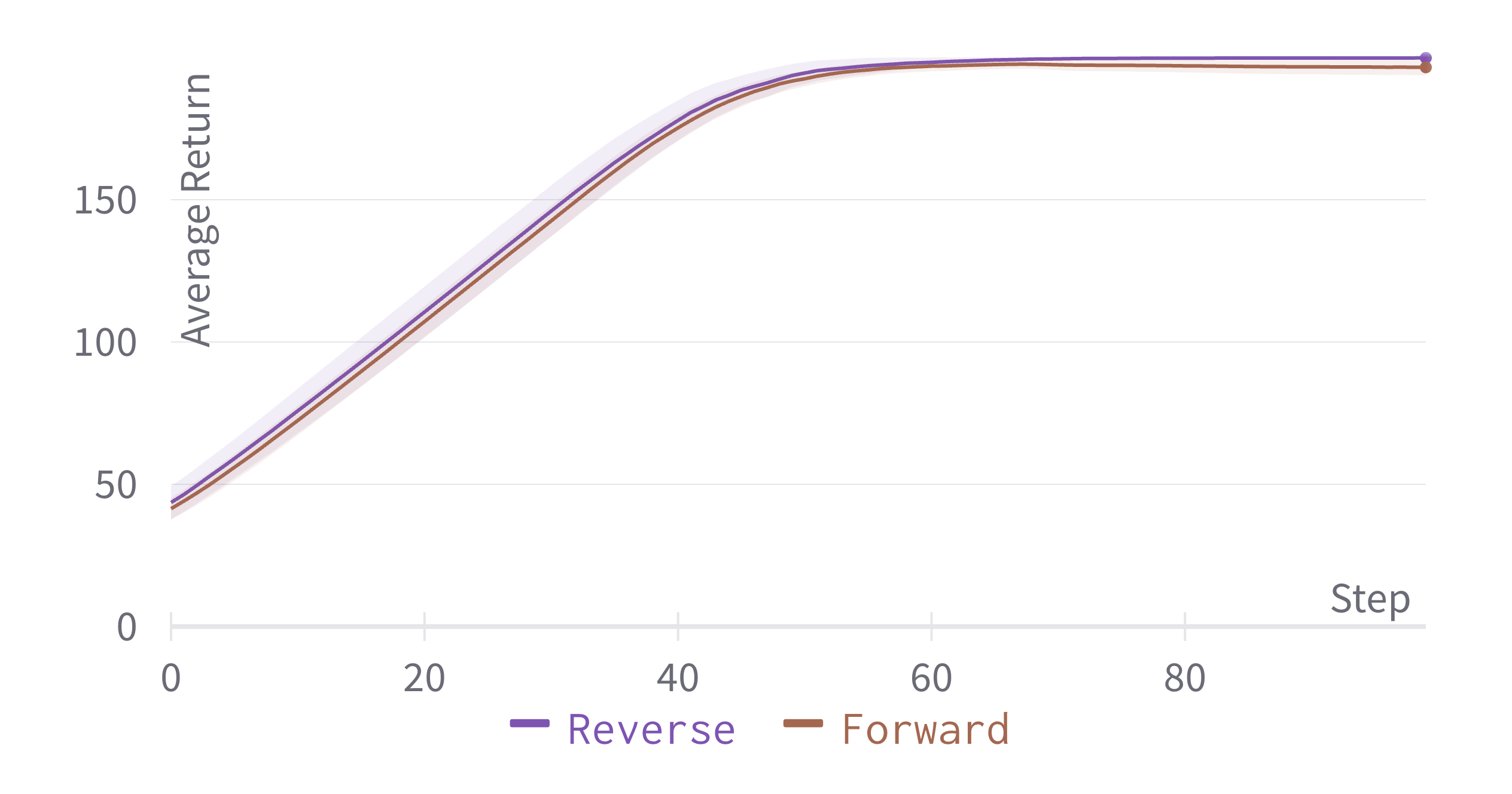}
}
\quad
\subfigure[Acrobot-v1]{%
\includegraphics[width=0.30\linewidth]{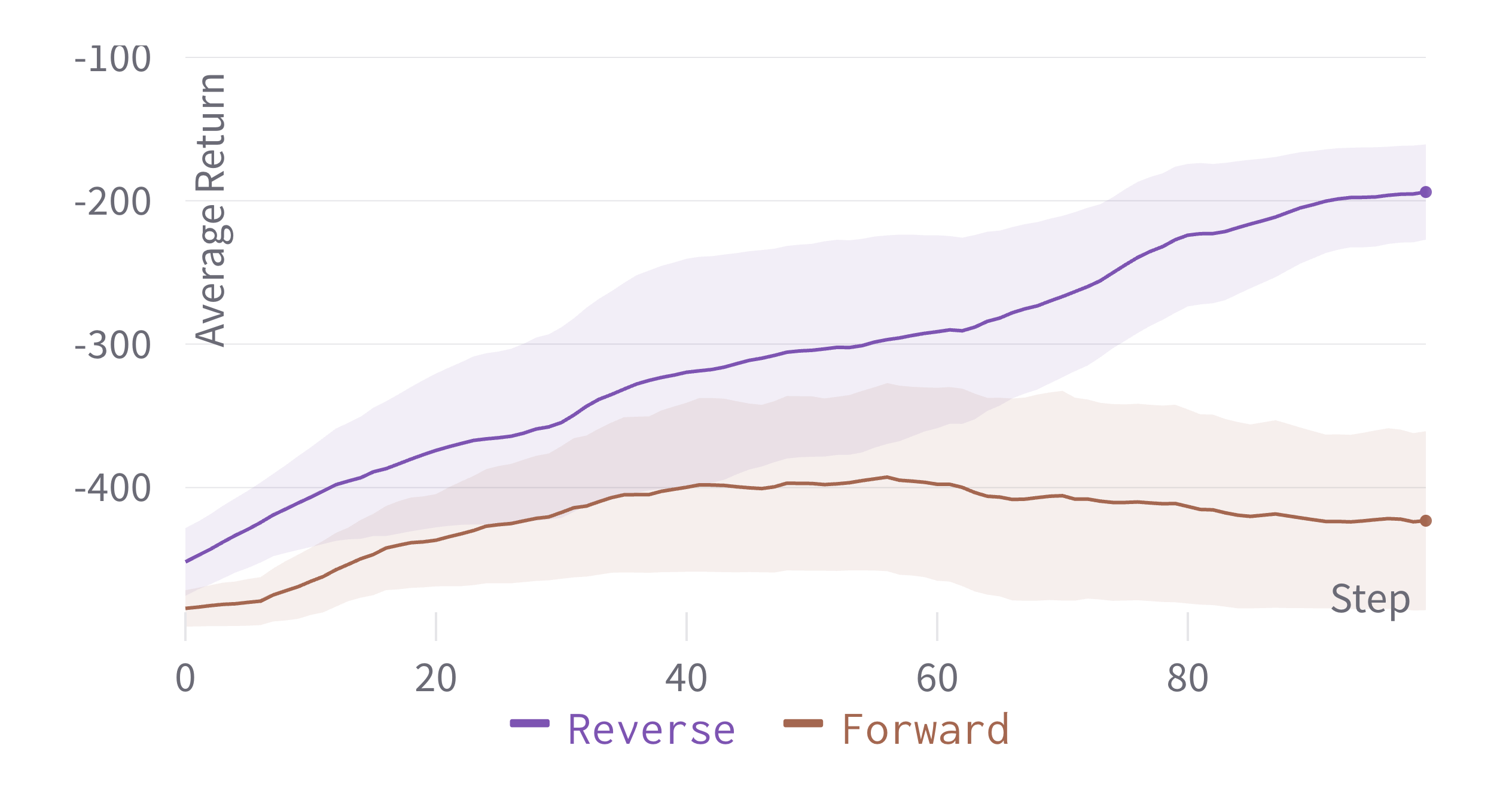}
}
\quad
\subfigure[Pendulum-v0]{%
\includegraphics[width=0.30\linewidth]{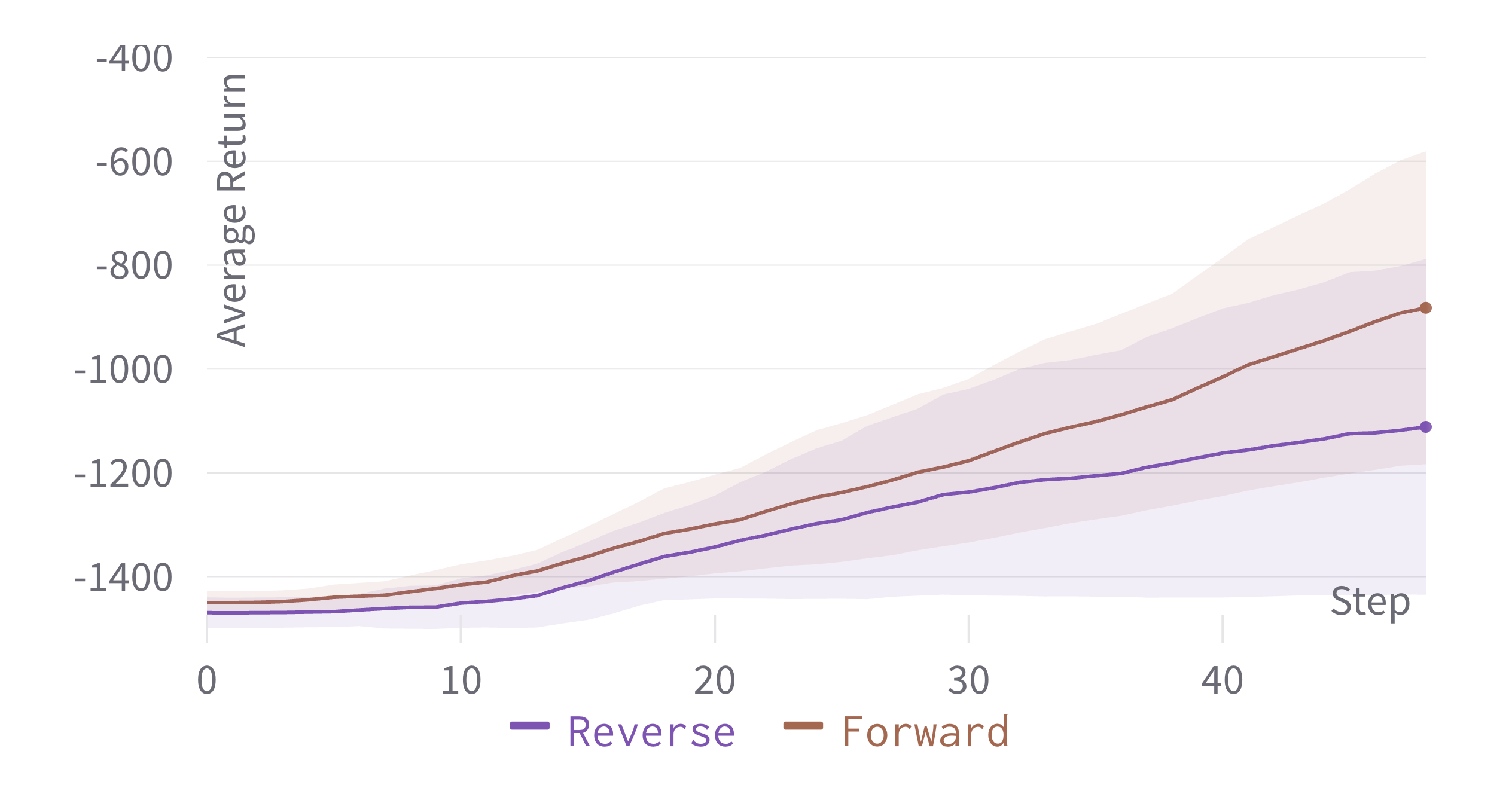}
}

\subfigure[LunarLander-v2]{%
\includegraphics[width=0.30\linewidth]{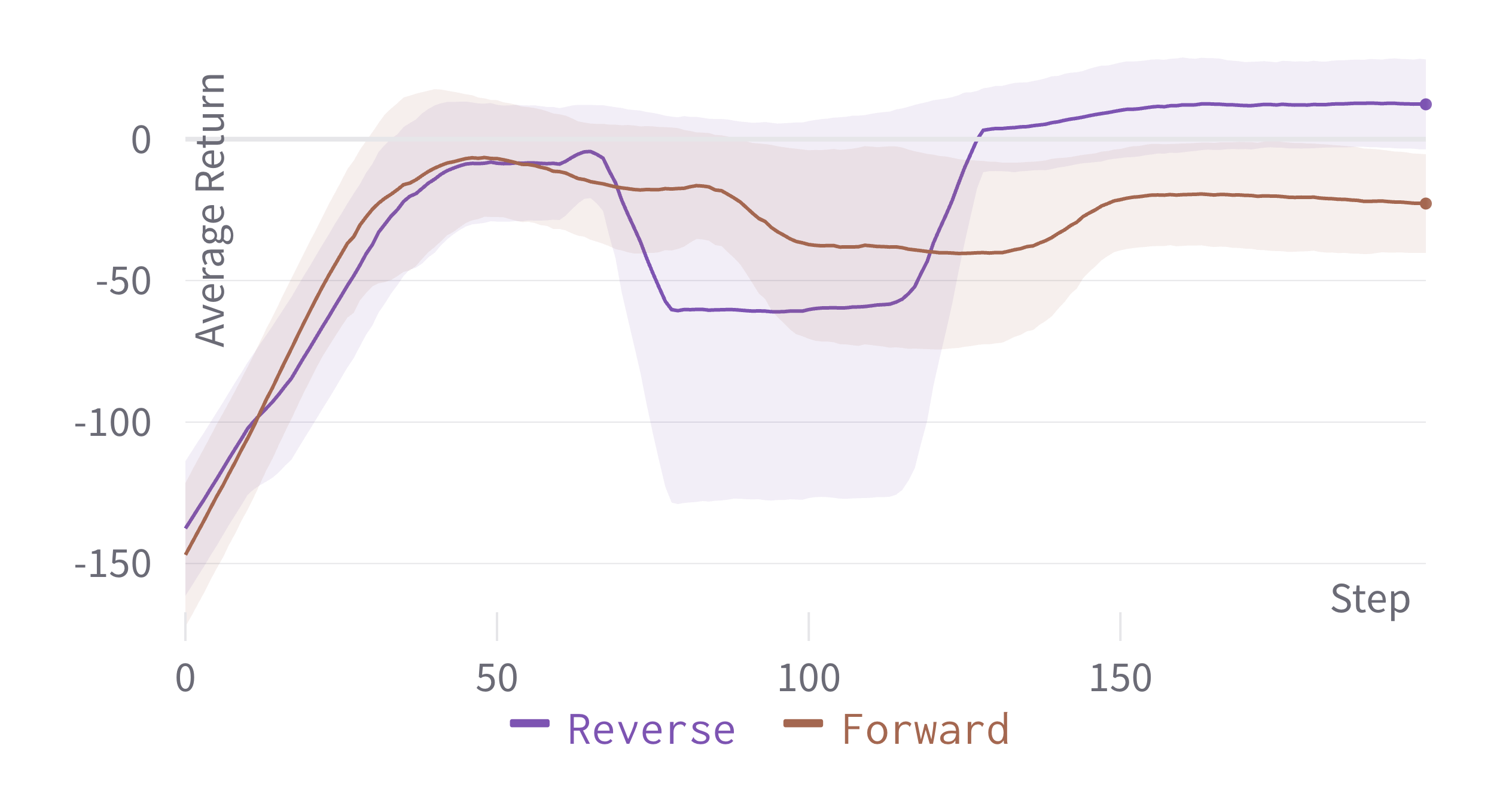}
}
\quad
\subfigure[HalfCheetah-v2]{%
\includegraphics[width=0.30\linewidth]{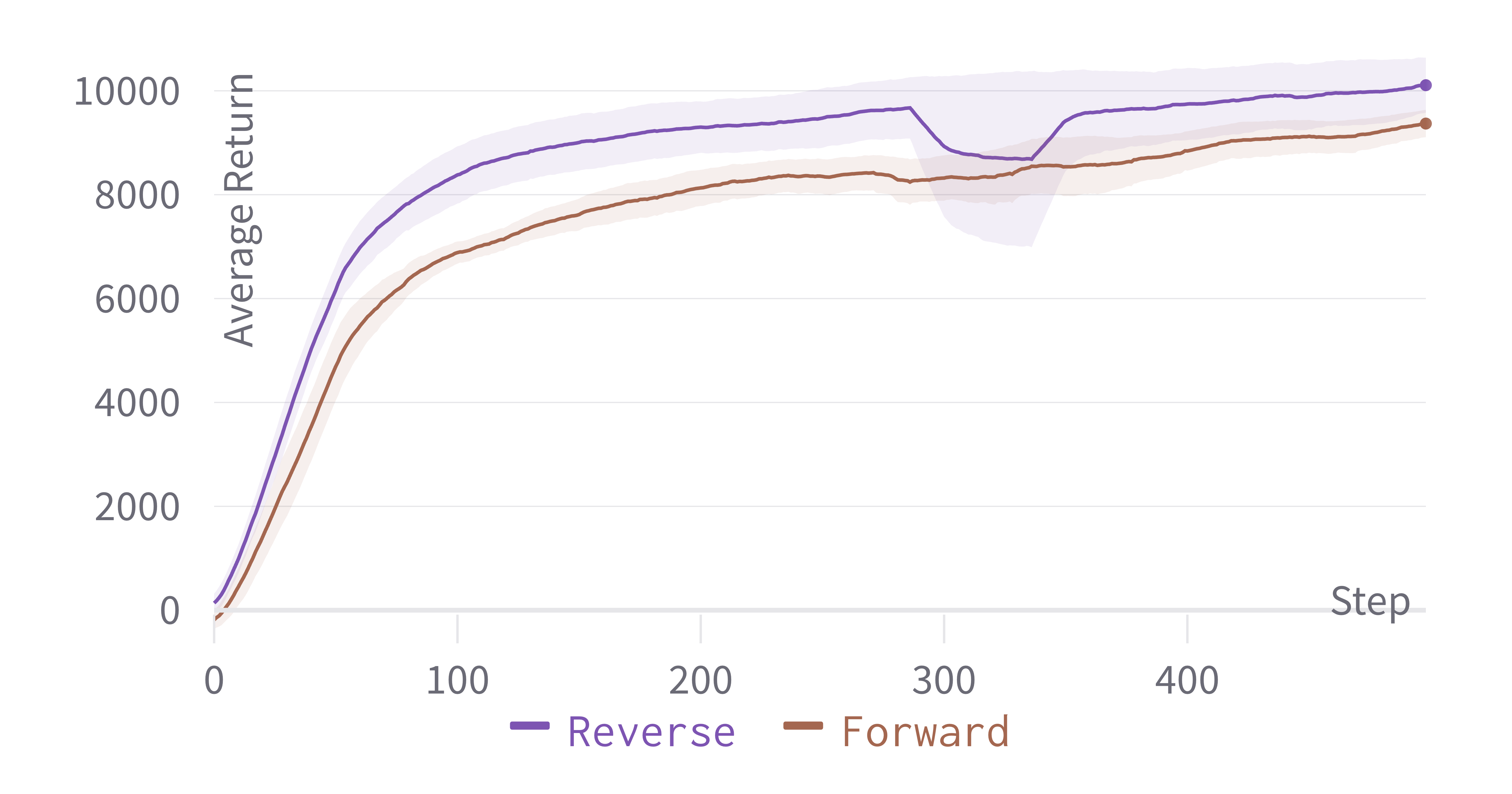}
}
\quad
\subfigure[Ant-v2]{%
\includegraphics[width=0.30\linewidth]{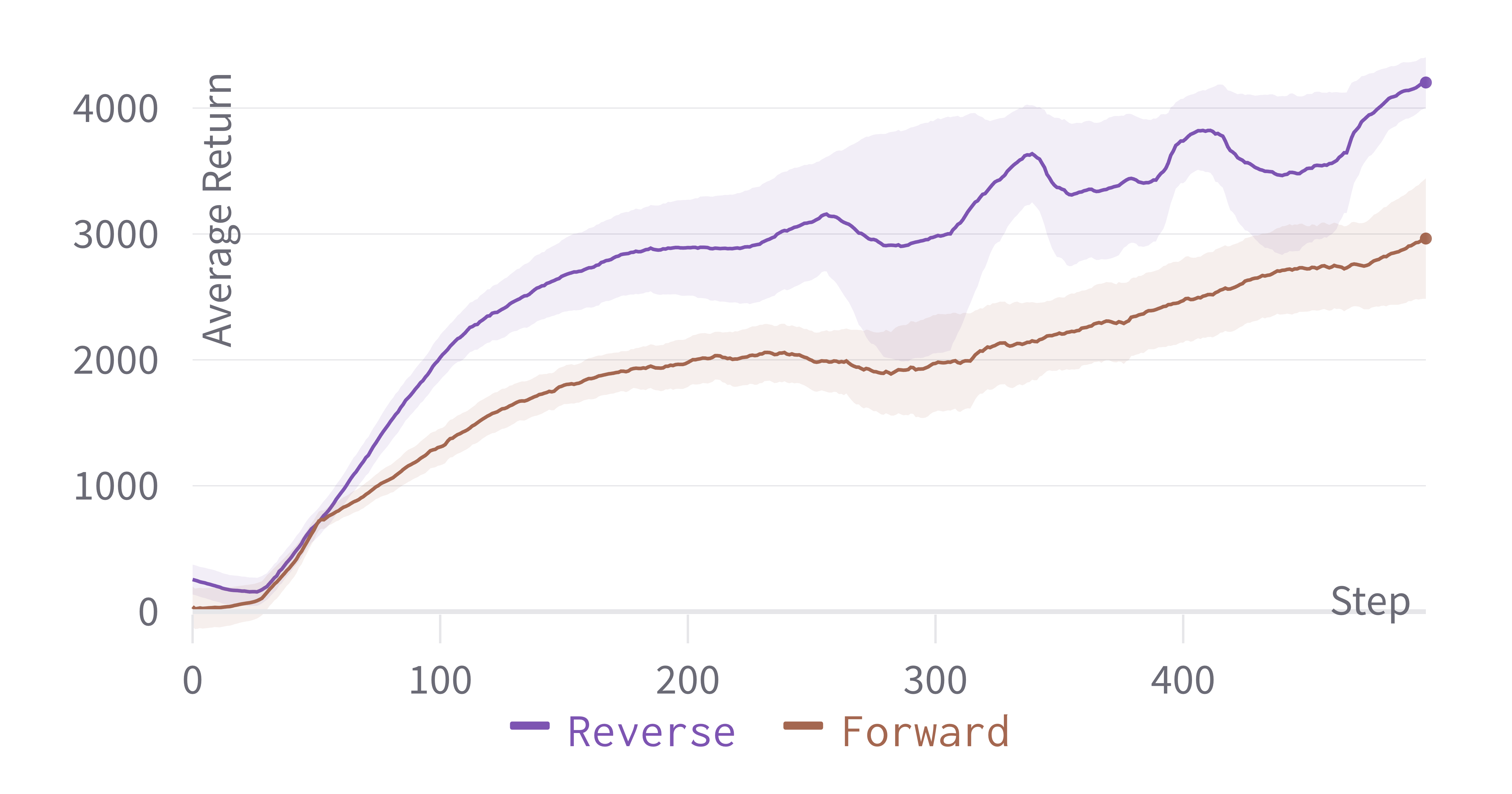}
}

\subfigure[Reacher-v2]{%
\includegraphics[width=0.30\linewidth]{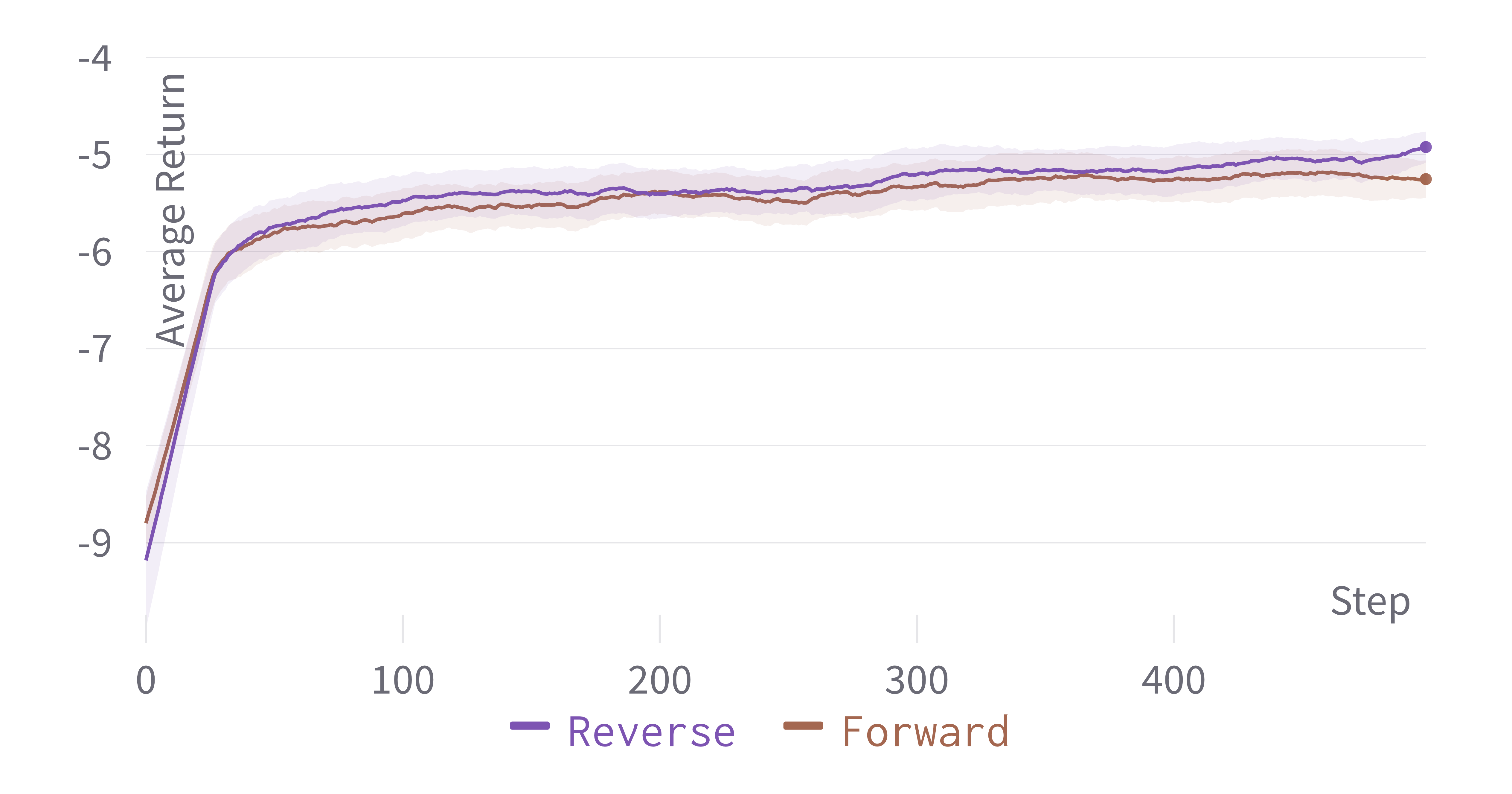}
}
\quad
\subfigure[Walker-v2]{%
\includegraphics[width=0.30\linewidth]{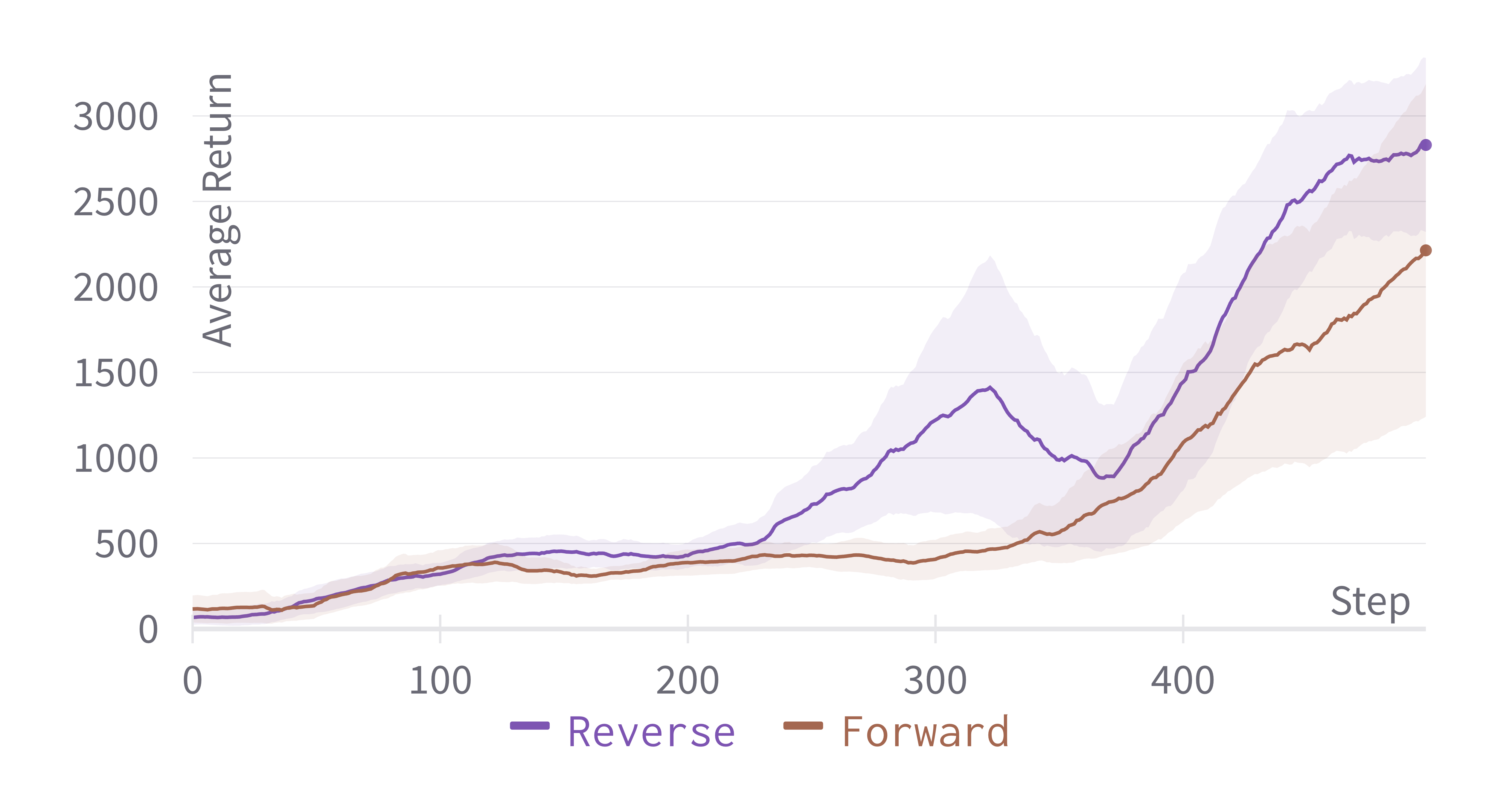}
}
\quad
\subfigure[Hopper-v2]{%
\includegraphics[width=0.30\linewidth]{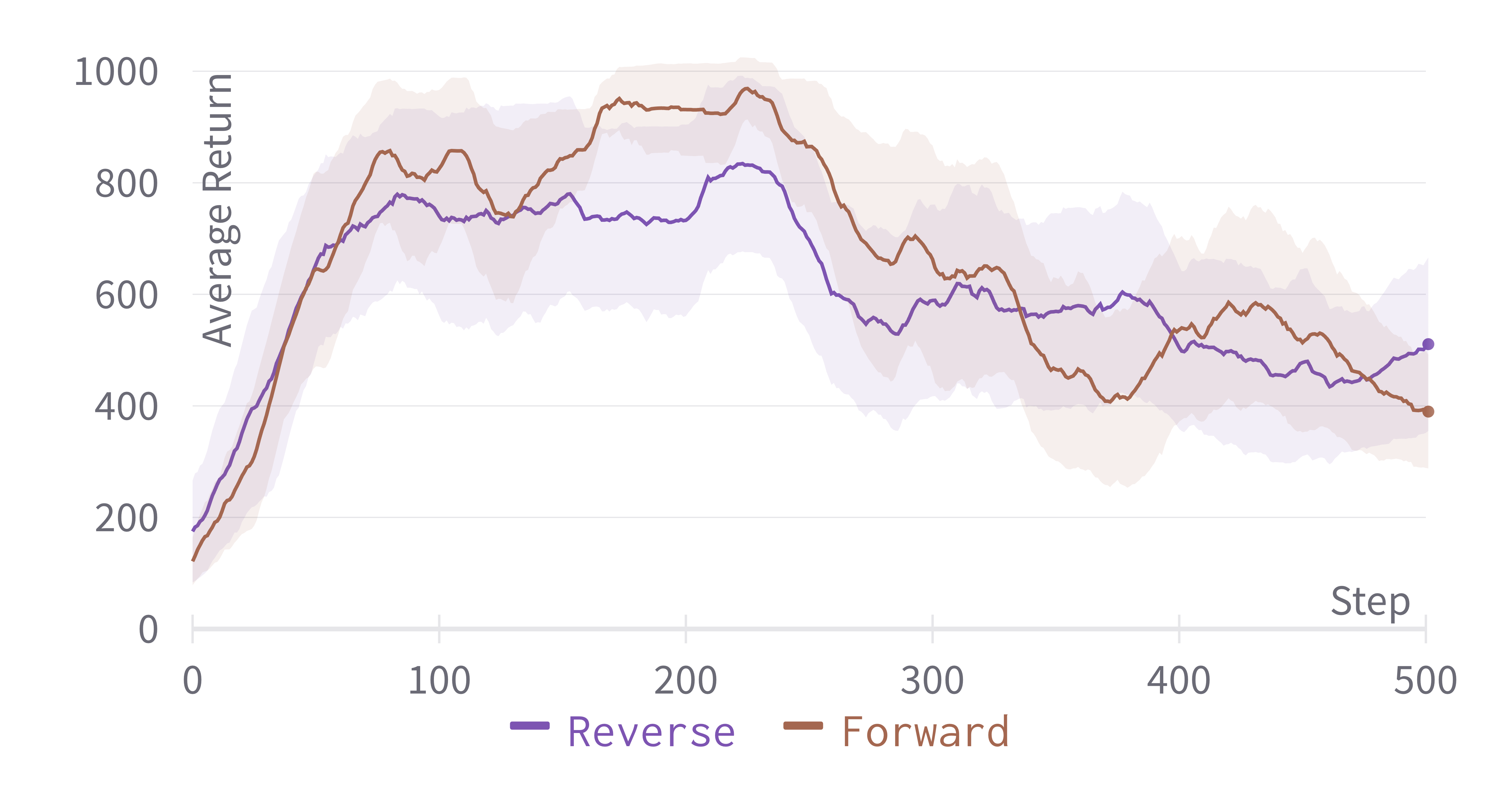}
}

\subfigure[InvertedDoublePendulum-v2]{%
\includegraphics[width=0.30\linewidth]{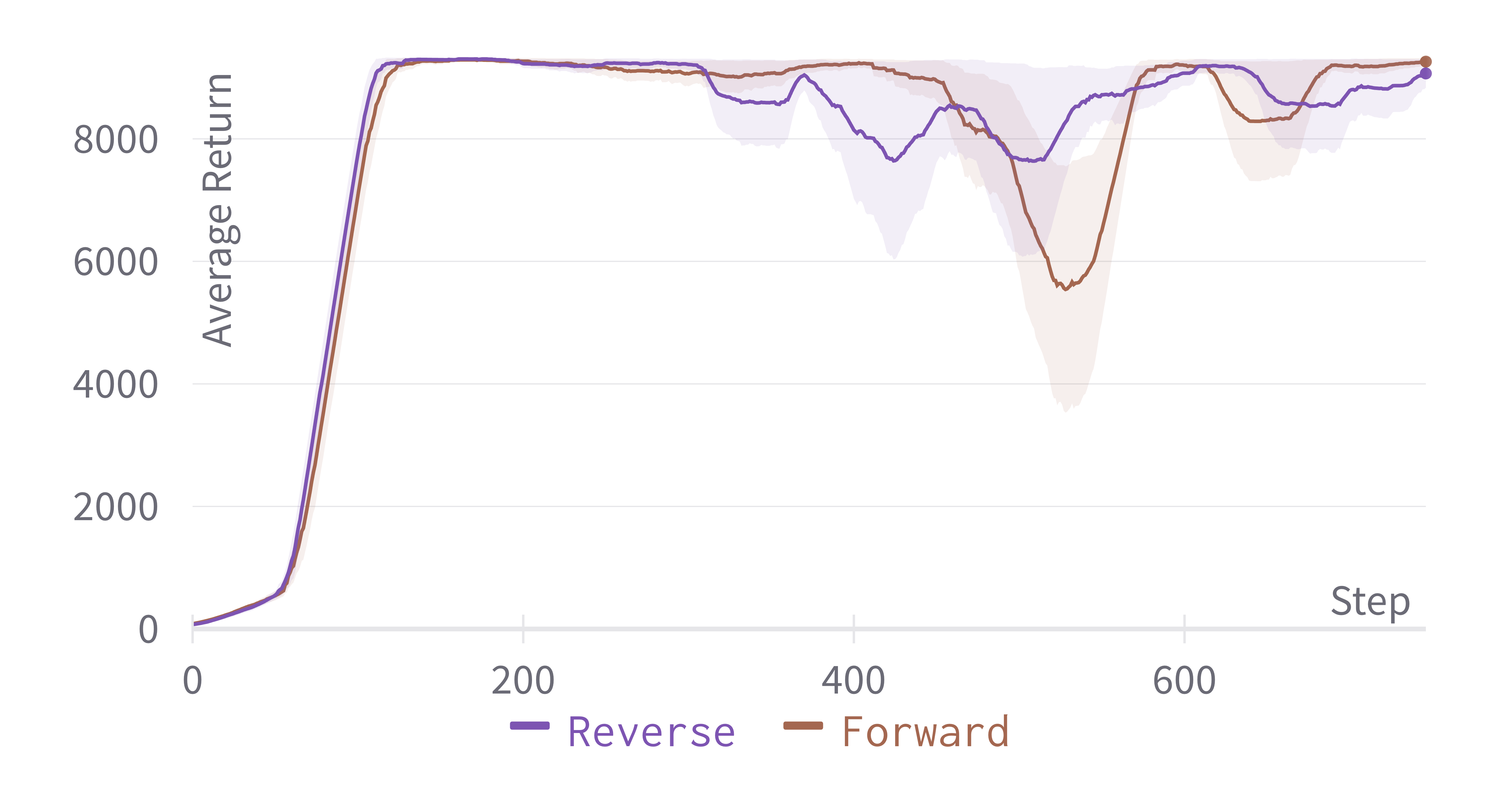}
}
\quad
\subfigure[FetchReach-v1]{%
\includegraphics[width=0.30\linewidth]{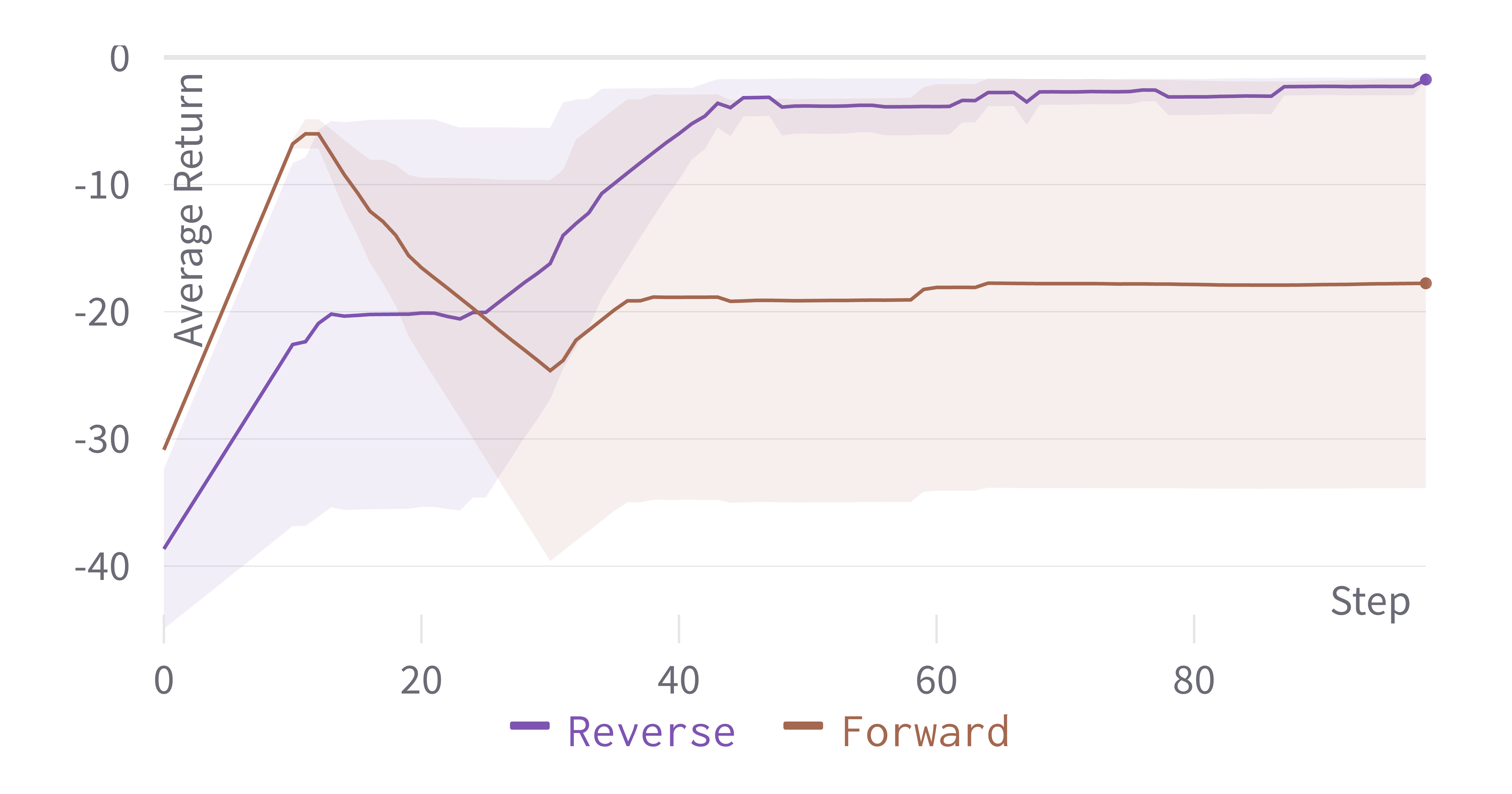}
}
\quad
\subfigure[Pong-v0]{%
\includegraphics[width=0.30\linewidth]{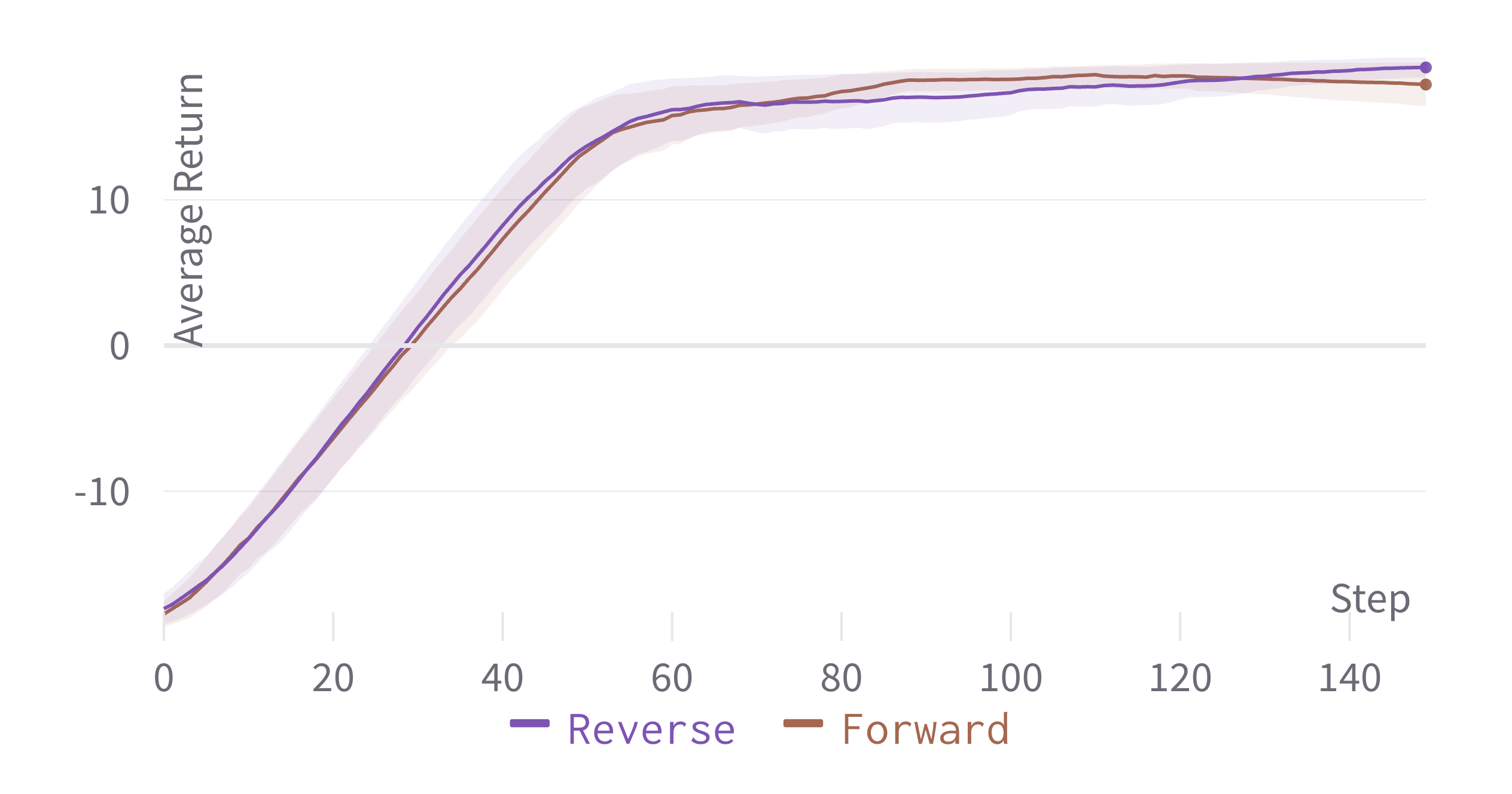}
}

\subfigure[Enduro-v0]{%
\includegraphics[width=0.30\linewidth]{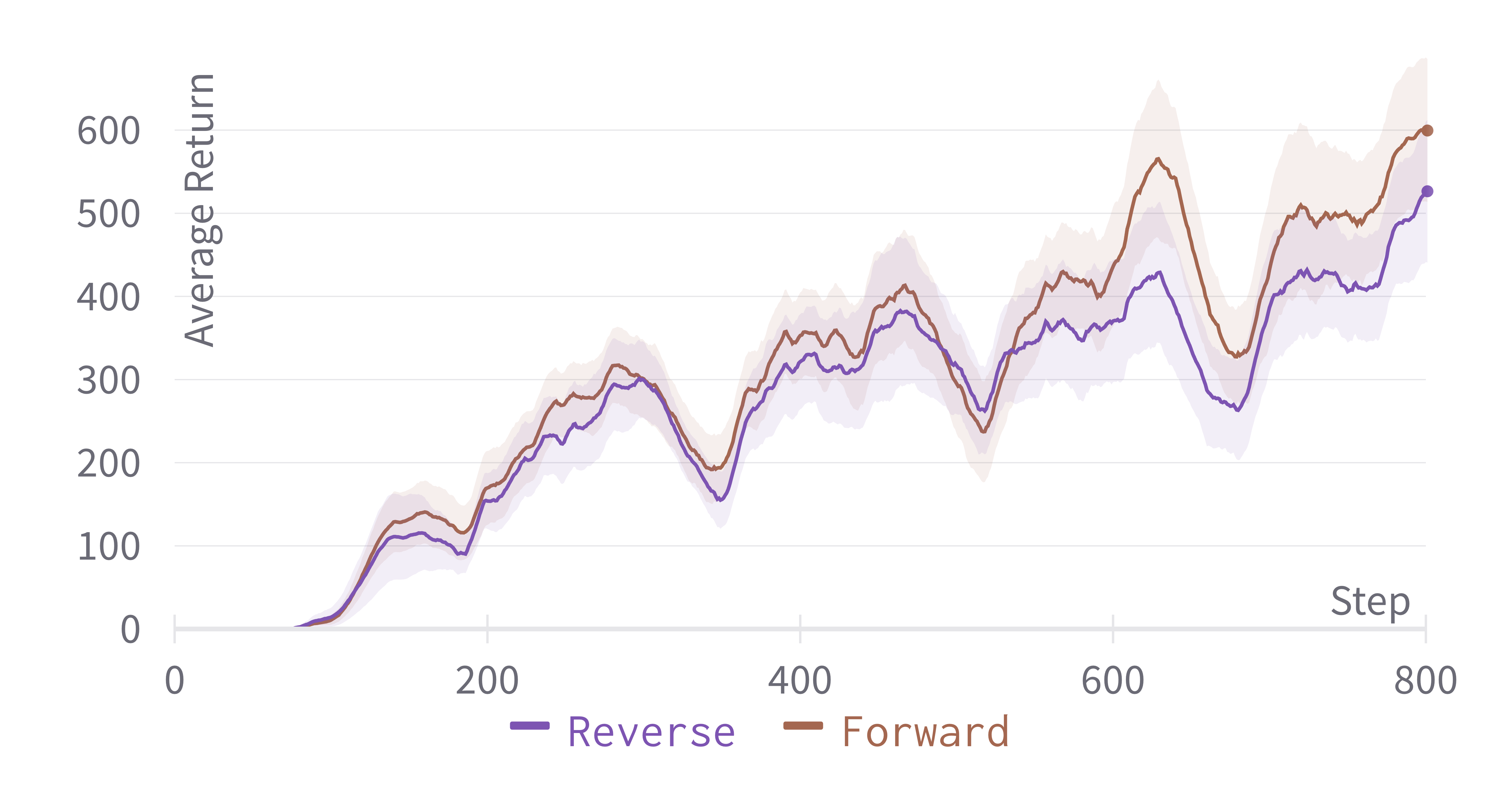}
}
\caption{Ablation study of the effects of the temporal structure on the performance of the agent.}
\label{fig:ablation_study_temporal}
\end{figure}

This section studies the ablation effects of going temporally forward and backward once we choose a pivot/surprise point. Furthermore, Figure~\ref{fig:ablation_study_temporal} depicts the learning curves of the two proposed methodologies. The forward sampling scheme is worse in most environments compared to the reverse sampling scheme.

\section{Sparsity and Rewards of Surprising States}
\label{sparse_surprise}

\subsection{Surprising States Have Large Rewards}
In this section, we study the ``learning from sparse reward'' intuition provided in Section~\ref{understanding_rer} -- i.e., we want to check if the states corresponding to large TD error correspond to states with large (positive or negative) rewards. To test the hypothesis, we consider a sampled buffer and plot the TD error of these points in the buffer against the respective reward. Figure~\ref{fig:scatterplot_ant} shows the distribution of TD error against reward for the sampled buffers in the Ant environment. We see that high reward states (positive or negative) also have higher TD errors. Therefore, our algorithm picks large reward states as endpoints to learn in such environments. 

\subsection{Surprising States are Sparse and Isolated}
\begin{figure}[h!]
\centering
\subfigure[UER]{%
\includegraphics[width=0.29\linewidth]{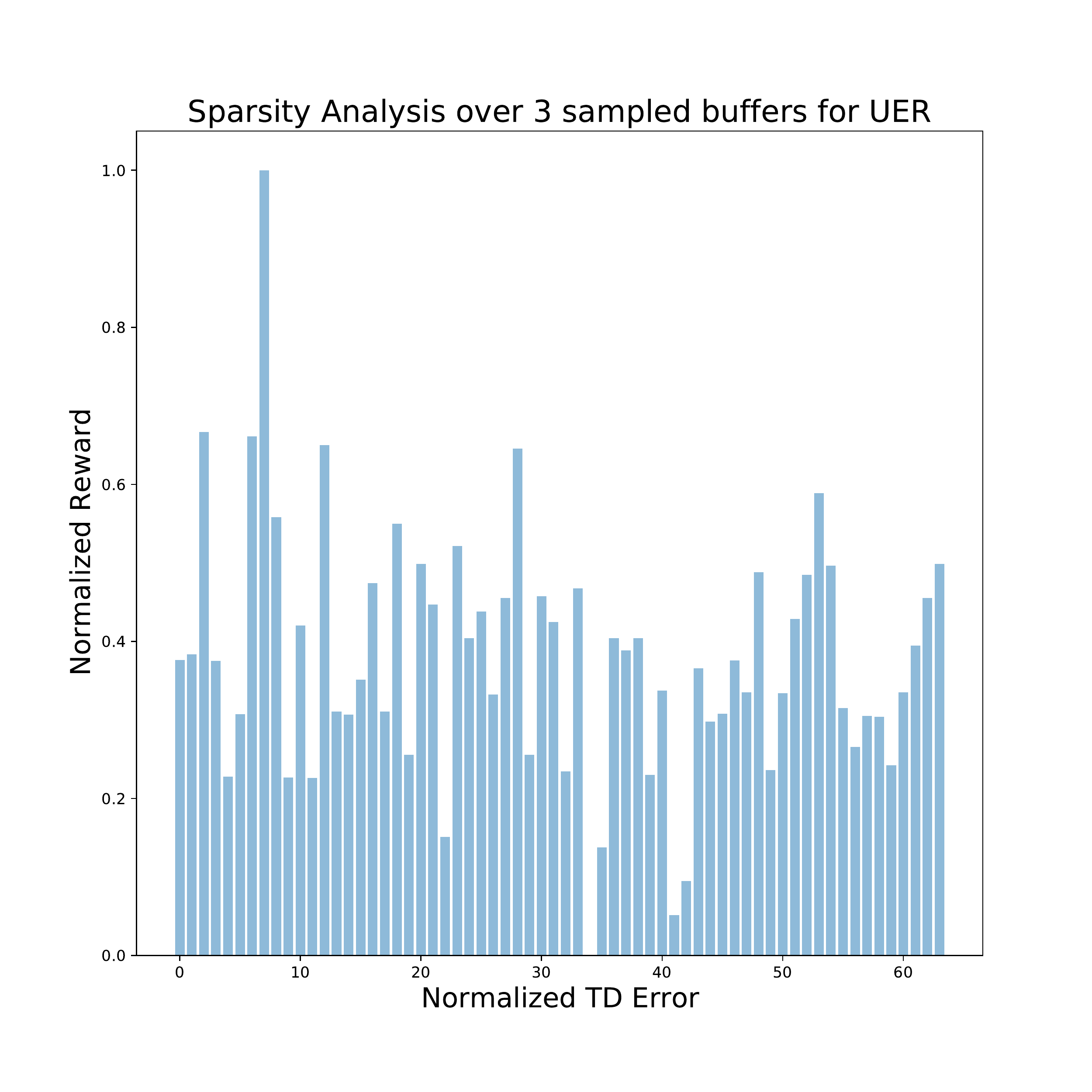}
\label{fig:uer_cartpole_sparsity}}
\subfigure[RER]{%
\includegraphics[width=0.29\linewidth]{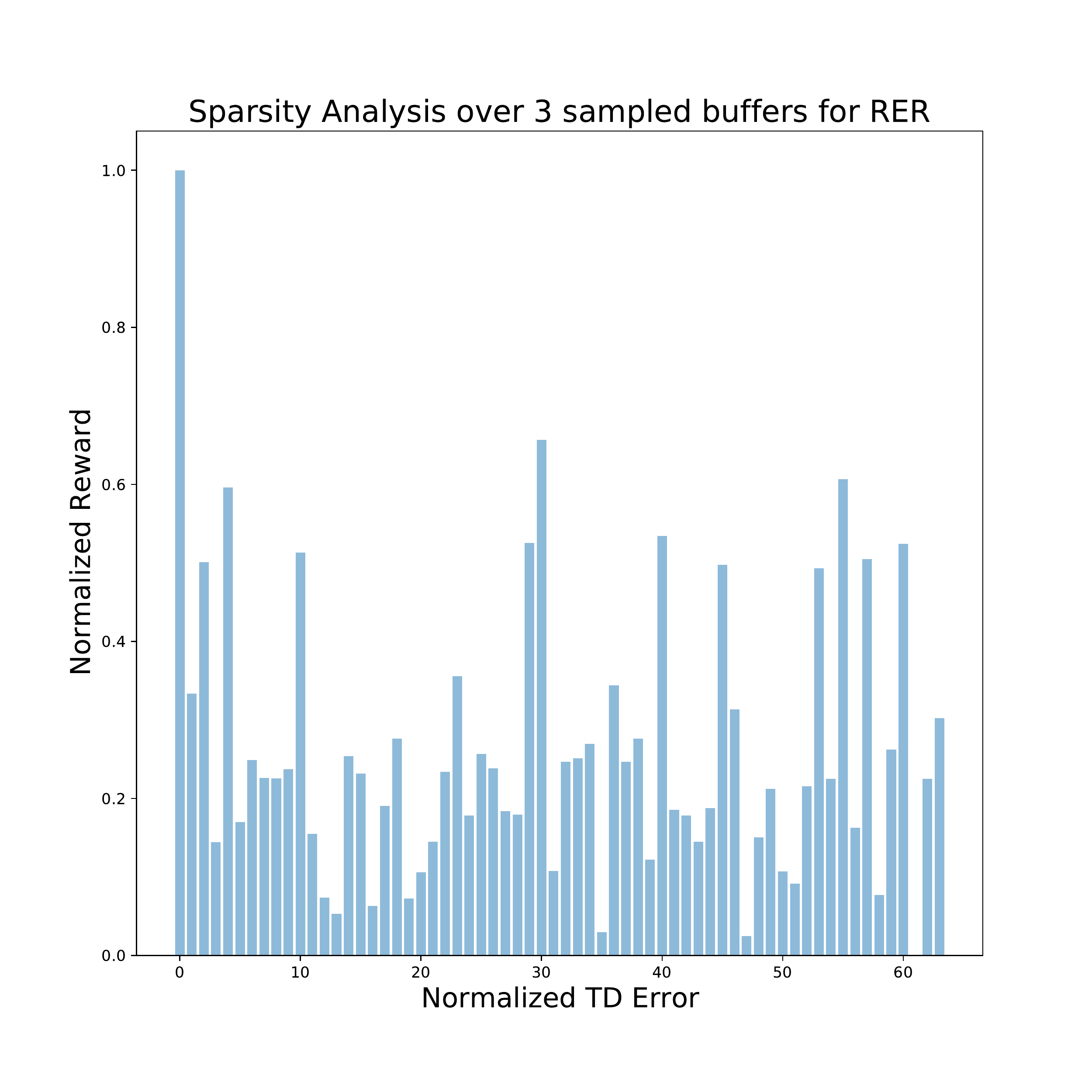}
\label{fig:rer_cartpole_sparsity}}
\subfigure[IER]{%
\includegraphics[width=0.29\linewidth]{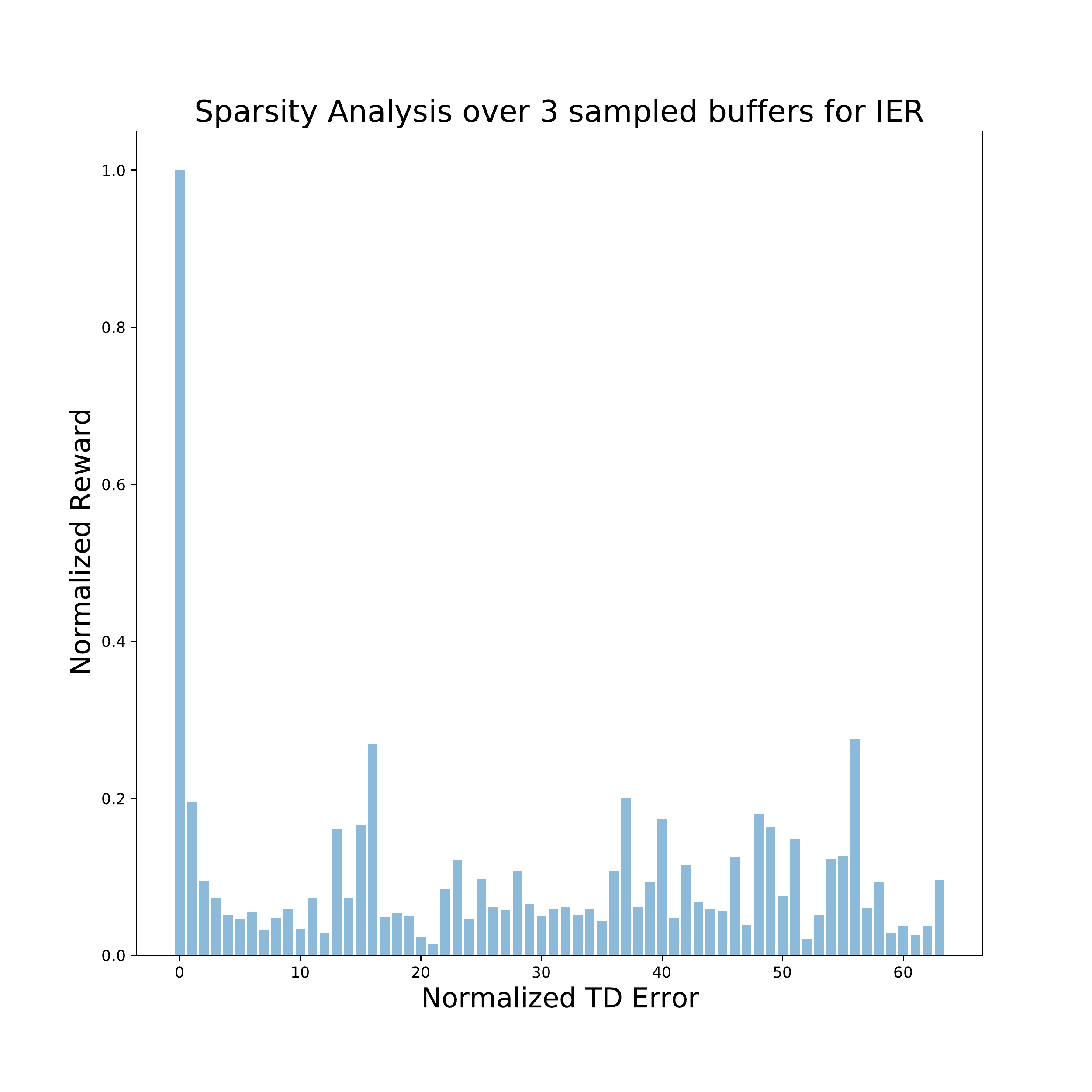}
\label{fig:rer++_cartpole_sparsity}}
\subfigure[IER Magnified]{%
\includegraphics[width=0.31\linewidth]{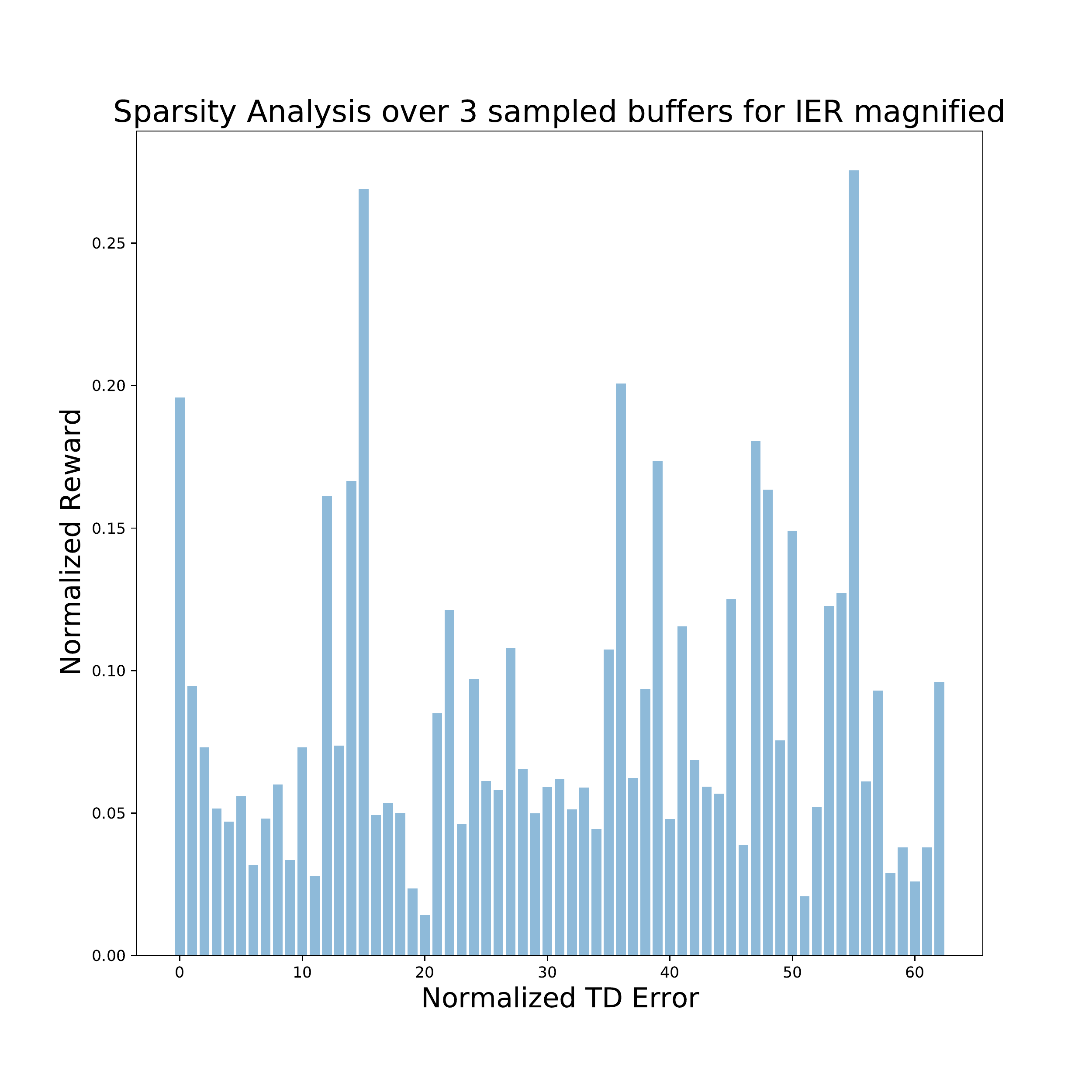}
\label{fig:rer++_magnified_cartpole_sparsity}}
\caption{Normalized TD Error ("Surprise factor") of each timestep over three different sampled buffers on the CartPole environment. Best viewed when zoomed.}
\label{fig:sparse_cartpole}
\end{figure}

\begin{figure}[h!]
\centering
\subfigure[UER]{%
\includegraphics[width=0.29\linewidth]{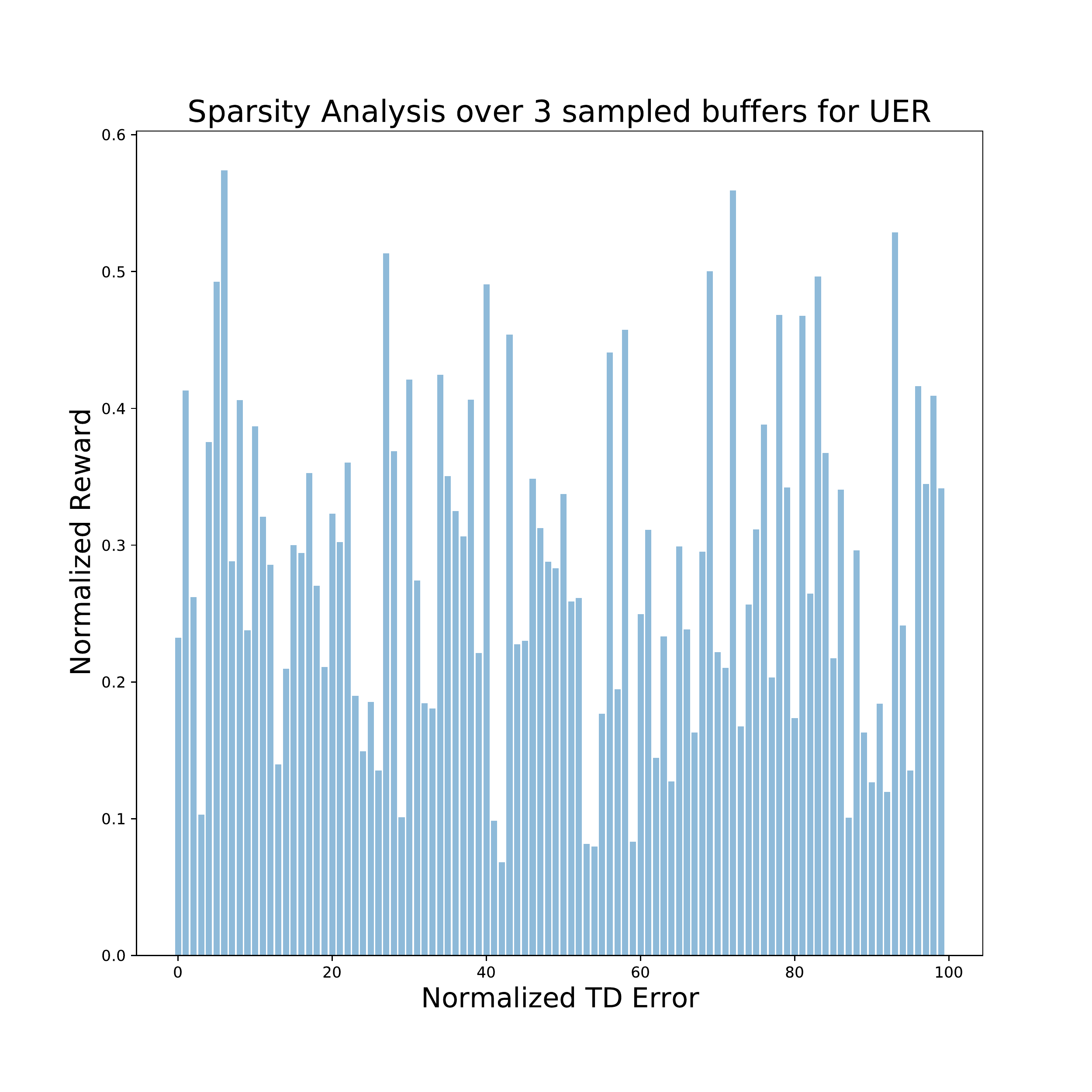}
\label{fig:uer_ant_sparsity}}
\subfigure[RER]{%
\includegraphics[width=0.29\linewidth]{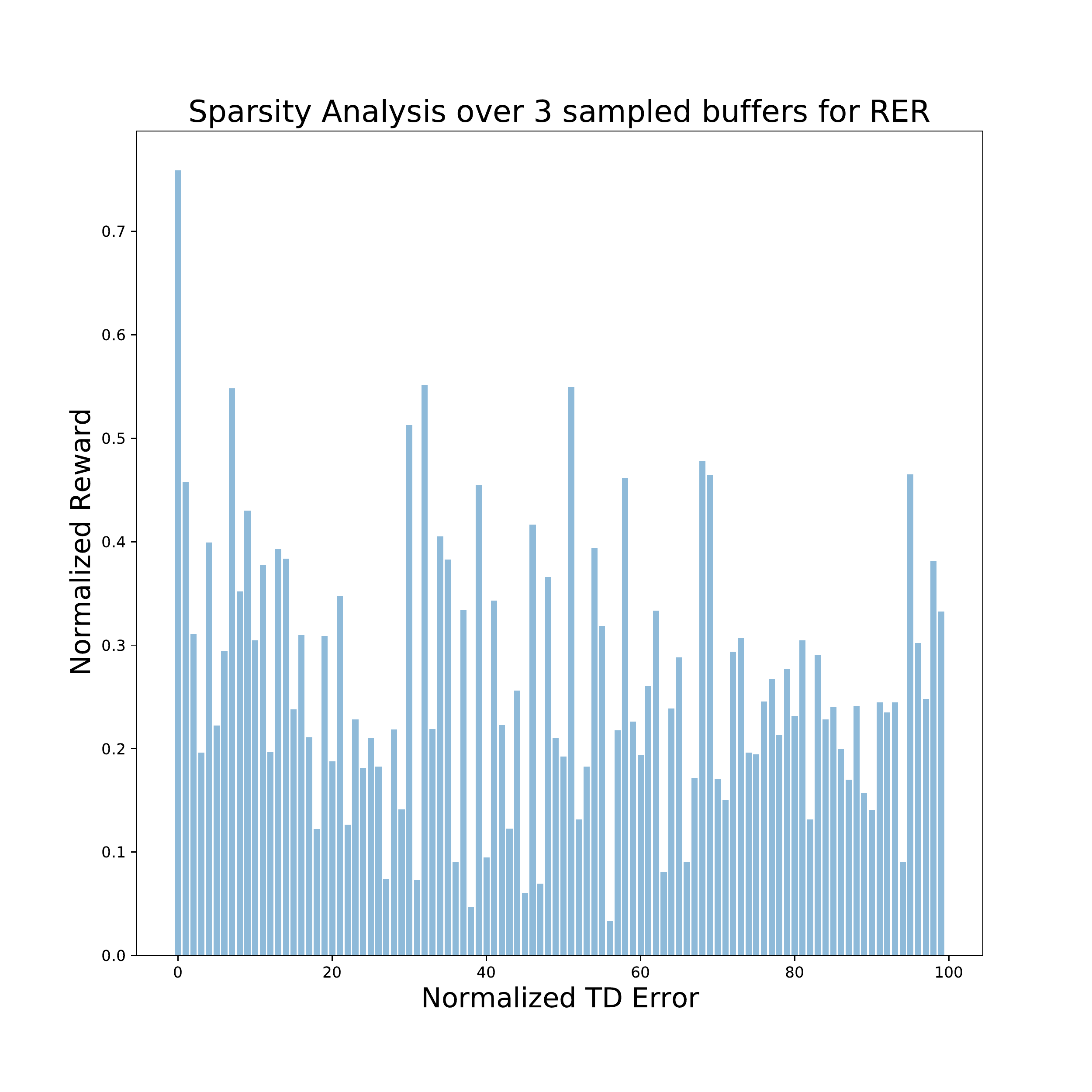}
\label{fig:rer_ant_sparsity}}
\subfigure[IER]{%
\includegraphics[width=0.29\linewidth]{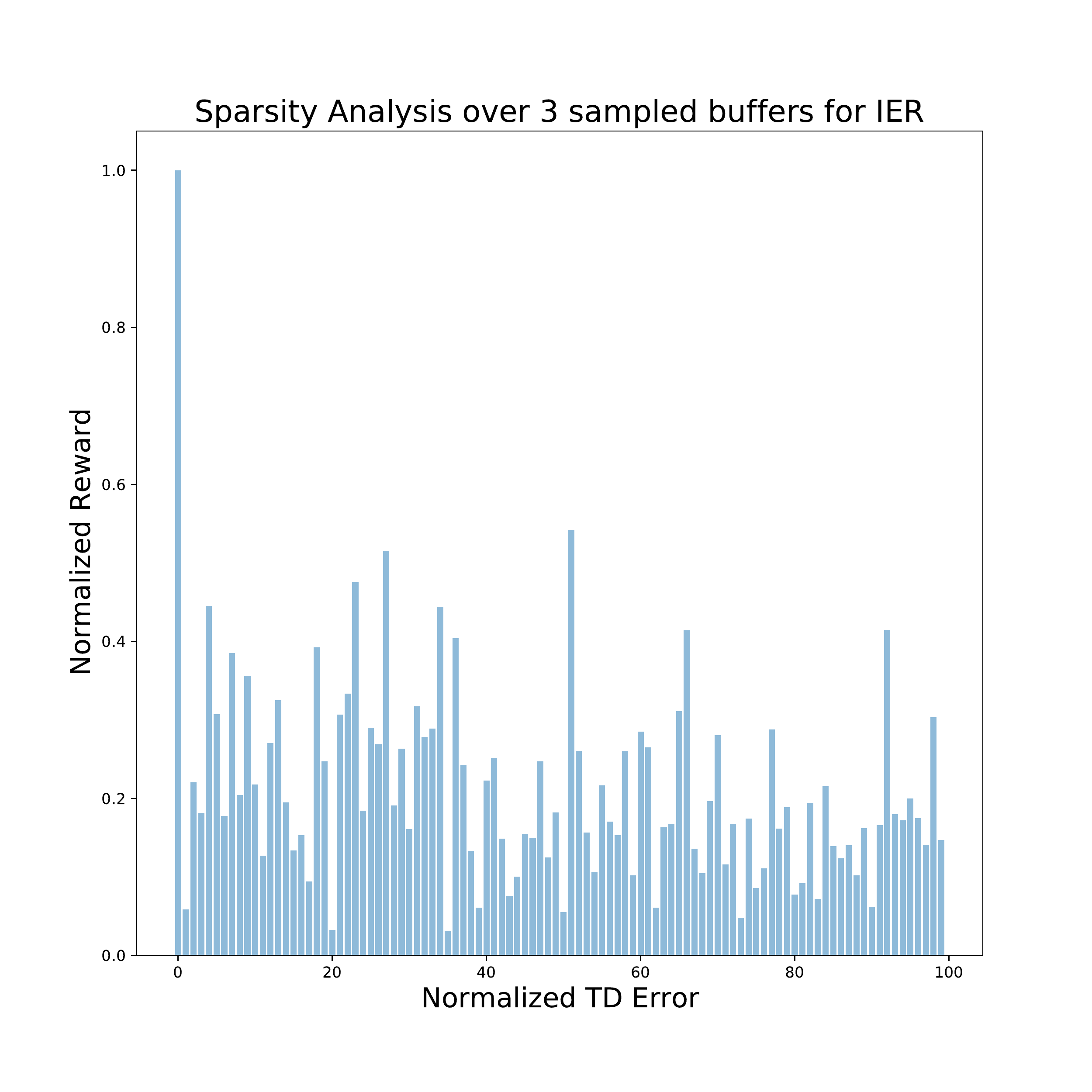}
\label{fig:rer++_ant_sparsity}}
\subfigure[IER Magnified]{%
\includegraphics[width=0.29\linewidth]{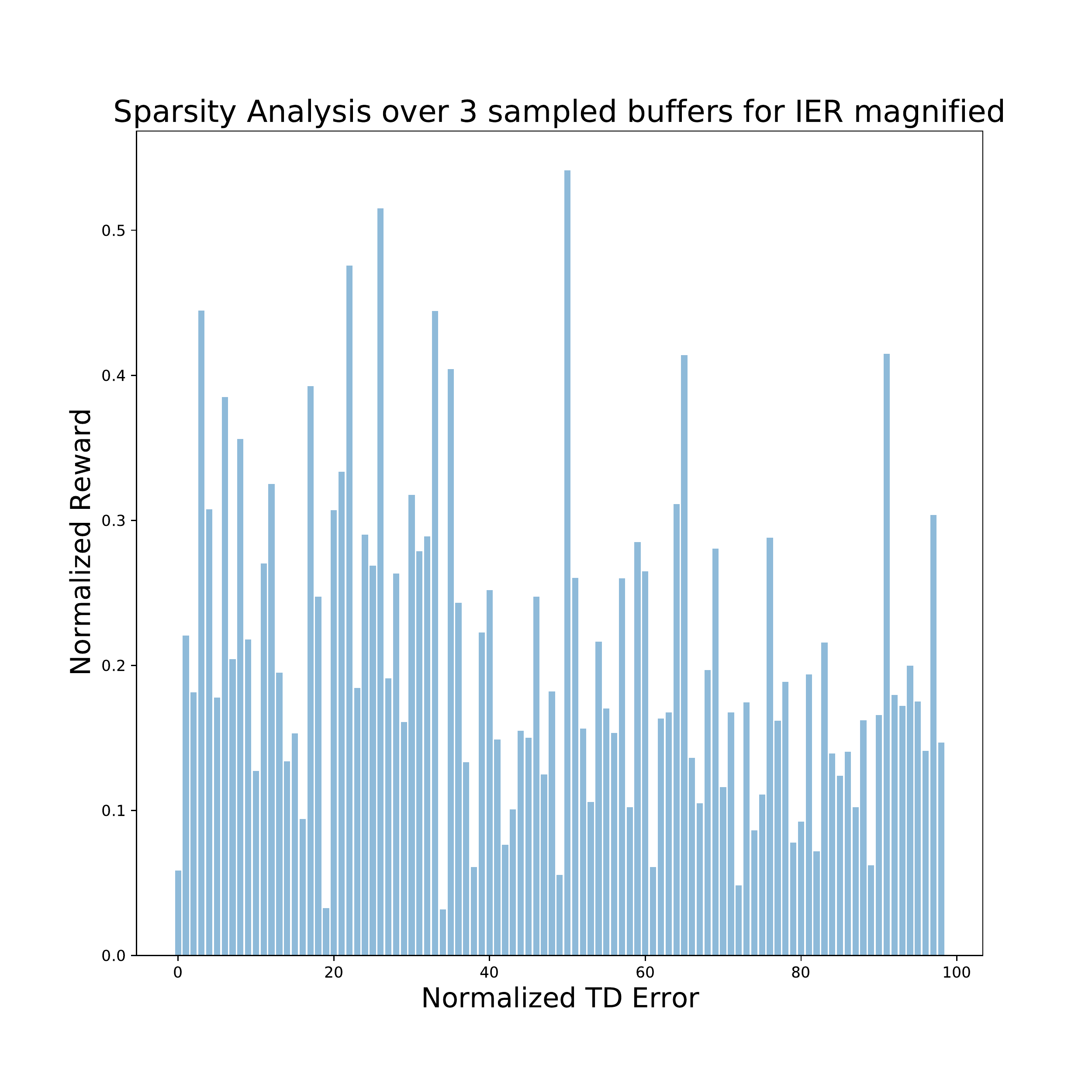}
\label{fig:rer++_magnified_ant_sparsity}}

\caption{Normalized TD Error ("Surprise factor") of each timestep over three different sampled buffers on the Ant environment. Best viewed when zoomed.}
\label{fig:sparse_ant}
\end{figure}

Figure~\ref{fig:sparse_cartpole} and Figure~\ref{fig:sparse_ant} depict the distribution of ``surprise''/TD error in a sampled batch for CartPole and Ant environments respectively. These two figures help show that the states with a large ``surprise'' factor are few and that even though the pivot of a buffer has a large TD error, the rest of the buffer typically does not.

Figure~\ref{fig:rer++_magnified_cartpole_sparsity} and Figure~\ref{fig:rer++_magnified_ant_sparsity} show a magnified view of Figure~\ref{fig:rer++_cartpole_sparsity} and Figure~\ref{fig:rer++_magnified_ant_sparsity} where the pivot point selected is dropped. This helps with a uniform comparison with the remaining timesteps within the sampled buffer. Again, we notice little correlation between the timesteps within the sampled buffer.

\section{Reverse Experience Replay (\namelo)}
\label{rer}

\begin{figure}[h!]
\centering
\includegraphics[width=0.7\linewidth]{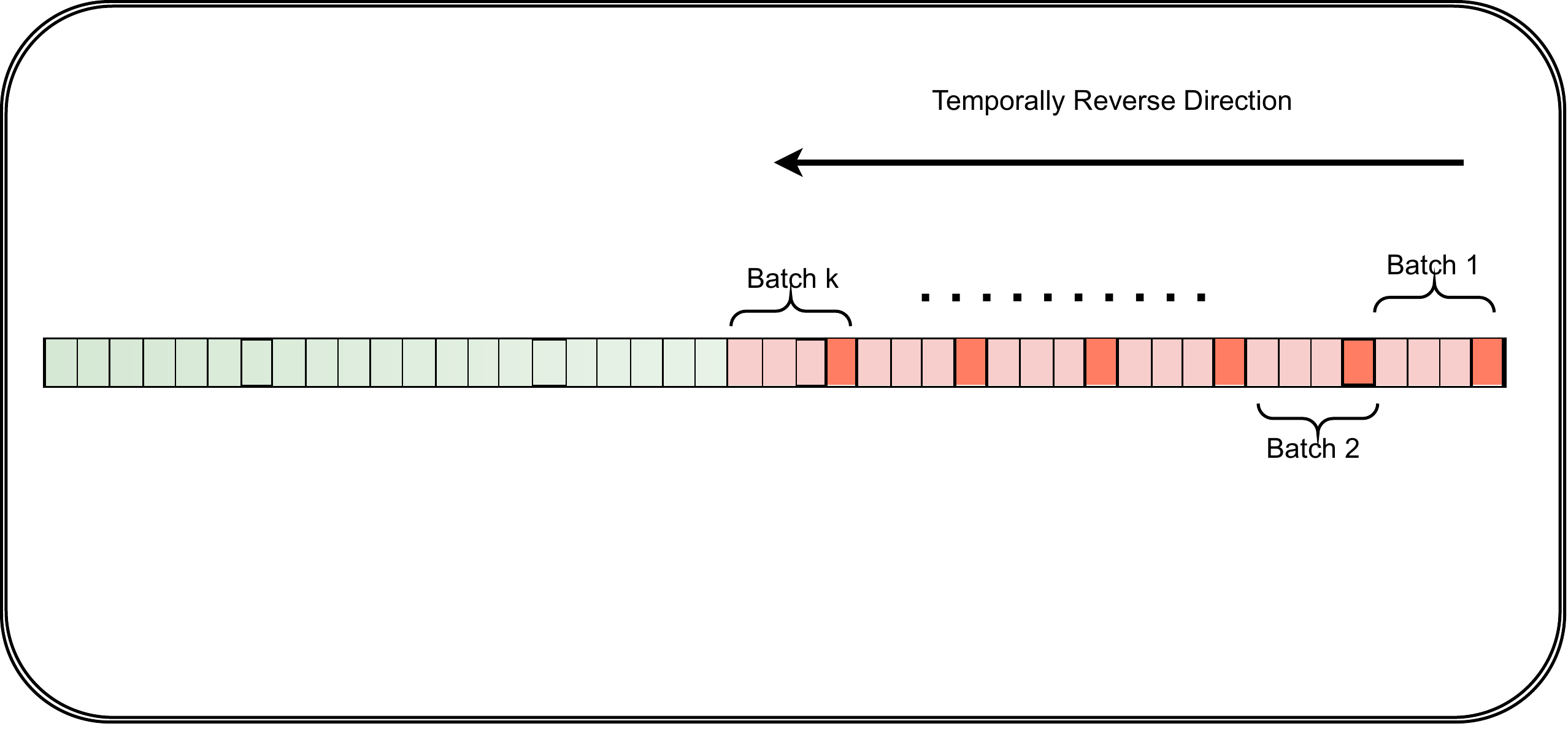}
\caption{An illustration of Reverse Experience Replay (RER) when selecting $k$ batches from the Replay Buffer.}
\label{fig:rer}
\end{figure}

This section discusses our implementation of Reverse Experience Replay (RER), which served as a motivation for our proposed approach. The summary of the RER approach is shown in Figure~\ref{fig:rer}. Furthermore, an overview of our implemented approach to RER is described briefly in Algorithm~\ref{alg:rer}.

\begin{algorithm}
\caption{Reverse Experience Replay}\label{alg:rer}
\KwIn{Data collection mechanism $\mathbb{T}$, Data buffer $\mathcal{H}$, Batch size $B$, grad steps per Epoch $G$, number of episodes $N$, learning procedure $\mathbb{A}$}
$n \leftarrow N$\;
$P \leftarrow \mathsf{len}(\mathcal{H})$ \tcp*{Set index to last element of Buffer $\mathcal{H}$}
\While{$n < N$}
{ $n \leftarrow n+1$\;
$\mathcal{H} \leftarrow \mathbb{T}(\mathcal{H})$ \tcp*{Add a new episode to the buffer}
$g \leftarrow 0$\;
\While{$g< G$}{
\eIf{$P-B<0$}{
$P \leftarrow \mathsf{len}(\mathcal{H})$ \tcp*{Set index to last element of Buffer $\mathcal{H}$}
}{
$P\leftarrow P-B$\;
}
$D \leftarrow \mathcal{H}[P-B,P]$ \tcp*{Load batch of previous $B$ samples from index $P$}
$g\leftarrow g+1$\;
$\mathbb{A}(D)$\tcp*{Run the learning algorithm with batch data $D$}
}
}
\end{algorithm}

\end{document}